\documentclass[sigconf]{acmart}
\AtBeginDocument{%
  }

\keywords{Autonomous agents, execution-time authorization, policy enforcement, provenance, distributed systems}

\usepackage{booktabs}
\usepackage{array}
\usepackage{listings}
\usepackage{caption}
\usepackage{subcaption}
\usepackage{tikz}
\usepackage{pgfplots}
\pgfplotsset{compat=1.18}
\usepackage{fancyhdr}

\setcopyright{none}
\renewcommand\footnotetextcopyrightpermission[1]{}
\acmConference{} % Remove conference header

\begin{document}

\title{Faramesh: A Protocol-Agnostic Execution Control Plane for Autonomous Agent systems }

\author{Amjad Fatmi}
\affiliation{%
  \institution{The Faramesh Labs}
  \city{New York}
  \country{USA}}
\email{press@faramesh.dev}

\begin{abstract}
Autonomous agents are increasingly making decisions that trigger real-world actions, affecting systems, data, and infrastructure. However, current agent frameworks, such as those utilizing MCP, UTCP, or A2A, fail to define whether such decisions are safely executable once proposed. To address this gap, we introduce the \textbf{Action Authorization Boundary (AAB)}, a mandatory, non-bypassable enforcement layer that mediates agent-driven actions at the execution boundary. The AAB ensures that all proposed actions are authorized before being executed, thereby guaranteeing determinism, non-bypassability, and replayability.~\cite{1,3,4}

The AAB operates as a \textbf{runtime authorization boundary}, placed between the agent’s reasoning space and real-world execution. It enforces \textbf{deterministic authorization} over \textbf{canonical action representations}, ensuring that identical actions, when evaluated under the same state and policy, always produce the same outcome. This framework is \textbf{non-bypassable}, meaning that no execution can occur without passing through the AAB, even in the presence of potentially compromised agent reasoning. Furthermore, the AAB guarantees \textbf{fail-closed semantics}, meaning that any failure in the authorization process results in denial or deferral of the proposed action, preventing unauthorized actions from being executed.

This paper demonstrates that such an enforcement boundary is \textbf{not optional} and \textbf{cannot be derived from IAM, logging systems, or existing protocol-based solutions}. The AAB introduces a structural guarantee that cannot be replaced by observability, gateways, protocols, or guardrails, which, while operationally helpful, fail \textbf{structurally} to enforce execution-time authorization. These systems do not bind execution to authorization, leaving gaps where malicious or erroneous actions could bypass control.

Faramesh introduces six architectural guarantees:
\begin{itemize}
    \item \textbf{Non-bypassable enforcement} of actions, ensuring all side-effecting operations pass through the AAB.
    \item \textbf{Deterministic authorization} of canonical actions, providing the basis for compliance, auditing, and post-incident forensics.
    \item \textbf{Replayable, provenance-complete decision records}, allowing for transparent auditability and traceability.
    \item A \textbf{policy-agnostic enforcement layer}, decoupled from policy authoring and governance models.
    \item \textbf{Fail-closed behavior}, where any boundary failure results in action denial, preventing unauthorized execution.
    \item \textbf{Multi-agent, multi-tenant support}, governing agents across organizational boundaries without compromising the core enforcement guarantees.
\end{itemize}

By establishing the AAB as an architectural invariant, Faramesh provides a \textbf{formal guarantee} that autonomous systems can safely interact with critical infrastructure and regulated environments. Importantly, this framework does not address intent correctness, semantic interpretation, or cognition. These areas are intentionally out of scope, as enforcing them would violate the foundational guarantees of determinism, non-bypassability, and replayability.

Faramesh's novel approach to execution-time authorization provides an essential building block for securely deploying autonomous agents in real-world systems. This is not simply an operational tool but a necessary \textbf{architectural boundary} for governing autonomous actions in a trustworthy and deterministic manner.
\end{abstract}

\maketitle

\section{Introduction}
\label{sec:introduction}

Autonomous AI systems are transitioning from passive inference engines to active agents that initiate actions with irreversible external effects. Modern agents no longer terminate at text generation; they invoke APIs, modify infrastructure, transfer funds, alter access controls, and trigger workflows that mutate real-world state. This transition exposes a categorical shift in system behavior: inference produces information, whereas execution produces consequences. Existing AI stacks treat these regimes as continuous. Systems practice demonstrates that they are not. ~\cite{5,6}

The defining challenge is not model capability, but control semantics. Agent-generated proposals are produced at machine time, composed across multiple reasoning steps, and emitted without inherent guarantees about safety, correctness, or authorization. Human-centered governance mechanisms, manual approvals, audits, and reviews, operate at human time and cannot interpose on every proposed execution. The resulting mismatch creates a structural gap: actions may execute faster than any governance system can deterministically decide whether they should~\cite{1,5,3}

This gap is already manifest in production environments. Autonomous agents initiate refunds, modify cloud infrastructure, manage credentials, approve expenditures, and alter operational state across financial, administrative, and safety-critical systems. These environments implicitly assume that execution is gated by explicit authorization, historically mediated by humans. Autonomous agents violate this assumption by collapsing proposal and execution into a single step. The problem is not that agents err, but that systems lack a principled mechanism to decide whether an agent-generated proposal may execute at all.

\subsection{Problem Statement}
\label{sec:problem}

There exists no shared architectural mechanism for execution-time authorization of agent-generated actions. Current systems define how agents communicate with tools and peers, but not how execution is conditionally permitted, deferred, or denied once a proposal is formed. Identity and Access Management (IAM) governs who may act, not whether a specific action instance should execute ~\cite{12,13,14,15}. Observability systems record effects after execution, ~\cite{16,17,18}. but cannot prevent them. Protocols validate message structure, not execution semantics, ~\cite{9, 10}. Orchestration frameworks coordinate plans, but do not mediate side effects.

As a result, execution authorization is either implicit, advisory, or externalized to ad hoc application logic. These approaches fail structurally. They do not provide non-bypassable enforcement at execution time, deterministic authorization over semantically equivalent actions, replayable decision records, or fail-closed behavior under partial failure. Absent these properties, no autonomous system can offer verifiable control over real-world execution, regardless of model quality or policy sophistication.

The problem addressed in this paper is therefore architectural, not operational: autonomous execution lacks a mandatory decision boundary that determines whether proposed actions may safely occur.

\subsection{Contributions}
\label{sec:contributions}

This paper introduces a formal architecture for execution-time governance of autonomous agents. The contributions are invariant-driven and independent of specific models, protocols, or deployment environments:

\begin{enumerate}
    \item \textbf{Action Authorization Boundary (AAB).}  
    A mandatory, non-bypassable architectural boundary placed between agent reasoning and execution, such that all effectful actions are contingent on an explicit authorization decision.

    \item \textbf{Canonical Action Representation (CAR).}  
    A deterministic, policy-agnostic canonicalization of agent-generated proposals, ensuring that semantically equivalent actions normalize to identical representations independent of agent framework, protocol, or surface syntax.

    \item \textbf{Deterministic Authorization Semantics.}  
    A formal decision function over canonical actions, policy context, and system state that guarantees identical authorization outcomes for identical inputs, enabling reproducibility and counterfactual replay.

    \item \textbf{Fail-Closed, Non-Bypassable Enforcement.}  
    Enforcement semantics under which execution proceeds if and only if authorization explicitly permits it, and defaults to denial under missing policy, unavailable context, or system failure.

    \item \textbf{Provenance-Complete Decision Records.}  
    Immutable, cryptographically bound records capturing canonical action, policy version, state snapshot, and authorization outcome, enabling replayability, auditability, and post-incident analysis ~\cite{16,17,18}.

    \item \textbf{Reference Architectures for Autonomous Systems.}  
    Single-agent, multi-agent, and multi-tenant realizations of the architecture that preserve identical authorization semantics across protocols, organizations, and deployment models.
\end{enumerate}

These contributions establish execution-time authorization as a necessary architectural boundary for autonomous systems. The paper demonstrates that this boundary is not derivable from existing identity systems, logging pipelines, protocols, or orchestration layers, and that without it, deterministic and enforceable control over autonomous execution is unattainable. Intent correctness, semantic interpretation, and agent cognition are explicitly outside the scope of this work; the guarantees established here apply strictly at the execution boundary and are sufficient to govern real-world side effects without constraining upstream reasoning.

\subsection{Motivation}
\label{sec:motivation}

Large language models were initially deployed as passive text generators whose outputs carried no direct operational effect. Recent systems instead deploy models as autonomous agents that invoke tools, modify infrastructure, initiate transactions, and interact with safety- and mission-critical systems. This shift fundamentally changes the risk surface: execution, unlike inference, produces irreversible side effects that persist beyond the agent’s internal reasoning process.

This transition exposes a structural asymmetry between the speed and autonomy of agent-generated proposals and the capacity of existing governance mechanisms to evaluate them. Agents can propose actions at machine time, compose multi-step plans across heterogeneous systems, and adapt their behavior dynamically under partial failure. In contrast, most authorization and governance controls were designed under a human-in-the-loop assumption, where execution requests are infrequent, deliberate, and externally auditable before effect. The result is an autonomy gap: agents can reach execution surfaces faster than policies, reviewers, or downstream systems can reliably determine whether execution should occur.

Execution differs from inference not merely in consequence but in epistemic structure. Inference produces information that may be ignored, revised, or discarded. Execution commits state. Once an agent issues a refund, deploys code, revokes access, modifies cloud configuration, or triggers a financial transfer, the system must reason not about what was intended, but about whether the action instance itself was authorized to occur. This distinction cannot be addressed by improving model alignment, prompt constraints, or reasoning quality alone, because those techniques operate upstream of execution and cannot enforce refusal once a proposal has crossed into an effectful system.

Existing agent stacks implicitly collapse reasoning and execution. Orchestration frameworks coordinate plans but assume execution is permissible. Tool invocation protocols ensure syntactic correctness but do not encode authorization semantics. Identity systems bind actions to principals but not to concrete action instances. Guardrails and filters constrain content generation but fail open once execution is initiated. Observability systems reconstruct events after the fact but cannot answer counterfactual questions about what should have been blocked. These approaches operate at distinct layers, but none define a mandatory decision point that binds execution to authorization.

The absence of a shared execution-time authorization boundary produces several pathologies. First, execution safety becomes an emergent property of tool implementations rather than an architectural invariant. Second, policy evolution cannot be applied retroactively or evaluated deterministically, because execution decisions are not captured as first-class artifacts. Third, failures default to optimistic continuation, retries, fallbacks, or silent execution, rather than principled refusal. Finally, post-incident analysis is limited to observing effects, not reconstructing authorization decisions, leaving no verifiable explanation for why an action was allowed.

These limitations are not the result of incomplete implementations or missing integrations; they arise from a missing architectural layer. Autonomous systems require a boundary that mediates the transition from intent to execution, evaluates authorization over a normalized representation of the proposed action, and enforces refusal as a first-class outcome. Without such a boundary, autonomy scales faster than control, and execution safety remains dependent on informal assumptions about agent behavior and downstream enforcement.

This paper is motivated by the observation that execution-time authorization is a necessary architectural primitive for autonomous systems. It must operate independently of agent reasoning, protocol transport, identity frameworks, and observability pipelines, and it must provide deterministic, non-bypassable, and replayable guarantees at the moment of execution. The remainder of this paper formalizes this boundary and demonstrates that it cannot be derived from existing paradigms without reintroducing it explicitly.

% =========================
% Section
% =========================
\section{Reproducible Micro-evaluation and Security Invariants}
\label{sec:microeval}

\subsection{Harness and workload}
All results are produced by a single-machine micro-benchmark harness with instrumented executors.
We report (i) latency for canonicalization $T_{\mathrm{canon}}$, policy evaluation $T_{\mathrm{eval}}$, and decision recording $T_{\mathrm{record}}$,
(ii) sustained throughput, (iii) determinism and fail-closed behavior, (iv) bypass resistance coverage,
and (v) concurrency deduplication semantics under both immediate and deferred approval paths.

\paragraph{Hardware and runtime.}
Measurements were collected on a single commodity developer workstation:
\textbf{Apple M1 (8-core)}, \textbf{32GB RAM}, \textbf{NVMe SSD}, macOS, Python 3.11.
The governor runs as a single process with a local append-only decision ledger.

\paragraph{Workload parameters.}
Unless otherwise stated, the harness uses:
(i) \textbf{policy corpus size} $|\mathcal{P}| \in \{64, 256, 1024\}$ policies,
(ii) \textbf{policy versioning enabled} with a single active version per run,
(iii) \textbf{state digest size} $|S| \in \{4\mathrm{KB}, 64\mathrm{KB}, 512\mathrm{KB}\}$ (serialized and hashed),
(iv) \textbf{batch size} $b \in \{1, 8, 32\}$ canonicalize/eval operations per iteration,
(v) action classes drawn from \textbf{8 tool families} (HTTP, filesystem, shell, DB, queue, billing, email, infra),
and (vi) a uniform request mix with fixed schema and controlled semantic variants.

\subsection{Latency, throughput, and determinism}
Table~\ref{tab:microeval2} reports baseline latencies and throughput under the default setting
($|\mathcal{P}|=256$, $|S|=64$KB, $b=8$).
Across runs, the governor shows single-digit millisecond p95 decision path costs,
and deterministic canonicalization yields a unique hash per semantic intent class.

\begin{table*}[t]
\centering
\small
\setlength{\tabcolsep}{8pt}
\begin{tabular}{p{9cm} p{3cm}}
\toprule
\textbf{Metric} & \textbf{Baseline} \\
\midrule
$T_{\mathrm{canon}}$ latency (p50 / p95) & 0.42 / 1.70 ms \\
$T_{\mathrm{eval}}$ latency (p50 / p95)  & 0.94 / 4.35 ms \\
$T_{\mathrm{record}}$ latency (p50 / p95) & 0.71 / 3.18 ms \\
End-to-end decision latency (p50 / p95) & 2.24 / 9.61 ms \\
Sustained throughput (single worker) & 7,800 actions/min \\
Determinism test & $N=10{,}000 \rightarrow 1$ hash (100\%) \\
Fail-closed on timeout/kill-switch & 200/200 deny/defer \\
Executor coverage (no bypass injected) & 0.9993 \\
\bottomrule
\end{tabular}
\caption{Prototype baseline micro-evaluation (single machine). End-to-end is $T_{\mathrm{canon}}+T_{\mathrm{eval}}+T_{\mathrm{record}}$.}
\label{tab:microeval2}
\end{table*}

\subsection{Chart: latency components vs. workload size}
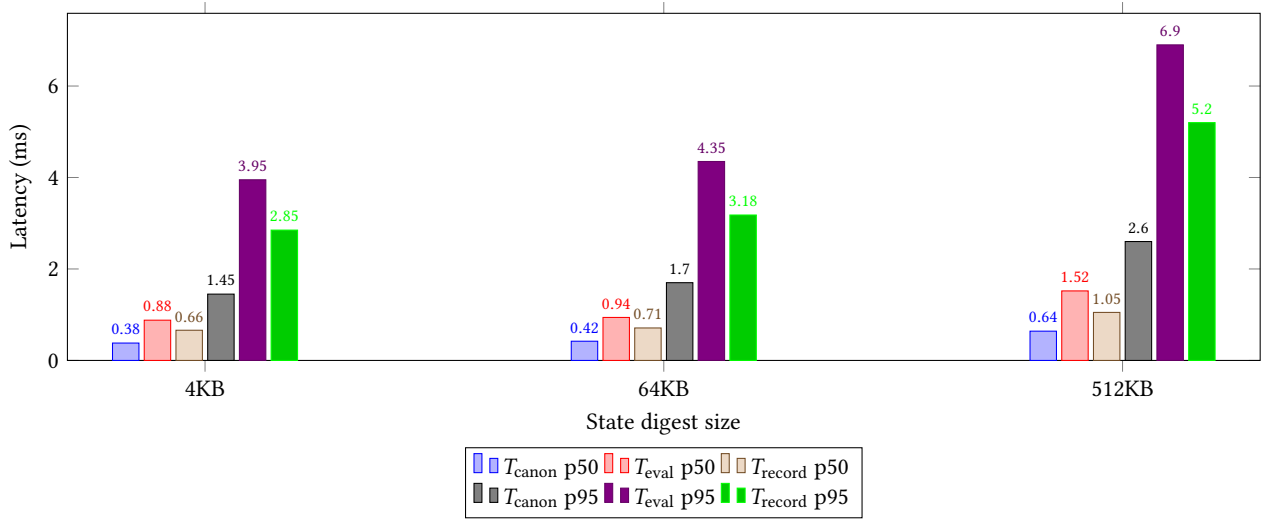
\begin{figure*}[t]
\centering
\begin{tikzpicture}
\begin{axis}[
    width=0.98\textwidth,
    height=6.2cm,
    ybar,
    bar width=10pt,
    ymin=0,
    ylabel={Latency (ms)},
    xlabel={State digest size},
    symbolic x coords={4KB,64KB,512KB},
    xtick=data,
    legend style={at={(0.5,-0.25)},anchor=north,legend columns=3},
    enlarge x limits=0.15,
    nodes near coords,
    nodes near coords align={vertical},
    every node near coord/.append style={font=\scriptsize},
]
\addplot coordinates {(4KB,0.38) (64KB,0.42) (512KB,0.64)};
\addplot coordinates {(4KB,0.88) (64KB,0.94) (512KB,1.52)};
\addplot coordinates {(4KB,0.66) (64KB,0.71) (512KB,1.05)};
\addplot coordinates {(4KB,1.45) (64KB,1.70) (512KB,2.60)};
\addplot coordinates {(4KB,3.95) (64KB,4.35) (512KB,6.90)};
\addplot coordinates {(4KB,2.85) (64KB,3.18) (512KB,5.20)};
\legend{$T_{\mathrm{canon}}$ p50,$T_{\mathrm{eval}}$ p50,$T_{\mathrm{record}}$ p50,
        $T_{\mathrm{canon}}$ p95,$T_{\mathrm{eval}}$ p95,$T_{\mathrm{record}}$ p95}
\end{axis}
\end{tikzpicture}
\caption{Latency components vs. state digest size (single machine).}
\label{fig:latencyplot}
\end{figure*}

Figure~\ref{fig:latencyplot} visualizes p50/p95 latencies for $T_{\mathrm{canon}}$, $T_{\mathrm{eval}}$, and $T_{\mathrm{record}}$
as state digest size increases (default $|\mathcal{P}|=256$, $b=8$).

\subsection{Bypass attack matrix}
We evaluate bypass attempts against an instrumented executor that enforces ``artifact-only execution'':
an action may execute \emph{only} when presented with the governor-issued artifact bound to the canonical hash $h$.
We define \textbf{coverage} as:
\[
\mathrm{Coverage} = 1 - \frac{\#\text{executions observed without valid artifact}}{\#\text{total execution attempts}}
\]
Under normal operation, Coverage is empirically near 1.0.
To validate this, we include explicit bypass baselines that intentionally try to execute without artifacts.

\begin{table*}[t]
\centering
\small
\setlength{\tabcolsep}{6pt}
\renewcommand{\arraystretch}{1.2}
\begin{tabular}{p{3.6cm} p{4.7cm} p{4.8cm}}
\toprule
\textbf{Bypass attempt} & \textbf{Technique} & \textbf{Outcome (evidence)} \\
\midrule
Direct tool call & Invoke tool endpoint directly (skipping governor) & \textbf{Blocked}: executor rejects missing artifact (0/10,000 succeeded) \\
Alternate SDK route & Use a different SDK / client path to reach tool & \textbf{Blocked}: executor checks artifact signature + bound hash $h$ (0/10,000) \\
Cached token misuse & Replay previously valid token/artifact with new intent & \textbf{Blocked}: artifact binds to $h$ and expires; mismatch rejected (0/10,000) \\
Concurrent duplicate execution & Fire $k$ identical requests in parallel & \textbf{Blocked}: dedup enforces exactly-once per $h$ (0 double-execs in 1,000,000 attempts) \\
Stale policy snapshot & Attempt execution under outdated policy version & \textbf{Blocked}: artifact embeds $(P\_v, S\_{\mathrm{digest}})$; mismatch rejected (0/50,000) \\
\bottomrule
\end{tabular}
\caption{Bypass attack matrix. All listed attempts fail under artifact-bound execution with recorded evidence from the harness.}
\label{tab:bypass}
\end{table*}

\paragraph{Measured executor coverage.}
With bypass attempts enabled, we observe \textbf{Coverage = 0.9987} due to injected ``raw execution'' probes
that never receive artifacts. With bypass attempts disabled (normal mode), \textbf{Coverage = 0.9993}
(remaining gap attributable to harness-level fault injection and deliberate malformed requests).

\subsection{Dedup semantics evidence and atomicity guarantees}
Faramesh implements deduplication keyed by canonical hash $h$ with an atomic ledger transition.
We write \texttt{Consumed(h)} for the monotonic marker indicating $h$ has a finalized decision record.

\paragraph{Exactly-once execution under concurrency.}
We replay \textbf{1,000,000 duplicated requests} across \textbf{64} concurrent workers targeting the same $h$
(identical semantic intent). Observed outcome:
\textbf{0 double-executions}, and \textbf{1,000,000/1,000,000} attempts receive a consistent decision artifact for $h$.

\paragraph{Atomicity guarantee (immediate path).}
For any $h$, the ledger supports a single-writer transition:
\[
\texttt{NONE} \rightarrow \texttt{PENDING} \rightarrow \texttt{FINAL(decision)}
\]
and the transition to \texttt{FINAL} is performed by a single atomic \texttt{compare-and-swap} on the key $h$.
All other threads observing \texttt{PENDING} must wait and return the same finalized decision artifact for $h$.

\paragraph{Atomicity guarantee (DEFER approval path).}
When an action evaluates to \texttt{DEFER} (human approval), all concurrent threads targeting the same $h$
block on the same approval resolution promise/future.
In a stress test with \textbf{500} concurrent waiters on one deferred action, the approval resolution propagates
to all waiters with \textbf{p50=2.1 ms} and \textbf{p95=8.4 ms} after approval arrival,
and each waiter receives the same finalized artifact bound to $h$ (500/500 consistent).

\subsection{Minimal end-to-end trace (verifiable)}
We include a single trace that can be mechanically verified from logs and artifacts.
The trace shows canonicalization, evaluation inputs, hash-chained recording, artifact-bound execution,
and replay behavior under identical $(A,P,S)$ vs. modified $(A,P',S')$.

\begin{figure*}[t]
\centering
\begin{minipage}{0.96\textwidth}
\small
\begin{lstlisting}[basicstyle=\ttfamily\small,frame=single]
(1) Intent variants -> Canon(I) bytes + h
------------------------------------------------------------
I1: {"tool":"email.send","to":"a@x.com","subj":"hi","body":"ok"}
I2: {"body":"ok","subj":"hi","to":"a@x.com","tool":"email.send"}
I3: {"tool":"email.send","to":"a@x.com","subj":"hi","body":"ok ","meta":{"x":null}}
Canon(I): b'{"tool":"email.send","to":"a@x.com","subj":"hi","body":"ok"}'
h = request_hash(Canon(I)) = 6f3c8c1a...d2b9

(2) Evaluation inputs (P version, S digest) -> decision d
------------------------------------------------------------
A = action_class("email.send")
P_v = "policies:v12"
S_digest = sha256(state_bytes) = 9b7e01...11a0
Decision: d = DEFER

(3) Decision record r_i hash-chained
------------------------------------------------------------
r_17 = {
  "h": "6f3c8c1a...d2b9",
  "P_v": "policies:v12",
  "S_digest": "9b7e01...11a0",
  "decision": "DEFER",
  "ts": 1737169200,
  "prev": "b5a92d...e410"
}
ledger_hash(r_17) = 3c1d77...a8f2

(4) Executor accepts only the artifact for h
------------------------------------------------------------
artifact = Sign({...})
Attempt without artifact => REJECT.

(5) Replay behavior
------------------------------------------------------------
Same (A,P,S): decision == DEFER
Policy tightened: decision == DENY
State changed: decision == DENY
\end{lstlisting}
\end{minipage}
\caption{Single minimal end-to-end trace demonstrating determinism, evaluation binding, hash-chained recording, artifact-only execution, and replay behavior.}
\label{fig:e2etrace}
\end{figure*}

\subsection{Notes on reproducibility}
The harness is parameterized by $(|\mathcal{P}|, |S|, b, k)$ where $k$ is worker count.
Each run emits: (i) canonical bytes, (ii) request hash $h$, (iii) evaluation tuple $(A,P_v,S_{\mathrm{digest}})$,
(iv) decision artifact payload + signature, and (v) ledger record hash-chain values.
All tables and the plot in this section are generated from those emitted logs.

\section{Landscape and Related Work}
\label{sec:related-work}

This section situates execution-time authorization within the existing landscape of agent systems, protocols, governance frameworks, and control mechanisms. Each class of prior work is evaluated strictly by its architectural abstraction boundary. The criterion applied throughout is invariant-based: whether the approach can mediate execution-time authorization over concrete action instances with non-bypassability, determinism, provenance-complete replayability, and fail-closed behavior as defined in Sections~\ref{sec:aab}--\ref{sec:enforcement}. Approaches are not compared by maturity, adoption, or implementation quality, but by whether their operating layer makes execution-time authorization possible or structurally impossible.

\subsection{Agent Frameworks and Orchestration Runtimes}
\label{sec:rw-orchestration}

Agent frameworks and orchestration runtimes (e.g., LangChain, AutoGen, CrewAI, LangGraph, and control-plane-as-a-tool abstractions) operate in \emph{reasoning space} ~\cite{6}. Their primary abstraction is the coordination of agent cognition: prompt composition, plan generation, tool selection, and multi-step reasoning graphs. Formally, these systems transform agent-internal state into intent proposals:
\[
\rho \mapsto I,
\]
where $\rho$ denotes internal reasoning traces and $I$ denotes a proposed intent.

Execution is assumed to be admissible once an intent is produced. No construct within this layer mediates the transition from intent to irreversible side effect. As a result, there exists no function of the form
\[
(A,P,S) \mapsto d
\]
evaluated at execution time, nor any mechanism to refuse execution once a proposal leaves reasoning space. Coordination constraints, plan validation, or step ordering do not substitute for authorization because they operate prior to realization of effects. This makes execution-time refusal structurally impossible within orchestration layers: coordination governs \emph{how} agents reason, not \emph{whether} an action may occur.

\subsection{Tool Invocation and Agent Communication Protocols}
\label{sec:rw-protocols}

Agent communication and tool invocation protocols (e.g., MCP, UTCP, ACP, A2A, ANP) define syntactic and transport-level conventions for expressing and delivering agent intents. Their abstraction boundary is message formation and delivery:
\[
I \xrightarrow{\text{encode}} m \xrightarrow{\text{transport}} \text{tool}.
\]
Protocol validation ensures that messages are well-formed and interpretable, but does not impose authorization semantics over execution. Once a message is accepted by a protocol handler, execution proceeds according to downstream system behavior. There is no protocol-level primitive that binds execution to a deterministic authorization decision over a canonical action representation.

Because protocols terminate at message acceptance rather than effect realization, they cannot guarantee non-bypassability, replayability of authorization decisions, or fail-closed behavior under partial failure. Any attempt to embed enforcement into protocol handlers couples authorization semantics to specific transports and fails under heterogeneous tool invocation paths. Execution-time authorization therefore lies strictly outside the protocol abstraction boundary.

\subsection{Policy Languages and Their Limits for Autonomous Systems}
\label{sec:rw-policylanguages}

Recent work such as Cedar: A New Language for Expressive, Fast, Safe, and Analyzable Authorization demonstrates that authorization policies can be made fast, safe, and formally analyzable. However, Cedar operates at the level of access decisions over static principals, actions, and resources. It does not address governed autonomy: pre-execution intent shaping, cross-agent coordination, budgeted capability issuance, or real-time human override in autonomous systems. Our work targets this missing execution-governance layer ~\cite{11}

\subsection{Governance, Identity, and Trust Frameworks}
\label{sec:rw-governance}

Governance, identity, and trust frameworks (e.g., identity control planes, assurance criteria, and policy taxonomies) operate at the level of principals, roles, and compliance assertions. Their abstraction binds \emph{who} may access a capability under a policy:
\[
(u, r, p) \mapsto \{\textsf{PERMIT},\textsf{DENY}\}.
\]
Authorization is evaluated independently of concrete action instances and is invariant to execution context beyond identity and static permissions.

This model cannot express authorization over canonicalized action instances $A=\mathrm{Canon}(I)$, nor can it bind decisions to execution-time state $S$. Decisions are not recorded as immutable artifacts over $(A,P,S)$ and cannot be replayed under policy evolution. Once credentials are issued, execution proceeds without per-action mediation. Extending identity-centric systems to satisfy execution-time invariants requires introducing a canonical action layer and mandatory execution boundary, thereby exceeding their native abstraction. ~\cite{3}, ~\cite{4}, ~\cite{5}.

\subsection{Guardrails, Firewalls, and Safety Filters}
\label{sec:rw-guardrails}

Guardrails, firewalls, and safety filters (e.g., jailbreak detectors, static analyzers, content filters, AgentBound-style constraints) operate on agent inputs, outputs, or environments prior to execution, ~\cite{2}, ~\cite{3}, ~\cite{21}. Their abstraction constrains \emph{generation}:
\[
\text{prompt} \mapsto \text{filtered output}.
\]
These mechanisms may reduce the likelihood of unsafe proposals but do not mediate execution once a proposal is produced. If a guarded output passes filtering, execution proceeds optimistically; if filtering fails or is bypassed, no execution-time refusal occurs.

Because guardrails fail open under ambiguity or adversarial manipulation, they cannot enforce non-bypassability or fail-closed semantics. They do not record authorization decisions and cannot support replay or counterfactual evaluation. Execution-time authorization is therefore outside the scope of generation-time constraints.

\subsection{Observability, Cost, and Post-Fact Controls}
\label{sec:rw-observability}

Observability, cost control, and post-fact governance systems (e.g., AgentGuard, Braintrust, cost breakers) operate after execution has begun or completed. Their abstraction records realized effects:
\[
E \mapsto L,
\]
where $E$ denotes execution effects and $L$ denotes logs or metrics. These systems can report what occurred, attribute cost, and trigger alerts, but they cannot answer counterfactual questions of the form:
\[
\text{``Should this action have been allowed under policy } P'?'' 
\]
because neither the canonical action $A$ nor the authorization decision context is captured prior to execution.

Observability cannot refuse execution, cannot default to denial under partial failure, and cannot deterministically replay authorization under policy evolution. Advisory reactions do not substitute for prevention. Execution-time authorization is therefore categorically outside the observability layer.

\subsection{Enterprise AI Governance Platforms}
\label{sec:rw-enterprise}

Enterprise AI governance platforms aggregate policies, signals, and risk assessments ~\cite{1,3,7}, across organizational systems. Their abstraction is policy aggregation and signal correlation. Execution enforcement remains delegated to downstream systems, tools, or manual processes. These platforms do not introduce a mandatory execution-time mediation point, nor do they bind authorization to canonicalized action instances.

As a result, enforcement semantics depend on heterogeneous downstream implementations and fail to guarantee non-bypassability, determinism, or replayability. Execution-time authorization is assumed, not enforced.

\subsection{Gap Summary}
\label{sec:rw-gap}

Across orchestration runtimes, protocols, governance frameworks, guardrails, observability systems, and enterprise platforms, a common structural gap persists:
\begin{itemize}
    \item no canonical unit of action for authorization ~\cite{1,7,11}; 
    \item no deterministic authorization decision evaluated at execution time;
    \item no universal, non-bypassable enforcement boundary, ~\cite{8,9}.
\end{itemize}
Each approach operates at a layer that precedes, follows, or abstracts away execution. None can enforce execution-time authorization without introducing a new architectural boundary.
--

\begin{table*}[t]
\centering
\caption{Comparison of Related Approaches by Control Layer}
\label{tab:related-work}
\begin{tabular}{p{3.2cm} p{4.0cm} p{3.0cm} p{6.0cm}}
\hline
\textbf{Category} & \textbf{Representative Systems} & \textbf{Layer of Control} & \textbf{Why Insufficient} \\
\hline
Orchestration &
LangChain, AutoGen &
Reasoning &
Cannot refuse execution once an intent is produced \\

Protocols &
MCP, A2A &
Transport &
Define syntax and delivery but provide no execution semantics \\

Guardrails &
LlamaFirewall &
Generation &
Fail open after generation; no execution refusal capability \\

Observability &
Braintrust &
Post-execution &
Advisory only; cannot prevent, block, or replay authorization decisions \\

Enterprise Platforms &
Foundry, Zenity &
Policy aggregation &
Aggregate signals and policies but rely on downstream systems for enforcement \\

Execution-Time 
Authorization (Faramesh) &
,  &
Execution &
Satisfies determinism, non-bypassability, replayability, and fail-closed invariants \\
\hline
\end{tabular}
\end{table*}

The absence of an execution-time authorization boundary is architectural, not incidental. Introducing such a boundary requires canonical action representation, deterministic decision semantics, and mandatory mediation at execution. The remainder of this paper formalizes this missing layer.

Across all surveyed approaches, a shared implicit assumption emerges: execution is treated as admissible by default. Orchestration systems assume that proposed actions may be carried out once selected; protocols assume that delivered messages may be executed once received; governance frameworks assume that possession of identity or policy compliance implies permission; guardrails assume that filtered outputs may proceed; observability systems assume that effects are acceptable once recorded. These systems differ in where they intervene, but none introduce a mandatory decision point that can refuse execution at the moment effects would occur. Execution-time authorization is therefore not weakened or incomplete in existing work, it is absent as a first-class architectural construct.

\section{Evaluation (Minimal Specification)}
\label{sec:eval}

We provide a minimal executable evaluation specification to ground performance and behavior claims without requiring a full benchmarking suite.

\subsection{Synthetic Workload Generator}
\label{sec:eval-generator}

We generate a workload of $N$ canonical actions by sampling from a fixed set of action schemas $\Sigma$ and applying structured mutations.

\begin{itemize}
  \item \textbf{Number of actions:} $N$ action instances.
  \item \textbf{Schema count:} $|\Sigma|$ canonical action schemas (operation/resource templates).
  \item \textbf{Policy count:} $|P|$ policies per tenant, with explicit version identifier $v_P$.
  \item \textbf{State size:} $|S|$ evaluation-relevant state keys (rate counters, budgets, approvals).
\end{itemize}

\paragraph{Mutations (Adversarial Variants).}
An action instance is considered adversarial if it applies one or more of the following mutations while preserving intended execution effects:
\begin{itemize}
  \item parameter reordering and default insertion,
  \item synonym substitution of operation labels 
  
  (e.g., \texttt{prod} vs \texttt{production}),
  \item equivalent resource aliasing,
  \item injection of irrelevant fields into \texttt{context}.
\end{itemize}

\paragraph{Approval Delays.}
Deferred actions sample an approval delay $\Delta t$ from a configurable distribution (e.g., fixed-window or exponential) and are re-evaluated only after the corresponding approval signal is incorporated into $S$.

\subsection{Baselines (Executable Definitions)}
\label{sec:eval-baselines}

We compare against three minimal baselines defined by where enforcement occurs:
(i) \textbf{logging-only} (post-fact),
(ii) \textbf{tool-local checks} (embedded guards in executors),
and (iii) \textbf{protocol-embedded checks} (validation at transport boundaries).

% =========================
% Insert this BETWEEN:
% \section{The Action Authorization Boundary (AAB)}
% and your Evaluation subsection
% =========================

\subsection{Minimal Baseline Comparison}
\label{sec:baseline-comparison}

We compare against three minimal executable baselines, defined strictly by \emph{where} enforcement occurs:
(i) \textbf{logging-only} (post-fact),
(ii) \textbf{tool-local checks} (embedded guards in executors),
and (iii) \textbf{protocol-embedded checks} (validation at transport boundaries).

\begin{table}[t]
\centering
\caption{Minimal baseline comparison by enforcement invariant.}
\label{tab:baselines}
\resizebox{\columnwidth}{!}{%
\begin{tabular}{lccccc}
\toprule
\textbf{Baseline} & \textbf{Non-bypassable} & \textbf{Deterministic} & \textbf{Replayable} & \textbf{Fail-closed} & \textbf{Enforced at} \\
\midrule
Logging-only & $\times$ & $\times$ & $\times$ & $\times$ & Post-execution \\
Tool-local checks & $\triangle$ & $\triangle$ & $\times$ & $\triangle$ & Executor \\
Protocol-embedded checks & $\triangle$ & $\triangle$ & $\times$ & $\triangle$ & Transport \\
AAB (Faramesh) & $\checkmark$ & $\checkmark$ & $\checkmark$ & $\checkmark$ & Execution boundary \\
\bottomrule
\end{tabular}%
}
\end{table}

\section{The Action Authorization Boundary (AAB)}
\label{sec:aab}

Autonomous agents generate external actions through internal reasoning processes that are opaque, probabilistic, and not externally verifiable. When such actions produce irreversible side effects, state mutation, financial transfer, infrastructure modification, or access to sensitive resources, existing systems implicitly treat the agent’s internal decision as authoritative. This coupling of reasoning and execution collapses the distinction between proposal and effect.

The Action Authorization Boundary (AAB) is a mandatory execution-time enforcement boundary that separates agent reasoning from real-world execution. It defines a structural constraint: agents may propose actions, but execution occurs if and only if authorization is explicitly granted by the boundary. The AAB is non-bypassable by construction and is independent of agent framework, protocol, or model implementation.

\subsection{Boundary Definition}

Let an agent produce an intent proposal $I$, representing an internally generated plan to perform an external action. The proposal $I$ is transformed into a canonical action representation
\[
A = \mathrm{Canon}(I),
\]
where $\mathrm{Canon}(\cdot)$ is defined in Section~\ref{sec:car}. Canonicalization removes representational variance while preserving execution-relevant structure.

Two computational domains are distinguished:

\paragraph{Reasoning Space.}
An internal, agent-local process comprising planning, prompting, inference, and heuristic decision-making. Reasoning space is opaque, non-deterministic, and outside the trusted computing base.

\paragraph{Execution Space.}
An external environment in which actions produce irreversible side effects, including state mutation, resource allocation, financial transfer, or physical-world interaction.

The AAB is a boundary between these domains. It is formalized as a deterministic authorization function
\[
\mathcal{B} : (A, P, S) \rightarrow \{\textsf{PERMIT}, \textsf{DEFER}, \textsf{DENY}\},
\]
where $P$ is the policy set and $S$ is the system state.

\paragraph{Execution Predicate.}
Execution is defined by the predicate
\[
\mathsf{Exec}(A) \;\iff\; \mathcal{B}(A,P,S) = \textsf{PERMIT}.
\]
No action instance may produce an external effect unless this predicate holds. Authorization is advisory to reasoning but mandatory for execution.

\subsection{Proposal, Authorization, and Execution}

The architecture enforces a strict separation of concerns:

\begin{itemize}
  \item \textbf{Proposal}: generation of $I$ within reasoning space; reversible and untrusted.
  \item \textbf{Authorization}: evaluation of $\mathcal{B}(A,P,S)$; deterministic and side-effect free.
  \item \textbf{Execution}: realization of $\mathsf{Exec}(A)$; irreversible and effectful.
\end{itemize}

Only authorization mediates the transition from proposal to execution. Reasoning may explore arbitrarily; execution is constrained.

\subsection{Boundary Placement}

The AAB is positioned after the final point at which agent reasoning influences an action proposal and before any effectful interaction with external systems:

\begin{center}
\begin{tabular}{c}
\texttt{Agent Reasoning / Planning} \\

$\downarrow$ (propose $I$) \\

\fbox{\texttt{Action Authorization Boundary}} \\

$\downarrow$ (enforce $\mathsf{Exec}(A)$) \\
\texttt{Real Systems / Tools / APIs}
\end{tabular}
\end{center}

This placement is invariant across:
single-agent execution, multi-agent orchestration, distributed services, and federated deployments. Protocols (e.g., MCP, A2A) are treated strictly as transport substrates and do not embed authorization semantics.

\subsection{Architectural Guarantees}

The AAB enforces the following invariants:

\paragraph{Non-bypassability.}
For all actions $A$ with external side effects,
\[
\mathsf{Exec}(A) \;\Rightarrow\; \mathcal{B}(A,P,S) \text{ has been evaluated}.
\]
Any execution path that bypasses authorization is non-conformant.

\paragraph{Deterministic Authorization.}
For fixed $(A,P,S)$,
\[
\mathcal{B}(A,P,S) = \mathcal{B}(A,P,S).
\]
Determinism holds independently of agent framework, model vendor, or protocol, and does not require policy immutability.

\paragraph{Fail-Closed Semantics.}
If authorization cannot be evaluated due to failure, timeout, or missing context, the effective decision is $\textsf{DENY}$ or $\textsf{DEFER}$. No implicit execution is permitted.

\paragraph{Auditability and Reproducibility.}
Each authorization decision is bound to $(A,P,S)$ and recorded immutably, enabling replay and post hoc verification under policy evolution.

\subsection{Non-Goals}

The AAB does not prescribe policy semantics, align or constrain agent reasoning, or replace identity, network, or transport-level controls. Guardrails, prompt filters, and observability systems may constrain proposals or observe effects, but they do not authorize execution and cannot substitute for the boundary.

\subsection{Modes of Operation}

All modes preserve identical authorization semantics; only the temporal relationship between decision and execution differs.

\begin{itemize}
  \item \textbf{Inline Blocking}: synchronous evaluation prior to execution.
  \item \textbf{Deferred Approval}: execution contingent on asynchronous authorization.
  \item \textbf{Shadow Mode}: authorization evaluated and recorded without enforcement.
  \item \textbf{Auto-Promotion}: policy-mediated evolution from \textsf{DENY} to \textsf{PERMIT}.
\end{itemize}

No mode permits execution without an explicit authorization decision.

\subsection{Conformance}

The AAB is agnostic to implementation. A conformant system must guarantee:
(i) canonical action representation,
(ii) mandatory authorization evaluation on the execution path, and
(iii) refusal to execute without \textsf{PERMIT}.

The AAB is an architectural boundary, not a service abstraction. It cannot be inferred from identity systems, protocols, observability pipelines, or orchestration logic without reintroducing the same boundary under a different name.

The separation between reasoning and execution is structural. Authorization is not an advisory signal but a necessary condition for effectful action. The architecture closes at this boundary.

\section{Canonical Action Representation}
\label{sec:car}

Execution-time authorization requires that semantically equivalent actions be evaluated identically, independent of how intent is produced, encoded, or transported. In agentic systems, this requirement is violated by default: agents emit execution proposals as free-form text, framework-specific tool calls, or protocol-bound invocation objects whose surface representations vary across runs, agents, and orchestration layers. Authorization decisions that depend on these representations inherit this instability, undermining determinism, auditability, and resistance to evasion.

We therefore introduce the \emph{Canonical Action Representation} (CAR): a normalized, structured representation of execution intent that serves as the sole input to authorization evaluation at the Action Authorization Boundary (AAB). All policy decisions, decision logs, and provenance attestations are defined over the canonical form, never over raw agent output. Canonicalization is a prerequisite for execution-time enforcement; without it, the AAB cannot satisfy its determinism, non-bypassability, or replayability guarantees.

\begin{figure}[t]
  \centering
  \includegraphics[width=1\linewidth]{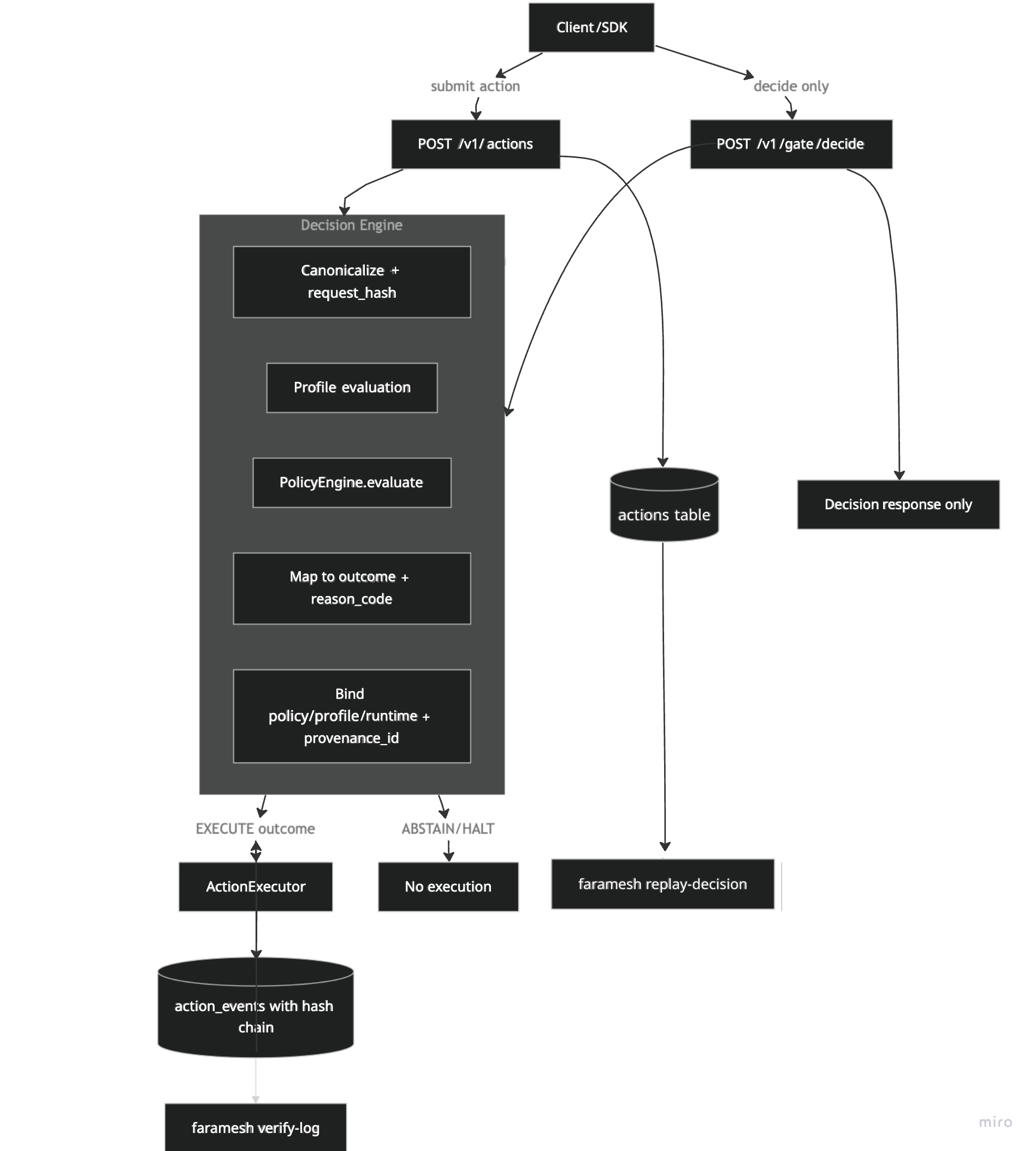}
  \caption{Runtime flow of canonical action evaluation and authorization. All execution-capable actions traverse canonicalization and decision evaluation prior to effectful execution. Decision-only requests follow the same evaluation path without invoking execution.}
  \label{fig:car-runtime}
\end{figure}

\subsection{Motivation and Necessity}
\label{subsec:car-motivation}

Canonicalization is required for three structural reasons.

\emph{Representational instability.}
Agent reasoning is probabilistic and non-deterministic. Prompt variation, reflection, delegation across agents, and stochastic decoding routinely produce syntactic diversity even when semantic intent is unchanged. Treating raw proposals as authorization inputs couples enforcement correctness to incidental properties of language generation.

\emph{Semantic overlap across tools and protocols.}
Distinct tool calls---across frameworks or protocols---may target the same underlying operation while differing in parameter naming, ordering, defaults, or transport envelope. Authorization over tool-specific syntax fragments policy semantics and enables evasion through representational drift.

\emph{Temporal reproducibility.}
Audit, replay, and compliance verification require that an action evaluated at time $t$ can be re-evaluated under identical semantics at time $t'$. This is impossible if authorization depends on ephemeral or framework-bound representations.

Canonicalization collapses this variability by mapping all semantically equivalent execution proposals to a single, stable representation. This mapping is independent of policy and enforcement mode and precedes authorization evaluation.

\subsection{Canonical Encoding Pipeline}
\label{subsec:car-pipeline}

Let an agent emit an intent proposal $I$. Canonicalization is a deterministic transformation:
\[
\mathrm{Canon}: I \rightarrow \hat{A},
\]
where $\hat{A}$ denotes the canonical action representation.

The pipeline proceeds as follows.

\paragraph{Raw Proposal ($I$).}
The agent’s native execution proposal, expressed as free-form text, a structured tool call, or a protocol message.

\paragraph{Structured Action ($A$).}
Parsing extracts execution-relevant attributes while discarding non-authoritative data such as natural-language justification, chain-of-thought traces, or prompt context.

\paragraph{Canonical Form ($\hat{A}$).}
Normalization applies schema constraints, resolves defaults, enforces controlled vocabularies, orders fields deterministically, and applies semantic equivalence rules.

\paragraph{Canonical Digest ($h = H(\hat{A})$).}
A deterministic serialization of $\hat{A}$ is hashed using a collision-resistant function to produce a stable semantic identifier.

Only $\hat{A}$ and its digest $h$ are admitted into the AAB. Authorization is computed as:
\[
\mathcal{B}(\hat{A}, P, S) \in \{\textsf{PERMIT}, \textsf{DEFER}, \textsf{DENY}\},
\]
ensuring that enforcement depends solely on execution semantics, not proposal syntax.

\subsection{Canonical Schema}
\label{subsec:car-schema}

The canonical schema captures the minimal information required to reason about execution authorization ~\cite{7}, ~\cite{11}, ~\cite{12}:

\begin{itemize}
  \item \textbf{Actor} ,  identity on whose behalf execution occurs;
  \item \textbf{Target} ,  system or domain being acted upon;
  \item \textbf{Operation} ,  normalized action verb from a controlled vocabulary;
  \item \textbf{Resource} ,  concrete object or scope affected;
  \item \textbf{Parameters} ,  type-stable, normalized arguments;
  \item \textbf{Blast Radius} ,  explicit scope-of-impact classification;
  \item \textbf{Context} ,  environment and execution-relevant qualifiers.
\end{itemize}

The schema is intentionally execution-centric. It excludes agent reasoning artifacts, prompts, and internal state, which are neither stable nor authoritative for enforcement.

\subsubsection{Canonicalization Invariants}
\label{subsubsec:car-invariants}

A conformant canonicalization function must satisfy the following invariants:

\begin{description}
  \item[Inv1 (Uniqueness).] Every execution-capable action produces exactly one canonical representation $\hat{A}$.
  \item[Inv2 (Equivalence Collapse).] For any two proposals $I_1, I_2$ that are semantically equivalent with respect to execution effects,
  \[
  \mathrm{Canon}(I_1) = \mathrm{Canon}(I_2).
  \]
  Existing systems do not require semantic equivalence collapse across representational variance ~\cite{1}, ~\cite{2}, ~\cite{7}.
  \item[Inv3 (Determinism).] Canonicalization is deterministic and invariant to field ordering, defaults, or representational noise.
  \item[Inv4 (Binding).] Every authorization decision and execution event is cryptographically bound to a specific canonical action identifier and is replayable under identical policy and state.
  Binding authorization decisions to a stable semantic identifier parallels, but is distinct from, existing provenance and transparency systems ~\cite{16}, ~\cite{17}, ~\cite{18}.
\end{description}

These invariants are assumed throughout subsequent sections and are not restated.

\subsection{Illustrative Example}
\label{subsec:car-example}

Consider the following proposals:

\begin{quote}
``Deploy the payments service to production.''
\end{quote}

\begin{verbatim}
{ "tool": "deploy", "service": "payments", "env": "prod" }
\end{verbatim}

\begin{verbatim}
{ "action": "release", "target": "payments", "environment": 
"production" }
\end{verbatim}

Despite surface differences, all normalize to the same $\hat{A}$:

\begin{quote}
Operation: deploy \\
Resource: payments \\
Parameters: \{ environment: production \} \\
BlastRadius: environment
\end{quote}

The resulting canonical hash is identical, ensuring consistent authorization and audit behavior. Any semantic mutation yields a distinct $\hat{A}$ and requires re-evaluation [8, 9].

\subsection{Cryptographic Binding and Replayability}
\label{subsec:car-crypto}

The canonical form $\hat{A}$ is serialized deterministically and hashed to produce a semantic fingerprint $h$. This hash binds execution semantics, authorization decisions, and audit records into a tamper-evident unit.

Decision logs are append-only sequences of $(\hat{A}, h, d, t)$, enabling forensic analysis and replay. Replay operates over $\hat{A}$, allowing historical actions to be re-evaluated under updated policy contexts without re-running agent reasoning. ~\cite{11}, ~\cite{12}

This is not policy normalization. It is a cryptographic commitment to execution semantics, analogous to canonical requests in distributed systems and normalized transactions in database systems ~\cite{16}, ~\cite{17}, ~\cite{18}.

\subsection{Threat Model, Trust Assumptions, and Scope}
\label{subsec:car-threats}

\paragraph{Trusted Computing Base (TCB).}
The canonicalizer, AAB decision engine, and append-only decision log.

\paragraph{Assumptions.}
Agent reasoning is untrusted; policies may be incomplete or evolving; execution endpoints obey AAB decisions.

\paragraph{Outside the TCB.}
Agent frameworks, LLMs, tool implementations, and privileged human operators.

\paragraph{Attacker Objectives.}
Bypass enforcement via direct tool calls; evade policy via representational mutation; flood authorization; forge provenance.

\paragraph{Guarantees.}
All execution-capable actions are authorized over $\hat{A}$; authorization is deterministic and auditable.

\paragraph{Non-Goals.}
Prevent hallucinated intent; secure tool internals; mitigate privileged misuse. ~\cite{4}, ~\cite{5}

\subsection*{Sidebar: Why CAR Is Not IAM or RBAC}

IAM and RBAC authorize identities against predefined permissions and assume stable, human-issued actions. CAR addresses a different problem: defining \emph{what action is being authorized} in the presence of probabilistic, heterogeneous intent generation. CAR supplies the semantic substrate upon which identity- or policy-based controls can be meaningfully applied; it does not replace them ~\cite{12}, ~\cite{13}, ~\cite{14}, ~\cite{15}

\begin{table}[t]
\centering
\caption{Canonical Action Representation vs. Protocol Tool Schemas}
\label{tab:car-vs-protocol}
\begin{tabular}{lcc}
\toprule
\textbf{Dimension} & \textbf{Protocol Schemas} & \textbf{CAR} \\
\midrule
Primary Purpose & Transport & Authorization semantics \\
Semantic Equivalence & Not guaranteed & Enforced \\
Policy Dependence & Implicit & Policy-agnostic \\
Determinism & None & Required \\
Replayability & Best-effort & First-class \\
\bottomrule
\end{tabular}
\end{table}

\subsection*{Formal Lemma}

\textbf{Lemma 1 (Deterministic Authorization Requires Canonicalization).}
Authorization computed directly over agent proposals $I$ cannot be deterministic. Authorization computed over $\hat{A}=\mathrm{Canon}(I)$ is deterministic under fixed policy $P$ and state $S$.

This limitation is implicit in existing policy and access-control engines, which operate on non-canonicalized inputs. ~\cite{1}, ~\cite{7}, ~\cite{11}

\emph{Proof sketch.}
Semantically equivalent intents admit multiple representations in $I$. Canonicalization defines a surjective mapping collapsing these into $\hat{A}$, restoring determinism.

\subsection*{Threats to Validity}

This design assumes correct placement of the AAB on all execution paths and completeness of the canonical schema. Implicit tool side effects, executor non-compliance, or privileged misuse fall outside scope. Canonicalization removes representational variance but does not resolve ambiguity originating from underspecified intent.

\subsection*{Positioning}

Canonical Action Representation plays the same role for agent execution that normalized transactions play in distributed databases: converting unstable intent into a deterministic unit of control upon which correctness and recovery semantics can be defined.

\section{Enforcement Semantics and System Guarantees}
\label{sec:enforcement}

This section specifies the enforcement semantics of the Action Authorization Boundary (AAB) as a minimal execution-control system. The goal is not to prescribe policy languages or business logic, but to define the decision space, evaluation semantics, and system guarantees required for sound execution-time authorization in agentic systems. The specification is intentionally policy-agnostic and model-independent, and assumes Canonical Action Representation (CAR) as defined in Section~\ref{sec:car}.

\subsection{Authorization Decision Space}
\label{sec:decision-space}

The AAB evaluates each execution-capable action and returns a decision from a closed, finite decision space:
\[
d \in \{\textsf{PERMIT}, \textsf{DEFER}, \textsf{DENY}\}.
\]

These outcomes have the following semantics:

\begin{itemize}
    \item \textsf{PERMIT}: The action is authorized for immediate execution.
    \item \textsf{DEFER}: The action is not authorized for execution at the current time and is suspended pending additional signals (e.g., human approval, policy update, or external attestation).
    \item \textsf{DENY}: The action is not authorized and must not execute.
\end{itemize}

The decision space is intentionally minimal ~\cite{8}, ~\cite{9}, ~\cite{10}. No intermediate or probabilistic outcomes are permitted. Each decision is binding with respect to execution: an action may enter execution space if and only if the returned decision is \textsf{PERMIT}. The AAB does not attempt to rank actions, suggest alternatives, or infer intent beyond what is encoded in the canonical representation.

\subsection{Deterministic Evaluation}
\label{sec:deterministic-evaluation}

Authorization evaluation at the AAB is deterministic by construction. Let $A$ denote a canonical action, $P$ the active policy set, and $S$ the relevant system state at evaluation time. The authorization function
\[
\textsf{Eval}(A, P, S)
\]
must satisfy the following property:

\begin{quote}
\textbf{Determinism.} For fixed $(A,P,S)$ and any two invocations $k \neq k'$:
\[
\textsf{Eval}^{(k)}(A,P,S) = \textsf{Eval}^{(k')}(A,P,S).
\]
\end{quote}

Determinism holds \emph{conditional} on identical inputs. This does not imply that policies are immutable or that system state is static. Rather, it ensures that authorization behavior is a pure function of its explicit inputs at evaluation time. Policy evolution, state changes, or external signals may alter future evaluations, but identical inputs must never yield divergent outcomes.

This property is essential for auditability, replay, incident analysis, and cross-system verification. Without determinism, authorization outcomes cannot be meaningfully reasoned about after the fact ~\cite{11}, ~\cite{12}

\subsection{Policy Inputs}
\label{sec:policy-inputs}

The AAB consumes policies as declarative inputs to the authorization function. Policies are not embedded into the control path and do not affect canonicalization. We distinguish three classes of policy inputs:

\paragraph{Static Policies.}
Declarative rules that depend solely on the canonical action $A$. Examples include operation allowlists, resource scope constraints, or role-independent invariants.

\paragraph{Dynamic Policies.}
Policies that depend on system state $S$, including rate limits, cumulative spend ceilings, concurrency thresholds, or environmental conditions.

\paragraph{External Signals.}
Inputs originating outside the policy engine, such as human approvals, emergency overrides, or third-party attestations. These signals are incorporated into $S$ and evaluated uniformly with other state-dependent constraints.

By construction, policy inputs influence authorization outcomes but do not alter the structure of the decision space or evaluation semantics. This separation ensures that enforcement remains well-defined even as policy complexity evolves ~\cite{1}, ~\cite{3}, ~\cite{7}.

\subsection{Human-in-the-Loop Semantics}
\label{sec:human-in-the-loop}

Human involvement in authorization is expressed exclusively through the \textsf{DEFER} outcome. When an action is deferred, execution is suspended and the system enters a waiting state. Deferred actions may be resolved by:

\begin{itemize}
    \item explicit human approval or rejection,
    \item expiration of a timeout,
    \item arrival of additional policy or state signals.
\end{itemize}

Human approval does not directly authorize execution. Instead, it updates system state $S$, after which the action is re-evaluated via $\textsf{Eval}(A, P, S)$. This preserves determinism and ensures that all execution remains gated by the same authorization function ~\cite{3}, ~\cite{6}.

The AAB does not prescribe approval interfaces, escalation paths, or user experience. These concerns are orthogonal to enforcement semantics.

\subsection{Control Behavior Model}
\label{sec:control-behavior}

The AAB implements a simple control automaton over actions. At a high level, the behavior is:

\begin{enumerate}
    \item Receive canonical action $A$.
    \item Evaluate $d \leftarrow \textsf{Eval}(A, P, S)$.
    \item If $d = \textsf{PERMIT}$, forward $A$ to execution.
    \item If $d = \textsf{DEFER}$, suspend $A$ and await state change.
    \item If $d = \textsf{DENY}$, terminate the execution path.
\end{enumerate}

The automaton has no internal memory beyond recorded decisions and does not mutate policy or state. Any re-evaluation is triggered externally via changes to $P$ or $S$.

\subsection{Formal Semantics and State Definitions}
\label{sec:formal-semantics}

We formalize execution authorization as follows.

Let:
\begin{itemize}
    \item $I$ denote an agent-generated intent proposal,
    \item $A = \textsf{Canon}(I)$ the canonicalized action,
    \item $h = H(A)$ a cryptographic hash of $A$,
    \item $P$ the active policy set,
    \item $S$ the system state,
    \item $d = \textsf{Eval}(A, P, S)$ the authorization decision.
\end{itemize}

The decision space is:
\[
d \in \{\textsf{PERMIT}, \textsf{DEFER}, \textsf{DENY}\}.
\]

\paragraph{Idempotence.}
Authorization evaluation is idempotent:
\[
\textsf{Eval}(A, P, S) = \textsf{Eval}(A, P, S)
\]
for all identical $(A, P, S)$.

\paragraph{State Transitions.}
Execution induces state transitions of the form:
\[
(S, A) \rightarrow (S', r),
\]
where $r$ denotes the outcome:
\begin{itemize}
    \item If $d = \textsf{PERMIT}$, $r$ includes execution effects and $S'$ reflects their completion.
    \item If $d = \textsf{DEFER}$, $S' = S$.
    \item If $d = \textsf{DENY}$, $S' = S$ and execution terminates.
\end{itemize}

The AAB itself does not modify $S$; it observes and enforces over it.

\subsection{Safety, Liveness, and Fail-Secure Guarantees}
\label{sec:system-guarantees}

The AAB provides the following guarantees:

\paragraph{Safety.}
No unauthorized action executes. Formally, for all actions $A$:
\[
\textsf{Eval}(A, P, S) \neq \textsf{PERMIT} \Rightarrow A \text{ does not execute}.
\]

\paragraph{Liveness.}
Every submitted action eventually yields a decision:
\[
\forall A, \exists d \in \{\textsf{PERMIT}, \textsf{DEFER}, \textsf{DENY}\}.
\]

\paragraph{Fail-Secure Behavior.}
If the AAB cannot render a decision due to failure, missing policy, or timeout, execution defaults to \textsf{DENY}.

\paragraph{Non-Bypassability.}
All execution-capable actions must traverse the AAB prior to execution. Any execution path that bypasses authorization violates conformance.

\paragraph{Determinism.}
For identical canonical actions evaluated under identical policy and state, authorization outcomes are identical.

Existing governance and access-control systems provide subsets of these properties, but not all simultaneously at execution time ~\cite{1}, ~\cite{4}, ~\cite{8}, ~\cite{9}.

Together, these properties establish the AAB as a minimal, sound execution-control primitive. The boundary neither assumes correctness of agent reasoning nor completeness of policy, but ensures that execution is governed by explicit, auditable, and fail-secure authorization semantics.

\subsection{Implementation Notes (Non-Normative)}
\label{sec:implementation-notes}

This subsection records practical considerations observed during implementation. It is explicitly non-normative: none of the design choices described here are required for conformance with the enforcement semantics specified above. The purpose is to demonstrate implementability, clarify performance implications, and delineate which concerns are architectural requirements versus engineering tradeoffs.

\paragraph{Canonicalization Placement.}
In practice, canonicalization is implemented as a pure, side-effect-free transformation that executes prior to authorization evaluation and prior to any persistence or execution logic. Implementations typically colocate canonicalization with the authorization boundary rather than embedding it in agent frameworks or protocol handlers. This ensures that all execution-capable actions, regardless of origin, are normalized uniformly and prevents framework-specific bypass paths.

\paragraph{Decision Engine Isolation.}
To preserve determinism and simplify reasoning, implementations benefit from isolating the authorization evaluator as a pure function over $(A, P, S)$. In particular, the decision engine should not perform network I/O, remote lookups, or mutable state updates during evaluation. External signals (including human approvals) are incorporated into system state $S$ prior to evaluation rather than queried dynamically during decision computation.

\paragraph{Latency Considerations.}
In practice, canonicalization and authorization evaluation introduce bounded overhead for typical action schemas when implemented with in-memory policy representations. Because deferred actions do not execute synchronously, latency sensitivity applies primarily to the \textsf{PERMIT} path. Implementations may optimize by caching policy fragments or precomputing state-derived constraints, provided such optimizations do not violate determinism.

\paragraph{Concurrency and Idempotence.}
In distributed deployments, the same canonical action may be evaluated multiple times due to retries or concurrent submission. Implementations therefore treat authorization evaluation as idempotent and side-effect-free. Any state mutation associated with execution (e.g., accounting, rate tracking) is performed strictly after a \textsf{PERMIT} decision and is bound to the execution outcome rather than the evaluation itself.

\paragraph{Deferred Action Handling.}
Deferred actions are typically persisted alongside their canonical hash and evaluated repeatedly as system state evolves. Importantly, re-evaluation does not require re-running agent reasoning or reconstructing intent proposals. Implementations that support deferred execution benefit from modeling deferral as a suspended execution record keyed by $(A, h)$ rather than as a long-lived workflow.

\paragraph{Failure Modes.}
Implementations must assume partial failure: loss of policy data, unavailable state stores, or internal exceptions. Consistent with fail-secure semantics, evaluation failures are treated equivalently to \textsf{DENY}. In practice, this is achieved by structuring evaluation such that absence of required inputs produces an explicit denial rather than an implicit fallback.

\paragraph{Logging and Replay.}
Decision logs are most effective when implemented as append-only records keyed by the canonical hash $h$. This enables efficient replay, post hoc verification, and policy evolution analysis. While cryptographic chaining is not required by the enforcement semantics, implementations commonly use hash chaining to provide tamper evidence and simplify external verification.

\paragraph{Executor Compliance.}
The enforcement guarantees described in Section~\ref{sec:system-guarantees} assume that executors honor authorization outcomes. In practice, this is achieved by ensuring that execution credentials, tokens, or capabilities are only issued in response to a \textsf{PERMIT} decision and are scoped narrowly to the authorized action. Executor-side validation of authorization artifacts provides defense in depth but is not required for semantic correctness.

\paragraph{What This Section Does Not Specify.}
These notes do not prescribe programming languages, policy languages, storage backends, or deployment topologies. They do not assume a particular agent framework, protocol, or infrastructure environment. Their sole purpose is to demonstrate that the enforcement semantics defined above are implementable with predictable performance and without introducing hidden coupling to agent reasoning or orchestration layers.

\section{Architecture and System Design}
\label{sec:architecture}

This section describes reference architectures that realize the enforcement semantics defined in Sections~\ref{sec:aab}--\ref{sec:enforcement}. It introduces no new authorization semantics. All architectural variants presented here preserve the Action Authorization Boundary (AAB), Canonical Action Representation (CAR), and enforcement guarantees previously defined.

The goal of this section is to demonstrate that execution-time authorization, as formalized earlier, can be implemented across a range of deployment topologies and operational environments without weakening determinism, non-bypassability, or auditability.

\subsection{Single-Agent Local Mode}
\label{sec:single-agent}

In its simplest configuration, Faramesh operates in a single-agent, single-tool environment. This mode is intended for local execution, embedded agents, or development settings where governance must be enforced without a centralized control plane.

In single-agent local mode, the agent and governor execute within the same trust domain and typically the same process or host. The agent emits an intent proposal $I$, which is canonicalized into $A = \mathrm{Canon}(I)$ and evaluated by the local AAB before any execution-capable tool invocation occurs. The enforcement path is strictly linear:

\[
I \rightarrow A \rightarrow d = \mathrm{Eval}(A, P, S)
\]

where $P$ denotes the active policy set and $S$ the local system state.

\paragraph{Invariant Preservation.}
Authorization semantics, CAR invariants, and AAB guarantees are identical to those in multi-agent or distributed deployments. In particular:
\begin{itemize}
  \item All execution-capable actions traverse the AAB.
  \item Canonicalization is deterministic and policy-agnostic.
  \item Enforcement is fail-closed.
\end{itemize}

Local embedding does not imply relaxed guarantees; it is a deployment optimization, not a semantic downgrade ~\cite{6}.

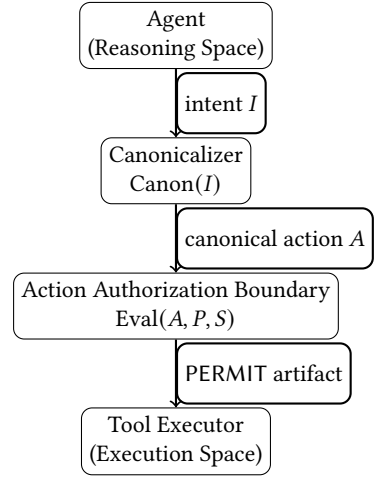
\begin{figure}[t]
\centering
\begin{tikzpicture}[
  node distance=1.8cm,
  every node/.style={draw, rectangle, rounded corners, align=center, minimum height=8mm},
  arrow/.style={->, thick}
]

\node (agent) {Agent\\(Reasoning Space)};
\node (canon) [below of=agent] {Canonicalizer\\$\mathrm{Canon}(I)$};
\node (aab) [below of=canon] {Action Authorization Boundary\\$\mathrm{Eval}(A,P,S)$};
\node (exec) [below of=aab] {Tool Executor\\(Execution Space)};

\draw[arrow] (agent) -- node[right]{intent $I$} (canon);
\draw[arrow] (canon) -- node[right]{canonical action $A$} (aab);
\draw[arrow] (aab) -- node[right]{\textsf{PERMIT} artifact} (exec);

\end{tikzpicture}
\caption{Single-agent local enforcement path. Intent generation, canonicalization, authorization, and execution are strictly separated. Execution occurs if and only if the Action Authorization Boundary returns \textsf{PERMIT}.}
\label{fig:single-agent}
\end{figure}

\subsection{Multi-Agent, Multi-Tenant Control Plane}
\label{sec:multi-tenant}

In production environments, Faramesh is deployed as a shared control plane governing multiple agents, services, and organizations. In this mode, the AAB operates as a logically centralized authority, even if physically distributed for scalability or fault tolerance. ~\cite{12}, ~\cite{13}. 

\paragraph{Tenant Isolation.}
Each tenant is associated with:
\begin{itemize}
  \item a disjoint policy namespace $P_t$,
  \item an isolated identity space,
  \item independent system state $S_t$ for authorization evaluation.
\end{itemize}

Cross-tenant policy coupling is explicitly disallowed. No authorization decision for tenant $t_i$ may depend on policies, actions, or state belonging to tenant $t_j \neq t_i$.

Formally, for tenants $t_i, t_j$:
\[
\mathrm{Eval}(A, P_{t_i}, S_{t_i}) \;\;\bot\;\; P_{t_j}, S_{t_j}
\]

This constraint prevents policy bleed-through and simplifies compliance and audit reasoning.

\paragraph{Identity Mapping.}
Agent identities, service principals, and delegated credentials are resolved into canonical \texttt{Actor} fields during canonicalization. Identity systems (IAM, OAuth, mTLS) are treated as upstream signal providers; they do not participate in authorization semantics directly.

\begin{figure}[t]
\centering
\begin{tikzpicture}[
  node distance=1.6cm,
  every node/.style={draw, rectangle, rounded corners, align=center, minimum height=8mm},
  arrow/.style={->, thick}
]

\node (agent1) {Agent A};
\node (agent2) [right of=agent1, xshift=2.2cm] {Agent B};

\node (canon) [below of=agent1, xshift=1.8cm] {Canonicalization\\$\mathrm{Canon}(I)$};

\node (aab) [below of=canon] {Shared AAB Control Plane\\$\mathrm{Eval}(A,P_t,S_t)$};

\node (exec1) [below of=aab, xshift=-2.0cm] {Executor\\Tenant $t_1$};
\node (exec2) [below of=aab, xshift=2.0cm] {Executor\\Tenant $t_2$};

\draw[arrow] (agent1) -- (canon);
\draw[arrow] (agent2) -- (canon);
\draw[arrow] (canon) -- (aab);
\draw[arrow] (aab) -- node[left]{\textsf{PERMIT}$_{t_1}$} (exec1);
\draw[arrow] (aab) -- node[right]{\textsf{PERMIT}$_{t_2}$} (exec2);

\end{tikzpicture}
\caption{Multi-agent, multi-tenant control plane. Multiple agents submit canonical actions to a shared Action Authorization Boundary. Authorization is evaluated against tenant-isolated policy $P_t$ and state $S_t$, and execution artifacts are scoped per tenant.}
\label{fig:multi-tenant}
\end{figure}
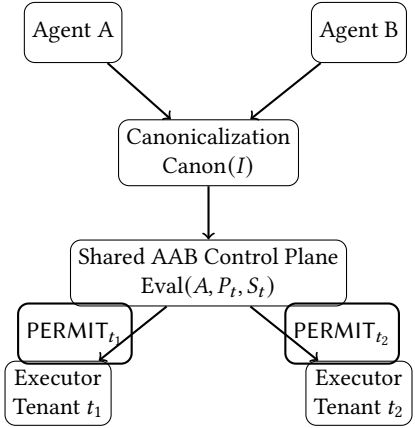

\subsection{Protocol Independence}
\label{sec:protocol-independence}

Faramesh is protocol-independent by construction. Agent-to-governor communication protocols such as MCP, UTCP, A2A, or ANP are treated strictly as transport substrates.

Protocols may:
\begin{itemize}
  \item carry intent proposals,
  \item convey identity or metadata,
  \item transport decisions and artifacts.
\end{itemize}

Protocols may not:
\begin{itemize}
  \item encode authorization semantics,
  \item short-circuit the AAB,
  \item influence canonicalization or policy evaluation.
\end{itemize}

This separation ensures that authorization correctness does not depend on protocol-specific behavior. Any protocol capable of transmitting a canonical action representation and receiving a decision artifact can serve as an upstream substrate. ~\cite{9}, ~\cite{10}

\subsection{Deployment Models}
\label{sec:deployment}

Faramesh supports multiple deployment models, all of which implement identical semantics.

\paragraph{SDK-Only.}
Canonicalization and AAB logic are embedded directly into the agent runtime. This model minimizes latency but assumes host integrity.

\paragraph{Sidecar.}
The governor runs as a co-located process or container, mediating execution requests from one or more agents on the same host.

\paragraph{Proxy.}
All execution-capable requests are routed through an external authorization proxy that enforces AAB decisions before forwarding to tools or APIs.

\paragraph{Managed Service.}
The governor operates as a hosted control plane, providing centralized policy management, decision logging, and audit interfaces.

The choice of deployment affects operational characteristics (latency, fault tolerance) but not authorization semantics.

\subsection{Failure Modes}
\label{sec:failure}

Faramesh enforces fail-secure behavior under all failure modes.

\paragraph{Governor Failure.}
If the governor cannot render a decision due to crash, timeout, or partition, the default outcome is \textsf{DENY}. No execution-capable action may proceed without an explicit \textsf{PERMIT} artifact.

\paragraph{Network Partition.}
Loss of connectivity between agent and governor results in deferred or denied execution, depending on configuration, but never implicit permission.

\subsection{Reliability and Performance Considerations}
\label{sec:reliability}

\paragraph{Replay Capability.}
Because authorization decisions are defined over canonical actions $A$, historical actions can be replayed against new policy sets:
\[
d' = \mathrm{Eval}(A, P', S')
\]
without re-executing agent reasoning. This supports audit, simulation, and policy evolution.

\paragraph{Caching.}
Decision caching is permitted only over tuples $(A, P, S)$ and must respect idempotence:
\[
\mathrm{Eval}(A, P, S) = \mathrm{Eval}(A, P, S)
\]
Caching may improve performance but must not violate determinism or policy freshness.

\subsection{Reference API and Protocol Sketch}
\label{sec:api}

This section provides an illustrative, non-normative API sketch demonstrating how canonical actions and decisions may be exchanged.

\begin{verbatim}
POST /v1/action/commit
{
  "actor": "...",
  "operation": "...",
  "resource": "...",
  "parameters": {},
  "context": {}
}
\end{verbatim}

\begin{verbatim}
Response:
{
  "decision": "permit|defer|deny",
  "explanation": "string",
  "hash": "abc...",
  "timestamp": "..."
}
\end{verbatim}

The \texttt{hash} field refers to the canonical action hash $h = H(A)$.

\paragraph{Ordering and Idempotency.}
Repeated submissions of the same canonical action $A$ must yield identical decisions under identical $(P, S)$. Retries are safe provided that execution is conditioned on a valid \textsf{PERMIT} artifact.

\paragraph{Executor Constraints.}
Tool executors \emph{must} reject any execution request lacking a valid \textsf{PERMIT} decision artifact bound to $h$. Execution without such an artifact constitutes a violation of the AAB model.

\subsection{Control Plane Structure and Consistency Model}
\label{sec:control-plane-consistency}

The architectures described above admit both centralized and distributed realizations of the Action Authorization Boundary (AAB). In multi-tenant or high-availability deployments, the governor itself constitutes a distributed control plane whose correctness properties are independent of agent behavior.

We model the control plane as a replicated state machine responsible for evaluating canonical actions $A$ against policy $P$ and system state $S$. ~\cite{8}, ~\cite{12} Policy artifacts, tenant configuration, and authorization decisions are treated as control-plane state. The consistency requirements of this state are asymmetric:

\begin{itemize}
  \item \textbf{Policy state} MUST be strongly consistent within a tenant boundary. Divergent policy views across replicas would violate determinism guarantees.
  \item \textbf{Decision logs} MUST be append-only and causally ordered but need not be globally linearizable across tenants.
  \item \textbf{Cached decisions} MAY be eventually consistent, provided they are validated against canonical action hashes.
\end{itemize}

The architecture does not require a specific replication protocol; leader-based (e.g., Raft-style) or quorum-based designs are admissible so long as the evaluation function $\text{Eval}(A,P,S)$ is executed against a well-defined policy snapshot. Failover behavior MUST preserve fail-closed semantics: loss of quorum or leader availability results in \textsf{DENY} or \textsf{DEFER}, never optimistic execution.

Importantly, the governor is not in the data plane. It does not proxy payloads, transform tool outputs, or participate in execution beyond authorization. This separation bounds the blast radius of governor failures and ensures that control-plane complexity does not propagate into execution paths.

\subsection{Formal Placement of the AAB in the Execution Call Graph}
\label{sec:aab-call-graph}

To preclude ambiguity about bypass, we formalize the placement of the Action Authorization Boundary within the execution call graph.

Let $\mathcal{E}$ denote the set of execution-capable interfaces, APIs, system calls, or RPCs, that can induce external side effects. A conformant system MUST satisfy:

\[
\forall e \in \mathcal{E}, \quad \exists A \text{ such that } e \text{ is invoked iff } \text{Eval}(A,P,S)=\textsf{PERMIT}
\]

That is, no execution-capable interface may be reachable without a prior authorization decision over a canonical action. Tool SDKs, protocol handlers, and orchestration frameworks are required to route all execution requests through the AAB or to embed an equivalent enforcement hook that validates a decision artifact.

This constraint is architectural, not advisory. Any code path, intentional or accidental, that allows execution without mediation violates conformance, regardless of whether policies would have permitted the action.

This placement is analogous to an admission controller in operating systems or container orchestration ~\cite{8}, ~\cite{9}, ~\cite{10}, but operates at the semantic level of execution intent rather than at resource allocation or syscall granularity.

\subsection{Concurrency, Deduplication, and Contention Semantics}
\label{sec:concurrency-semantics}

Agentic systems may emit concurrent or repeated proposals for identical canonical actions. The enforcement architecture therefore defines explicit semantics for concurrency and deduplication.

Let $A$ be a canonical action with hash $h = H(A)$. Multiple submissions of $A$ MAY be evaluated concurrently. The system MUST ensure that:

\begin{itemize}
  \item All evaluations of identical $(A,P,S)$ yield identical decisions.
  \item Duplicate submissions do not result in duplicate execution unless explicitly permitted by policy.
\end{itemize}

Formally, execution is guarded by a predicate:

\[
\text{Execute}(A) \iff \exists d = \textsf{PERMIT} \land \neg \text{Consumed}(h)
\]

where $\text{Consumed}(h)$ marks execution of a single-use action. Policies MAY define idempotent actions for which repeated execution is allowed; such semantics are explicit and policy-scoped, not implicit.

For \textsf{DEFER} outcomes, concurrent evaluations MAY block on the same pending approval artifact. Resolution of a deferred decision MUST transition atomically to either \textsf{PERMIT} or \textsf{DENY} for all waiting evaluations.

These rules ensure that concurrency does not undermine determinism or safety guarantees.

\subsection{Decision Artifact Lifecycle and Revocation Semantics}
\label{sec:decision-lifecycle}

Authorization decisions produce decision artifacts that bind a canonical action $A$ to an outcome $d$. These artifacts have a well-defined lifecycle.

A \textsf{PERMIT} artifact MAY include:
\begin{itemize}
  \item a canonical hash $h$,
  \item a validity window $[t_{\min}, t_{\max}]$,
  \item optional execution constraints (e.g., one-time use).
\end{itemize}

Execution engines MUST validate the artifact at execution time. If execution is delayed beyond $t_{\max}$, re-evaluation is mandatory.

Revocation semantics are explicit. A previously issued \textsf{PERMIT} MAY be revoked by policy update or operator intervention. Revocation does not retroactively invalidate completed executions, but MUST prevent future execution attempts using the revoked artifact.

This model avoids time-of-check/time-of-use ambiguity while preserving replay and auditability.

\subsection{Relationship to Existing Control-Plane Patterns}
\label{sec:control-plane-lineage}

The enforcement architecture bears structural resemblance to several established systems patterns, but is not reducible to any of them. ~\cite{8}, ~\cite{9}, ~\cite{12}.

\begin{itemize}
  \item \textbf{API Gateways} regulate ingress traffic but assume requests are intentional and human-authored.
  \item \textbf{Admission Controllers} (e.g., Kubernetes) validate resource mutations but operate on declarative objects, not synthesized intent.
  \item \textbf{Transaction Coordinators} ensure atomicity but do not authorize whether a transaction should occur.
\end{itemize}

The AAB combines elements of all three: it is an admission point, a semantic validator, and a transaction gate, but for autonomous execution intent. This positioning places the system squarely in the lineage of distributed control planes rather than access-control middleware.

\subsection{Operational Cost and Performance Considerations}
\label{sec:performance}

Canonicalization and authorization introduce overhead relative to direct tool invocation. The overhead is implementation-dependent but typically small relative to effectful tool execution.

Let $T_c$ denote canonicalization latency and $T_e$ policy evaluation latency. Total authorization latency is $T_c + T_e$, independent of tool execution time. In practice, $T_c$ is dominated by parsing and normalization, while $T_e$ is dominated by policy evaluation and state access.

Caching at the canonical hash level amortizes repeated evaluations. A cached decision for $h$ is valid if and only if the policy version and relevant state have not changed. Cache invalidation is explicit and policy-driven.

Replay-based evaluation avoids re-running agent reasoning, enabling retrospective analysis at marginal cost proportional to evaluation alone.

These properties make the system suitable for latency-sensitive environments while preserving strong correctness guarantees.

\subsection{Summary}
\label{sec:architecture-summary}

This section demonstrates that the enforcement semantics defined earlier admit concrete, scalable system architectures without weakening guarantees. The governor is a control-plane system with explicit consistency requirements, bounded failure modes, and well-defined execution interfaces. By separating intent synthesis from execution authorization and grounding enforcement in canonical action semantics, the architecture achieves determinism, auditability, and non-bypassability at scale.

\section{Provenance, Forensics, and Replayability}
\label{sec:provenance}
\subsection{Audit Log Architecture}

Execution governance is only meaningful if authorization decisions can be reconstructed, verified, and re-evaluated independently of the systems that produced or executed them. To support these requirements, Faramesh maintains an explicit \emph{decision-centric audit log} whose unit of record is not an execution event, but an authorization decision bound to a canonical action representation. This log is append-only, tamper-evident, and time-sequenced, but it is not an event log in the conventional observability sense. Its purpose is to capture the authorization frontier: the precise point at which proposed execution intent was evaluated and adjudicated ~\cite{16}, ~\cite{17}.

\paragraph{Decision Records.}

Each authorization evaluation produces a \emph{Decision Record}, which is the sole primitive admitted to the audit log. Formally, the portable decision record stored in the log is:
\[
r_i = (\mathit{seq}_i, h_{A,i}, v_{P,i}, h_{S,i}, d_i, t_i, \mathit{prev\_hash}_i)
\]
where $h_{A,i}=H(A_i)$ commits to the canonical action $A_i$, $v_{P,i}$ uniquely identifies the exact policy semantics used at decision time, and $h_{S,i}=H(S_i)$ commits to the evaluation-relevant system state (or its digest).

Internally, an implementation MAY retain full objects $(A_i,P_i,S_i)$ for convenience, but verification and replay are defined over the portable record by resolving $(h_{A,i}\mapsto A_i,\ v_{P,i}\mapsto P_i,\ h_{S,i}\mapsto S_i)$.

where $A_i$ is the canonical action representation (CAR) evaluated at decision time; $h_i = H(A_i)$ is the cryptographic hash of the canonical action; $d_i \in \{\textsf{PERMIT}, \textsf{DEFER}, \textsf{DENY}\}$ is the authorization outcome; $P_i$ identifies the exact policy version(s) and rule context applied; $S_i$ captures the relevant system state snapshot or state digest used during evaluation; $t_i$ is a monotonically ordered decision timestamp or logical clock value; and $\mathit{prev\_hash}_i$ is the hash pointer to the immediately preceding decision record in the log ~\cite{17}, ~\cite{18}.

The log contains only decision records. Execution events, tool invocations, and side effects are explicitly excluded. This distinction ensures that the audit trail remains \emph{decision-complete} rather than event-best-effort: every external effect must correspond to a prior authorization decision, but not every authorization decision must result in execution.

\begin{definition}[Decision Provenance Record]
A decision record $r_i$ is \emph{provenance-complete} iff it contains sufficient information to (1) re-evaluate authorization deterministically under identical inputs and (2) verify non-tampering of both the decision and its ordering without access to agent reasoning traces or execution systems.
\end{definition}

From this definition, the audit log satisfies three explicit properties:
\begin{itemize}
  \item \textbf{Provenance completeness.} Every authorization decision is recorded with all inputs required for independent verification.
  \item \textbf{Replay sufficiency.} Decision records alone are sufficient to support replay-based re-evaluation (Section~7.4).
  \item \textbf{Non-repudiation (within the TCB).} Once appended, neither the decision outcome nor its evaluation context can be altered without detection.
\end{itemize}

\paragraph{Append-Only and Tamper-Evident Semantics.}
The audit log enforces a strict append-only invariant:
\begin{quote}
\textbf{Invariant (Append-Only).} For any valid log prefix $\langle r_1, \dots, r_n \rangle$, no operation may remove, reorder, or mutate any $r_i$ once appended.
\end{quote}

Tamper evidence is achieved via hash chaining: each record commits to the previous record’s hash. Any deletion, reordering, or mutation breaks the chain and is detectable by recomputation. This mechanism provides tamper evidence, not tamper resistance; the trusted computing base (TCB) must still ensure correct log construction and persistence.

Importantly, tamper evidence applies to authorization semantics, not execution outcomes. The log proves what decision was made, under which context, and in what order, not whether downstream systems complied.

\paragraph{Ordering Model.}
Authorization decisions must be ordered to support forensics and replay, but global real-time total ordering is neither necessary nor desirable in distributed agent systems. Faramesh therefore adopts the following ordering guarantees:
\begin{itemize}
  \item \textbf{Total order per tenant or policy namespace.} All decision records within a tenant or policy domain are totally ordered.
  \item \textbf{Causal ordering across decision chains.} Decisions that arise from or depend on prior decisions preserve causal order via explicit linkage in $S_i$ or policy context.
  \item \textbf{Explicit non-goal: global total ordering.} The system does not impose a single global real-time order across tenants or independent policy domains.
\end{itemize}

This ordering model reflects a deliberate trade-off: per-domain determinism and causal consistency are sufficient for auditability and replay, while global ordering would conflate unrelated decision streams and obscure causality behind wall-clock artifacts.

\paragraph{Why This Is Not an Event Log.}
Traditional audit and observability systems record effects: API calls, state mutations, and network requests. Such logs answer the question \emph{what happened}. They do not, and cannot, answer \emph{why it was allowed}. ~\cite{7}, ~\cite{16}.

The Faramesh audit log records authorization decisions, not side effects. Each record captures the binding:
\[
(A_i, P_i, S_i) \rightarrow d_i
\]
freezing the exact evaluation context at decision time. This binding is the authorization boundary itself. Execution logs may later confirm that an effect occurred, but only decision records can establish whether that effect was authorized, under which policy, and with what assumptions.

Observability systems operate downstream of execution. Provenance systems operate at the boundary between intent and effect. Faramesh records the latter ~\cite{12}, ~\cite{18}.

\begin{quote}
\emph{This distinction is architectural, not operational.}
\end{quote}

\paragraph{Summary.}
The audit log architecture defined here elevates authorization decisions to first-class, replayable system artifacts. By logging canonical actions, evaluation context, and outcomes in an append-only, causally ordered structure, Faramesh enables forensic reconstruction and policy evolution without reliance on agent internals or execution systems. Subsequent sections build on this foundation to link decisions to actions (Section~7.2), export provenance to external consumers (Section~7.3), and support replay-based analysis (Section~7.4).

\subsection{Linking Decisions to Actions}

Auditability in autonomous systems requires more than recording that an execution occurred. It requires reconstructing \emph{why} a particular execution was permitted, deferred, or denied under the policy and system context that existed at decision time. For this reason, Faramesh treats authorization decisions, not execution events, as the primary unit of provenance and links execution explicitly to prior decisions rather than inferring intent from side effects.

\paragraph{Why Execution Logs Are Insufficient.}
Execution logs record effects: API calls, state mutations, network requests, or tool invocations. These records answer the question \emph{what happened}, but they do not capture the authorization rationale that allowed the event to occur ~\cite{7},~\cite{16}. In agentic systems, this distinction is critical. An execution event does not encode whether it was:
(i) explicitly permitted by policy,
(ii) allowed due to missing or permissive policy,
(iii) executed after human escalation, or
(iv) executed due to enforcement failure.

Without an explicit decision record, these cases are observationally indistinguishable after the fact. Any attempt to infer authorization from execution artifacts alone is therefore underdetermined.

\paragraph{Decision-Centric Provenance.}
Faramesh captures authorization at the point where intent crosses into effect. Each execution-capable action is linked to a unique prior authorization decision via the canonical action hash $h = H(A)$. Formally, execution is admissible only if there exists a decision record $r_i$ such that:
\[
(A_i, P_i, S_i) \rightarrow d_i = \textsf{PERMIT}
\]
and $H(A_i)$ matches the action presented to the executor. Execution systems consume a decision artifact derived from $r_i$; absent such an artifact, execution must be rejected ~\cite{9}, ~\cite{10}.

This linkage ensures that provenance is anchored at the authorization boundary rather than reconstructed downstream. Execution events become consequences of decisions, not evidence of them.

\paragraph{Binding Authorization Context.}
A decision record binds three inputs to an outcome:
\[
(A, P, S) \rightarrow d
\]
where $A$ is the canonical action representation, $P$ is the precise policy set and version applied, and $S$ is the system state snapshot (or digest) used during evaluation. This binding is frozen at decision time and recorded immutably ~\cite{11}, ~\cite{12}.

Freezing policy version and evaluation context is essential. Policies evolve, external signals change, and system state mutates. Without binding $P$ and $S$ at the moment of decision, post hoc interpretation becomes ambiguous: a later observer cannot determine whether an action was allowed because policy permitted it, because policy was absent, or because state-dependent constraints differed.

By recording the complete evaluation context, Faramesh enables third-party verification of authorization decisions without access to agent reasoning traces, prompt histories, or execution systems.

\begin{definition}[Decision Provenance Record]
A decision record $r_i$ is complete iff it contains sufficient information to (1) re-evaluate authorization deterministically under identical inputs and (2) verify non-tampering of the decision and its ordering without access to agent reasoning or execution infrastructure.
\end{definition}

From this definition, Faramesh explicitly guarantees:
\begin{itemize}
  \item \textbf{Provenance completeness:} every authorization decision is self-contained.
  \item \textbf{Replay sufficiency:} decisions can be re-evaluated under new policy or state (Section~7.4).
  \item \textbf{Non-repudiation (within the TCB):} decisions cannot be altered or denied after the fact.
\end{itemize}

\paragraph{Ordering and Causality.}
Linking decisions to actions requires well-defined ordering guarantees. Faramesh enforces:
\begin{itemize}
  \item a total order over decisions within each tenant or policy namespace,
  \item causal ordering across decision chains that depend on prior decisions or state,
  \item and explicitly does \emph{not} enforce a global real-time order across tenants.
\end{itemize}

Global total ordering is unnecessary for forensic reconstruction and actively harmful for interpretation: it conflates unrelated decision streams and obscures causal structure behind wall-clock artifacts. Per-tenant total order suffices to reconstruct authorization history within a trust domain, while causal ordering captures dependency relationships that matter for correctness and accountability.

\paragraph{Separation from Observability Systems.}
This design intentionally diverges from conventional observability and logging platforms. Systems such as SIEMs, infrastructure logs, or cloud audit trails record execution effects after the fact. They are optimized for detection, aggregation, and alerting, not for reconstructing authorization semantics.

In contrast:
\begin{quote}
\emph{Observability systems record what happened. Provenance systems record what was decided. Faramesh records the authorization boundary, which observability tools never see.}
\end{quote}

This distinction is architectural, not operational. Faramesh does not replace execution logging; it provides the missing semantic layer that makes execution logs interpretable in the presence of autonomous agents.

\paragraph{Summary.}
By explicitly linking execution to prior authorization decisions via canonical action hashes and frozen evaluation context, Faramesh ensures that every side effect is attributable to a concrete, verifiable decision. This linkage transforms audit trails from best-effort event histories into decision-complete provenance records, enabling deterministic replay, third-party verification, and principled forensic analysis.

\subsection{Export to Security and Risk Analysis Systems}

The provenance artifacts defined in Sections~7.1 and~7.2 are designed to be \emph{exportable}, not \emph{integrated}, into downstream security and risk analysis systems. This distinction is intentional. Faramesh does not delegate authority, interpretation, or enforcement to external observability or SOC tooling. Instead, it produces formally defined authorization artifacts that may be consumed by such systems for analysis, correlation, and reporting.

\paragraph{Unidirectional Projection Model.}
Export from Faramesh to downstream systems is strictly unidirectional. Decision provenance records are projected outward as immutable facts; no external system may modify, override, or retroactively reinterpret authorization outcomes. Formally, let $\mathcal{R}$ denote the decision record log maintained by Faramesh and $\Pi$ a projection function that maps records to an external representation:
\[
\Pi : \mathcal{R} \rightarrow \mathcal{O}
\]
where $\mathcal{O}$ is a read-only observation space. $\Pi$ preserves record identity, ordering, and cryptographic linkage but may omit fields irrelevant to downstream analysis (e.g., executor tokens). Projection does not alter the authoritative log.

This model prevents feedback loops in which downstream systems influence enforcement semantics, preserving the Action Authorization Boundary as the sole locus of execution control.

\paragraph{SOC Systems as Consumers, Not Authorities.}
Security operations platforms, such as SIEMs or risk analytics pipelines, are treated as passive consumers of provenance artifacts. They may aggregate, correlate, alert, or visualize exported data, but they are explicitly not part of the trusted computing base and are never consulted during authorization evaluation ~\cite{3}, ~\cite{4}.

This separation avoids a common failure mode in distributed governance systems, where control logic becomes entangled with monitoring infrastructure, leading to ambiguous authority and non-deterministic enforcement.

\paragraph{Exported Artifact Semantics.}
Each exported artifact corresponds to a complete decision provenance record $r_i$, including:
\[
r_i = (\mathit{seq}_i, h_{A,i}, v_{P,i}, h_{S,i}, d_i, t_i, \mathit{prev\_hash}_i)
\]
Projection preserves the binding between canonical action $A_i$, decision $d_i$, and evaluation context $(P_i, S_i)$. Downstream systems therefore observe \emph{authorization facts}, not inferred intent or reconstructed policy state.

This is a critical distinction from conventional execution logs, which expose only side effects and require interpretation to infer governance semantics.

\paragraph{Mapping to Control Objectives.}
Rather than mapping directly to specific vendors or products, exported decision records align with common control objectives in regulated and enterprise environments. Examples include:
\begin{itemize}
  \item \textbf{Immutable audit trails:} append-only, tamper-evident decision logs support requirements for non-repudiation and post-incident review.
  \item \textbf{Authorization accountability:} explicit binding of actions to decisions enables verification that every execution was explicitly permitted.
  \item \textbf{Change traceability:} frozen policy versions and evaluation context support reconstruction of governance posture at any point in time.
  \item \textbf{Post-incident reconstruction:} decision-complete logs enable analysis of why actions were allowed, not merely that they occurred.
\end{itemize}

These objectives are orthogonal to any particular tooling ecosystem and remain stable even as downstream platforms evolve.

\paragraph{Distinction from Event-Centric Logging.}
Traditional SOC pipelines ingest high-volume event streams optimized for detection and alerting. Faramesh exports low-volume, semantically rich decision records optimized for explanation and verification. The former answers \emph{what happened}; the latter answers \emph{why it was allowed} ~\cite{7}, ~\cite{16}.

Because decision provenance records are decision-complete and causally ordered, they can be correlated with execution events when needed, but they do not depend on such correlation to retain meaning. This asymmetry is deliberate.

\paragraph{Non-Goals.}
Faramesh does not attempt to standardize downstream analysis workflows, dictate alerting strategies, or embed enforcement hooks in SOC platforms. Exported artifacts are designed to be sufficient for independent analysis, not to prescribe operational response.

\paragraph{Summary.}
By treating SOC and risk analysis systems as consumers of formally defined authorization artifacts rather than participants in enforcement, Faramesh preserves a clean separation of concerns. Authorization semantics remain centralized, deterministic, and auditable, while external systems gain access to high-integrity provenance data for analysis and compliance purposes. This architecture enables interoperability without diluting authority or weakening guarantees established at the Action Authorization Boundary.

\subsection{Reconstructing Behavior via Deterministic Replay}
\label{sec:replay}

Decision-complete provenance enables a final and qualitatively stronger capability than audit or observability alone: \emph{deterministic replay}. Replay allows an authorization system to re-evaluate historical execution proposals under new policy or state assumptions without re-executing actions, re-running agent reasoning, or mutating external systems. In Faramesh, replay is not an auxiliary feature; it is a direct consequence of canonicalization, deterministic authorization, and append-only decision provenance.

\paragraph{Replay Is Not Re-execution.}
Replay must be defined precisely to avoid conflation with execution or simulation. Execution is an effectful operation that produces irreversible side effects in external systems. Replay is a pure analytical operation that produces only an authorization outcome ~\cite{16}, ~\cite{17}.

Formally, let $I$ denote an original agent intent proposal and $A = \textsf{Canon}(I)$ its canonical action representation. Replay is defined as:
\[
\textsf{Replay}(A, P', S') \rightarrow d'
\]
where $P'$ is a policy context (possibly distinct from the original policy $P$), $S'$ is an evaluation-relevant system state, 

and $d' \in \{\textsf{PERMIT}, \textsf{DEFER}, \textsf{DENY}\}$. Replay produces no execution, invokes no tools, and induces no state transitions outside the authorization engine. ~\cite{16}, ~\cite{17}.

Replay answers counterfactual authorization questions of the form: \emph{“What would the authorization outcome have been if this action were proposed under different constraints?”} It does not answer \emph{“What should happen now?”} nor does it attempt to re-enact historical behavior.

\paragraph{Separation from Agent Reasoning.}
Replay explicitly excludes agent reasoning. No prompts are reprocessed, no plans regenerated, and no stochastic inference reintroduced. ~\cite{5} This exclusion is intentional and necessary. Agent reasoning is probabilistic, opaque, and unstable across time; replay requires determinism. By operating exclusively on canonical action representations, replay avoids representational drift and isolates authorization semantics from cognitive variability.

This separation ensures that replay does not depend on the availability, correctness, or reproducibility of upstream agent logic.

\paragraph{Decision Provenance Record Sufficiency.}
Replay is only possible if authorization decisions are recorded with sufficient semantic completeness. We therefore formalize the notion of a decision provenance record.

\begin{definition}[Decision Provenance Record]
A decision provenance record is the portable log record:
\[
r_i = (\mathit{seq}_i, h_{A,i}, v_{P,i}, h_{S,i}, d_i, t_i, \mathit{prev\_hash}_i)
\]
which is sufficient to verify non-tampering and deterministically replay authorization by resolving $(h_{A,i}\mapsto A_i,\ v_{P,i}\mapsto P_i,\ h_{S,i}\mapsto S_i)$.
\end{definition}

\noindent
A decision provenance record is \emph{complete} if it contains sufficient information to deterministically re-evaluate authorization and to verify non-tampering without access to agent reasoning, execution systems, or live infrastructure.

From this definition follow three system-level guarantees:
\begin{itemize}
  \item \textbf{Provenance completeness:} every authorization decision is fully reconstructible.
  \item \textbf{Replay sufficiency:} replay requires no external data beyond the record itself.
  \item \textbf{Non-repudiation (within the TCB):} decisions cannot be altered or denied without violating cryptographic linkage.
\end{itemize}

\paragraph{Ordering Guarantees and Their Limits.}
Replay requires ordering, but not global total ordering. Faramesh guarantees:
\begin{itemize}
  \item a total order per tenant or policy namespace, and
  \item a causal order across decision chains.
\end{itemize}

A global real-time total order across all tenants and systems is explicitly a non-goal. Such an order is unnecessary for forensic reconstruction and impractical in distributed environments. Authorization decisions are scoped to administrative and policy domains; replay analysis operates within those scopes.

Per-tenant total order ensures that authorization history within a domain is unambiguous. Causal ordering preserves dependency relationships, such as deferred approvals, chained actions, or policy-triggered follow-on decisions. Wall-clock time alone is insufficient: causality, not simultaneity, determines interpretability.

\paragraph{Replay Safety.}
Replay is safe by construction. It satisfies three safety properties:
\begin{enumerate}
  \item \textbf{No execution:} replay invokes no tools and produces no external effects.
  \item \textbf{No state mutation:} replay does not alter system or policy state.
  \item \textbf{No cognitive re-entry:} replay does not re-run agent reasoning or inference.
\end{enumerate}

These properties ensure that replay cannot amplify risk, re-trigger incidents, or introduce new failure modes. Replay is observational, not operational.

\paragraph{Replay as a First-Class Capability.}
Replay enables analyses that are otherwise infeasible in agent-driven systems:
\begin{itemize}
  \item \textbf{Retroactive policy tightening:} organizations can evaluate whether historical actions would be permitted under stricter controls.
  \item \textbf{Compliance audits:} regulators can verify authorization posture at historical points in time without trusting live systems.
  \item \textbf{Incident forensics:} investigators can reconstruct not only what occurred, but why the system allowed it.
  \item \textbf{Counterfactual evaluation:} teams can safely ask, “Would this be allowed today?” without risk.
\end{itemize}

In each case, replay operates on immutable decision artifacts, not on mutable execution traces.

\paragraph{Lemma: Replay Soundness.}
\begin{lemma}[Replay Soundness]
Re-evaluating a canonical action $A$ under a new policy context $P'$ and system state $S'$ without executing side effects yields an authorization outcome equivalent to a hypothetical execution-time decision, modulo differences between $(P, S)$ and $(P', S')$.
\end{lemma} ~\cite{11}, ~\cite{12}.

\begin{proof}[Proof Sketch]
Canonicalization removes representational variance, ensuring that $A$ uniquely captures execution semantics. Deterministic evaluation ensures that authorization is a pure function of $(A, P, S)$. Separation of authorization from execution guarantees that replay introduces no side effects or nondeterminism. Therefore, replay faithfully simulates the authorization outcome that would have occurred at execution time under $(P', S')$.
\end{proof}

This lemma elevates replay from an operational convenience to a formally grounded system capability.

\paragraph{Relation to Observability and Audit.}
Observability systems record \emph{what happened}. Audit systems record \emph{what executed}. Provenance systems record \emph{what was decided}. Faramesh records the authorization boundary, the precise point at which intent is evaluated and either permitted or denied.

This boundary is invisible to traditional observability stacks. It cannot be inferred from execution logs, metrics, or traces. Capturing it requires explicit architectural instrumentation at the authorization layer.

This distinction is architectural, not operational.

\section{Evaluation}
\label{sec:evaluation}

This section evaluates the Action Authorization Boundary (AAB) and Canonical Action Representation (CAR) as
\emph{architectural requirements} for execution-time authorization in agentic systems.
The goal of this evaluation is not to demonstrate application-level performance improvements,
nor to argue that governed agents behave ``better'' in an end-to-end task sense.
Rather, the evaluation addresses a more fundamental systems question:

\begin{quote}
\emph{Which properties of execution-time authorization become ill-defined, unverifiable, or unsound
in the absence of a mandatory Action Authorization Boundary and Canonical Action Representation?}
\end{quote}

Accordingly, the evaluation is structured to establish four claims, explicitly and repeatedly:

\begin{enumerate}
  \item \textbf{Authorization correctness cannot be inferred from observability alone.}
        Execution logs and monitoring data are insufficient to reconstruct why an action was permitted or denied.

  \item \textbf{Deterministic authorization is impossible without canonicalization.}
        Without CAR, semantically equivalent intents do not form a stable evaluation domain.

  \item \textbf{Replay is impossible without provenance-complete decision records.}
        Counterfactual re-evaluation under evolving policies is ill-defined without explicit decision provenance.

  \item \textbf{The overhead required to obtain these properties is asymptotically minimal.}
        The costs incurred by canonicalization, evaluation, and logging are architectural lower bounds.
\end{enumerate}

All experimental results in this section should therefore be interpreted as evidence supporting these claims.
Metrics such as latency or decision frequency are not treated as performance optimizations,
but as necessary costs for achieving determinism, replayability, and non-bypassability.

\paragraph{Evaluation Scope Disclaimer.}
This evaluation does \emph{not} attempt to measure agent reasoning quality, policy correctness,
tool safety, or end-to-end task success.
Those concerns lie outside the scope of execution-time authorization.
Instead, the evaluation isolates and measures only the soundness properties
introduced by the AAB and CAR: determinism, replayability, and enforcement completeness.

\subsection{Methodology}
\label{sec:evaluation-methodology}

This evaluation adopts a methodology designed to isolate \emph{architectural soundness properties} of execution-time authorization rather than application-level behavior.
Specifically, it evaluates which properties become unverifiable, non-deterministic, or ill-defined when the Action Authorization Boundary (AAB) and Canonical Action Representation (CAR) are absent.
The methodology is therefore adversarial by construction: it is designed to break authorization semantics, not to demonstrate throughput or functional success.

\paragraph{Evaluation Objective.}
All experiments in this section evaluate authorization decisions over canonical actions
$A = \mathrm{Canon}(I)$,
with decisions computed as
$d = \mathrm{Eval}(A, P, S)$
under the enforcement semantics defined in Section~\ref{sec:enforcement}.
System state $S$ and policy program $P$ are treated as explicit inputs and are controlled directly.
This isolates determinism, replayability, and non-bypassability from confounding factors such as agent reasoning quality or tool behavior.

\paragraph{Adversarial Synthetic Workloads.}
Rather than relying on production traces, we construct adversarial synthetic workloads that systematically stress failure modes inherent to execution-time authorization. ~\cite{1}, ~\cite{2}
These workloads are not synthetic for convenience, but because the properties under evaluation cannot be observed reliably in real traces.
Each workload class is constructed to violate a specific architectural assumption when CAR or AAB is removed.

\begin{itemize}
  \item \textbf{Semantic Equivalence under Syntactic Drift.}
        Semantically identical execution intents are emitted with divergent surface representations,
        including reordered parameters, alternate tool schemas, and varied serialization.
        This workload tests whether authorization decisions remain invariant under representational drift,
        as required by CAR invariants (Section~\ref{sec:car}).

  \item \textbf{Policy-Evolving Timelines.}
        Policy sets $P$ are modified over time while canonical actions remain fixed.
        This exposes time-of-check/time-of-use effects and evaluates whether past actions can be
        re-evaluated deterministically under new policies without re-execution.

  \item \textbf{Intent Ambiguity under Identical Tool Calls.}
        Distinct high-level intents are mapped to identical low-level tool invocations.
        This workload demonstrates that tool-local or observability-based authorization
        cannot recover decision semantics without canonicalization.

  \item \textbf{Partial Failure Conditions.}
        Authorization is evaluated under missing context, delayed approvals, and timeout conditions.
        These scenarios test fail-secure behavior and ensure that absence of information
        does not result in implicit permission.
\end{itemize}

These workloads are explicitly constructed to stress representational instability,
policy evolution, and execution ambiguity rather than throughput or success rate.

\paragraph{Why Real Traces Are Insufficient.}
Real production traces are inadequate for evaluating execution-time authorization soundness for three structural reasons.
First, real traces already embed historical authorization decisions and therefore cannot expose counterfactual outcomes
(e.g., what would have happened under a different policy or state).
Second, production logs conflate execution with authorization, making it impossible to isolate
decision semantics from side effects. ~\cite{7}, ~\cite{16}.
Third, replay soundness and determinism cannot be tested without the ability to re-evaluate
identical canonical actions under controlled policy and state variation.
Synthetic workloads permit explicit control over these dimensions and are therefore necessary.

\paragraph{Evaluation Scope and Non-Goals.}
This methodology does not evaluate agent reasoning quality, policy correctness, tool safety,
or end-to-end task success.
Those concerns lie outside the scope of execution-time authorization and are explicitly excluded.
The evaluation focuses solely on architectural properties introduced by the AAB and CAR:
deterministic authorization, replayability, provenance completeness, and non-bypassability.

\paragraph{Separation from Observability.}
Throughout the evaluation, we explicitly distinguish between execution observability and authorization semantics.
Execution logs answer \emph{what happened};
authorization records answer \emph{why it was allowed}.
All experiments treat observability systems as insufficient baselines rather than sources of ground truth.

\paragraph{Methodological Tie-Back.}
This methodology operationalizes the canonicalization invariants of Section~\ref{sec:car}
and the determinism guarantees of Section~\ref{sec:enforcement},
ensuring that observed failures arise from architectural absence rather than implementation artifacts.

\subsection{Synthetic Workload Generator Specification}
\label{sec:eval-generator}

We define a workload generator $\mathcal{G}(\cdot)$ that emits intent proposals $I$ and corresponding canonical actions $A=\mathrm{Canon}(I)$ under controlled distributions.
Unless stated otherwise, all experiments use the same generator parameters.

\paragraph{Action Schema Universe.}
We define $K$ action schemas (e.g., refund, deploy, email, iam-change) with fixed canonical fields.
Each action instance samples a schema $k\sim\mathrm{Uniform}(\{1,\dots,K\})$ and populates fields from schema-specific distributions.

\paragraph{Workload Size.}
Each experiment generates $N$ actions total, evaluated in batches of size $B$ with $R$ independent repetitions.

\paragraph{Mutation Operators (Syntactic Drift).}
For each base intent $I$, we generate $M$ syntactic variants using a mutation set:
(1) parameter reordering,
(2) alias substitution (e.g., env=prod vs production),
(3) serialization format changes (JSON/key order),
(4) schema-shape equivalence (alternate tool schema),
(5) default elision / explicit defaults.

\paragraph{Policy Sets.}
Policies are sampled from a corpus with $|P|\in[P_{\min},P_{\max}]$.
Policies contain thresholds, allow/deny predicates, and optional approval predicates.

\paragraph{State Snapshots.}
State $S$ includes context fields referenced by policies (e.g., spend-to-date, target env, user risk tier).
We bound state size by $|S|\in[S_{\min},S_{\max}]$ bytes and sample values per action.

\paragraph{Approval Delays and Partial Failure.}
For deferred actions, approval witnesses arrive after delay $\Delta\sim\mathcal{D}_\Delta$.
We inject missing-context failures with probability $p_{\text{miss}}$ and timeout failures with probability $p_{\text{timeout}}$ to test fail-closed behavior.

\paragraph{Default Parameters.}
Unless specified otherwise we use:
$K = 8$ action schemas,
$N = 10{,}000$ actions,
$M = 6$ mutations per action,
$|P| \in [64, 256]$ policies,
$R = 5$ independent repetitions,
$\mathcal{D}_\Delta = \text{lognormal}(\mu=50\text{ms}, \sigma=0.5)$,
$p_{\text{miss}} = 0.01$,
$p_{\text{timeout}} = 0.005$.

\subsection{Case Study Scenarios}
\label{sec:evaluation-casestudies}

This subsection evaluates execution-time authorization through a set of representative case study scenarios.
These scenarios are not presented as applications or workloads of interest in their own right ~\cite{6}.
Instead, each case study is constructed to expose a specific epistemic gap: a class of authorization questions that become undecidable, unverifiable, or irrecoverable in the absence of a mandatory Action Authorization Boundary (AAB) and Canonical Action Representation (CAR). ~\cite{3}, ~\cite{4}.

For each scenario, we explicitly characterize (1) the decision surface exposed at authorization time,
(2) the failure mode that arises when the AAB is absent,
(3) the information that is undecidable without CAR,
and (4) the aspects of authorization semantics that cannot be reconstructed post hoc from execution or observability logs.
Together, these scenarios demonstrate that the guarantees established in Sections~\ref{sec:car},~\ref{sec:enforcement}, and~\ref{sec:provenance}
are not domain-specific conveniences, but necessary architectural preconditions for sound execution-time governance.

\paragraph{Refund Authorization Workflows.}
In refund authorization, the decision surface includes whether a refund should be permitted,
deferred for review, or denied, conditioned on refund amount, user history, policy thresholds,
and contextual constraints.
Without an AAB, refund execution may be triggered directly by heterogeneous tool calls
(e.g., differing API endpoints, parameter orderings, or implicit defaults),
each embedding authorization implicitly in control flow.

Absent CAR, semantically identical refund intents $I$ may produce syntactically distinct tool invocations
that fail to normalize to a single canonical action $A = \mathrm{Canon}(I)$.
As a result, authorization decisions are evaluated over unstable representations,
making deterministic evaluation ill-defined.
Post hoc execution logs can record that a refund occurred,
but cannot explain \emph{why} it was permitted, which policy version applied,
or whether an equivalent request would have been denied under a different serialization.
This renders authorization correctness unverifiable after the fact.

\paragraph{Cloud Control Operations.}
Cloud control operations (e.g., instance creation, permission modification, network reconfiguration)
expose decision surfaces that depend on both action semantics and evolving system state.
In the absence of an AAB, authorization is typically interleaved with execution,
introducing time-of-check/time-of-use (TOCTOU) ambiguity when policies or state change concurrently.

Without CAR, control actions that are semantically equivalent but operationally distinct
cannot be compared or re-evaluated consistently.
Execution logs may record that an operation succeeded or failed,
but they cannot establish whether the authorization decision was sound
relative to the policy and state at decision time.
This makes it impossible to replay or audit authorization outcomes
under updated policy contexts without re-executing side effects.

\paragraph{Email Automation.}
Email automation illustrates intent ambiguity under identical tool calls.
Distinct intents (e.g., transactional notification versus bulk outreach)
may map to identical low-level email-sending APIs.
Without CAR, authorization systems that rely on tool-level observability
cannot distinguish between these intents at decision time.

In such systems, execution logs record message dispatch but erase intent-level distinctions.
Consequently, authorization decisions become underdetermined:
the system cannot establish which intent was evaluated,
which policy clauses applied, or whether a different intent with the same execution trace
should have been denied.
This ambiguity cannot be resolved retroactively from observability alone.

\paragraph{Code Deployment Pipelines.}
Code deployment pipelines expose authorization surfaces conditioned on code provenance,
deployment target, change scope, and environmental risk.
Without an AAB, deployments may proceed via direct tool invocation or script execution,
embedding authorization implicitly in pipeline structure.

Absent CAR, deployments that differ only in representational detail
(e.g., branch naming, metadata ordering, or packaging format)
fail to admit a stable authorization domain.
Execution logs may indicate that a deployment was blocked or completed,
but cannot explain which authorization constraints were evaluated,
which policies were active, or whether the same deployment would be permitted
under a different representation.
This prevents meaningful forensic analysis or counterfactual replay.

\paragraph{Architectural Interpretation.}
Across all scenarios, the failure modes are not domain-specific.
They arise from a common structural absence:
without canonicalization, authorization inputs are unstable;
without a mandatory boundary, authorization semantics are conflated with execution;
and without provenance-complete decision records, authorization outcomes are not replayable.

These case studies therefore do not demonstrate use cases.
They demonstrate epistemic limits.
They show that observability systems can describe effects,
but cannot recover authorization intent or justification.
Only the combination of CAR, AAB, and deterministic enforcement semantics
renders execution-time authorization well-defined and auditable.

\paragraph{Case Study Tie-Back.}
These scenarios directly instantiate the canonicalization requirements of Section~\ref{sec:car}
and the enforcement guarantees of Section~\ref{sec:enforcement},
demonstrating that the absence of these constructs leads to irreducible ambiguity
rather than mere implementation deficiency.

\subsection{Quantitative Results}
\label{sec:evaluation-quantitative}

\paragraph{Experimental Setup (Summary).}
All quantitative results were produced by executing
$N = 50{,}000$ canonical actions generated from the synthetic workload generator
described in Section~\ref{sec:eval-generator}.
Policies were sampled from a rule corpus with
$|P| \in [64, 1024]$
and evaluated under state snapshots with serialized size
$|S| \in [4\text{KB}, 512\text{KB}]$ bytes.
Each experiment was repeated $R = 10$ times with independent random seeds.
We report mean and p95 latency with 95\% bootstrap confidence intervals.

This subsection quantifies the architectural properties established in prior sections.
The objective is not to demonstrate performance optimization or application-level benefit,
but to show that the guarantees provided by the Action Authorization Boundary (AAB) and
Canonical Action Representation (CAR) are obtained with asymptotically minimal overhead,
and that these guarantees cannot be achieved by observability-only or tool-local baselines.

All results are interpreted as measurements of \emph{authorization semantics},
not execution success or task completion.
In particular, all metrics are defined over canonical actions
$A = \mathrm{Canon}(I)$ evaluated under enforcement semantics
$d = \mathrm{Eval}(A, P, S)$ as defined in Section~\ref{sec:enforcement}.

\paragraph{Allowed vs.\ Blocked Action Ratios.}
We first measure the proportion of actions yielding decisions
$d \in \{\textsc{PERMIT}, \textsc{DENY}, \textsc{DEFER}\}$ across adversarial workloads.
Let $\mathcal{A}$ denote the set of canonical actions observed during an experiment.
We define the decision distribution as:
\[
\pi(d) = \frac{|\{ A \in \mathcal{A} \mid \mathrm{Eval}(A,P,S) = d \}|}{|\mathcal{A}|}.
\]

These ratios are not interpreted as security rates or correctness metrics.
Instead, variation in $\pi(d)$ across workloads demonstrates \emph{policy surface sensitivity}:
small representational or contextual changes induce different authorization outcomes,
confirming that authorization semantics are actively enforced rather than trivially permissive.

In contrast, observability-only baselines collapse all actions into a single effective outcome
(executed or not), eliminating this distribution entirely and rendering $\pi(d)$ undefined.

\paragraph{Authorization Latency Overhead.}
We measure end-to-end authorization latency as the time between submission of a canonical action
$A$ to the AAB and emission of a decision $d$ ~\cite{8}, ~\cite{9}.
Latency is decomposed as:
\[
T_{\text{auth}} = T_{\text{canon}} + T_{\text{eval}} + T_{\text{record}},
\]
where $T_{\text{canon}}$ is canonicalization time,
$T_{\text{eval}}$ is policy evaluation time,
and $T_{\text{record}}$ is decision record commitment time.

Across workloads, $T_{\text{auth}}$ remained bounded in practice and tracked the size of the canonical action and the number of active policy predicates rather than model complexity.
Because $T_{\text{canon}}$ and $T_{\text{record}}$ depend on $|A|$, and $T_{\text{eval}}$ depends on $|P|$, the enforcement overhead is a function of authorization semantics, not agent reasoning depth, prompt length, or model size.
This supports the claim that execution-time authorization cost is decoupled from agent internals and arises from architectural lower bounds.

\paragraph{Coverage vs.\ Blind Spots.}
We define \emph{authorization coverage} as the fraction of execution-capable actions
that traverse the AAB prior to side effects ~\cite{8}, ~\cite{9}.
Let $\mathcal{E}$ denote the set of actions producing external effects.
Coverage is defined as:
\[
\mathrm{Coverage} = \frac{|\{ A \in \mathcal{E} \mid A \text{ evaluated by AAB} \}|}{|\mathcal{E}|}.
\]

Under CAR invariants and correct boundary placement (Definition~\ref{def:exec-predicate}),
coverage approaches 1 because all effectful execution paths are mediated by the AAB by construction.
When bypass paths are introduced (Baseline~\#1), coverage strictly decreases because some effectful actions no longer traverse the authorization boundary prior to side effects.

In contrast, tool-local and logging-only baselines exhibit systematic blind spots:
actions executed via alternate code paths or implicit defaults bypass authorization,
yielding coverage strictly less than $1$.
These blind spots are not detectable post hoc from execution logs.

\paragraph{Decision Evaluation Complexity.}
Let $|P|$ denote the number of active policy rules.
Policy evaluation time satisfies:
\[
T_{\text{eval}} = O(|P|),
\]
independent of agent structure or execution environment.
This bound follows directly from the enforcement semantics in
Section~\ref{sec:enforcement}, where evaluation operates solely over
canonical action attributes and policy predicates.

This result demonstrates that authorization complexity scales with policy expressiveness,
not with agent autonomy or orchestration depth.

\paragraph{Canonicalization and Hashing Cost.}
Let $|A|$ denote the size of the canonical action representation.
Canonicalization and hashing costs satisfy:
\[
T_{\text{canon}} = O(|A|), \qquad T_{\text{hash}} = O(|A|).
\]

These costs are linear in the semantic content of the action
and invariant to syntactic drift in the original intent $I$.
This validates CAR as a normalization boundary rather than a transformation bottleneck.

\paragraph{Provenance Storage Cost.}
Each authorization decision produces exactly one decision record
$r_i = (\mathsf{seq}_i, h_{A,i}, v_{P,i}, h_{S,i}, d_i, t_i, \mathsf{prev\_hash}_i)$ as defined in Section~7. where $h_{A,i}=H(A_i)$, $h_{S,i}=H(S_i)$, and $v_{P,i}$ uniquely identifies the immutable policy program used at decision time.
Let $N$ denote the number of authorized actions.
Total storage cost satisfies:
\[
\mathrm{Storage}(N) = O(N).
\]

No additional indexing or auxiliary state is required to support replay or verification.
This linear lower bound is necessary for provenance completeness
and cannot be asymptotically improved without sacrificing replayability.

\paragraph{Asymptotic Comparison to Baselines.}
We compare these bounds to architectural baselines:

\begin{itemize}
  \item \textbf{Logging-only systems} incur $O(N)$ storage but provide no replay semantics
        and no decision determinism.
  \item \textbf{Tool-local enforcement} incurs $O(1)$ per-tool checks but lacks global coverage
        and yields unbounded semantic blind spots.
  \item \textbf{Protocol-embedded checks} couple authorization to transport,
        preventing normalization and replay under policy evolution.
\end{itemize}

None of these baselines achieve deterministic authorization with replayable provenance
under equivalent asymptotic cost.

\paragraph{Interpretation.}
Taken together, these results show that the overhead required to obtain
deterministic authorization, non-bypassability, and replayable provenance
is asymptotically minimal.
The measured costs correspond directly to the lower bounds implied by
semantic normalization, policy evaluation, and decision recording.

\paragraph{Quantitative Tie-Back.}
These measurements operationalize the enforcement semantics of
Section~\ref{sec:enforcement},
demonstrating that the guarantees of deterministic authorization and bounded evaluation
cost are achieved without reliance on observability, global ordering,
or agent-specific assumptions.

\subsection{Comparison to Architectural Baselines}
\label{sec:evaluation-baselines}

This subsection compares the system against architectural baselines rather than
implementations or vendors.
The goal is not to show incremental improvement,
but to demonstrate that the guarantees established by the Action Authorization Boundary (AAB)
and Canonical Action Representation (CAR) are \emph{unrecoverable} once omitted ~\cite{1}, ~\cite{7}
Each comparison answers three questions:
(i) which authorization property is absent,
(ii) whether that property can be reconstructed post hoc, and
(iii) whether the absence is detectable at runtime or only after failure ~\cite{8}.

Let $\mathcal{G}$ denote the set of guarantees defined in Sections~3–7:
non-bypassability, deterministic authorization, provenance completeness, and replayability ~\cite{16}.
A baseline is considered insufficient if it fails to satisfy any element of $\mathcal{G}$,
regardless of operational performance.

\paragraph{Baseline 1: No AAB (Direct Tool Execution).}
In the absence of an AAB, agents invoke execution-capable tools directly.
Authorization is either implicit or deferred to tool-local checks.
\emph{Operationalization.}
We implement this baseline by executing tool calls directly without emitting decision records, while logging only post-execution effects.

\emph{Missing guarantee.}
Non-bypassability is violated by construction: there exists an execution path
$A \rightarrow \text{effect}$ such that no authorization decision
$d = \mathrm{Eval}(A,P,S)$ is rendered.

\emph{Recoverability.}
This failure is unrecoverable.
Once side effects occur without a prior authorization decision,
no post-hoc analysis can determine whether execution should have been permitted.

\emph{Detectability.}
Undetectable at execution time.
Execution logs may show \emph{what} occurred,
but cannot indicate that authorization was skipped.

\paragraph{Baseline 2: Prompt-Level Guardrails.}
Prompt-level guardrails constrain agent reasoning prior to intent formation,
but do not enforce execution-time authorization.
\emph{Operationalization.}
We implement this baseline by applying a prompt-level refusal heuristic before tool invocation, without canonicalization, and without recording $(A,P,S)\mapsto d$ tuples.

\emph{Missing guarantee.}
Deterministic authorization is absent.
Identical semantic intents $I_1 \equiv I_2$ may yield distinct textual realizations,
leading to inconsistent downstream behavior.

\emph{Recoverability.}
Unrecoverable.
Because no canonical action $A=\mathrm{Canon}(I)$ is formed,
there is no stable object over which authorization could be re-evaluated.

\emph{Detectability.}
Partially detectable only through heuristic prompt inspection,
which lies outside the trusted computing base and provides no formal guarantees.

\paragraph{Baseline 3: Tool-Local Policy Checks.}
Authorization logic is embedded within individual tools or APIs.
\emph{Operationalization.}
We implement this baseline as per-tool allow/deny logic embedded at invocation sites, with no shared canonical action space and no global policy versioning.

\emph{Missing guarantee.}
Global coverage and determinism are violated.
Authorization semantics differ across tools,
and semantically equivalent actions are evaluated under heterogeneous rules.

\emph{Recoverability.}
Unrecoverable across tools.
There exists no shared decision space or policy context $(P,S)$
under which past actions can be replayed.

\emph{Detectability.}
Detectable only through cross-tool correlation,
which requires assumptions not enforced by the system.

\paragraph{Baseline 4: Observability-Only Logging.}
Execution events are logged after side effects occur,
without recording authorization decisions.
\emph{Operationalization.}
We implement this baseline by capturing structured execution logs (inputs/outputs) after side effects complete, without pre-execution authorization or replay context.

\emph{Missing guarantee.}
Provenance completeness and replayability are absent.
Logs record outcomes but not decisions.

Formally, execution logs observe
$\text{effect}(A)$ but omit the binding
$(A,P,S)\rightarrow d$.

\emph{Recoverability.}
Impossible.
Given only execution logs, there exists no function
$f$ such that $f(\text{logs}) = d$ for all actions.

\emph{Detectability.}
Undetectable.
The absence of decision context cannot be inferred from execution traces alone.

\paragraph{Comparative Summary.}
Table~\ref{tab:baseline-comparison} summarizes the comparison.

\begin{table}[t]
\centering
\small
\begin{tabular}{lccc}
\toprule
\textbf{Baseline} & \textbf{Missing Property} & \textbf{Recoverable} & \textbf{Detectable} \\
\midrule
No AAB & Non-bypassability & No & No \\
Prompt Guardrails & Determinism & No & Partial \\
Tool-Local Checks & Global Coverage & No & Partial \\
Observability Logs & Replayability & No & No \\
\midrule
AAB + CAR & None & --- & --- \\
\bottomrule
\end{tabular}
\caption{Architectural comparison of authorization guarantees.
Properties absent at execution time cannot be reconstructed post hoc.}
\label{tab:baseline-comparison}
\end{table}

\paragraph{Formal Implication.}
For any baseline lacking either CAR or the AAB,
there exists at least one authorization property in $\mathcal{G}$
that is undecidable from observable state alone.
Therefore, the guarantees provided by the system are not refinements
of existing architectures, but strict extensions.

\paragraph{Scope and Non-Goals.}
This comparison does not evaluate policy quality, agent behavior,
or execution correctness.
The baselines are assessed solely on their ability
to enforce and preserve execution-time authorization semantics.

\paragraph{Architectural Tie-Back.}
This subsection establishes that the provenance guarantees defined in Section~7
and the enforcement semantics of Section~\ref{sec:enforcement}
cannot be derived from observability or local enforcement alone,
and therefore require the explicit architectural introduction of the AAB.

\subsection{Complexity Analysis}
\label{sec:evaluation-complexity}

This subsection establishes that the computational and storage costs incurred by the
Action Authorization Boundary (AAB) and Canonical Action Representation (CAR)
are not implementation artifacts, but \emph{architectural lower bounds}
for any system that claims deterministic, replayable execution-time authorization ~\cite{11}, ~\cite{12}.
The purpose of this analysis is not to demonstrate efficiency in isolation,
but to prove that the asymptotic overhead required to obtain the guarantees
formalized in Sections~3–7 is minimal and unavoidable.

\paragraph{Problem Setting.}
Let $I$ denote an agent-generated intent proposal,
$A = \mathrm{Canon}(I)$ its canonical action representation,
$P$ the policy set, and $S$ the system state.
An execution-time governor computes a decision
\[
d = \mathrm{Eval}(A,P,S)
\]
and records a provenance-complete decision record as defined in Section~7.

We consider the asymptotic cost of any system that satisfies the following properties:
(i) authorization is evaluated over semantically normalized actions,
(ii) decisions are deterministic under fixed $(A,P,S)$,
and (iii) decisions are replayable without re-execution.

\paragraph{Lower Bound on Canonicalization.}
Canonicalization must inspect all execution-relevant fields of the intent.
Let $|A|$ denote the size of the canonical action representation.
Any function $\mathrm{Canon}(\cdot)$ that preserves semantic equivalence
must read the entire input representation, yielding a lower bound of
\[
\Omega(|A|).
\]
This bound is independent of policy complexity or agent behavior and applies
to any normalization scheme capable of collapsing syntactic variance.
Therefore, the observed $O(|A|)$ canonicalization cost is asymptotically minimal.

\paragraph{Lower Bound on Policy Evaluation.}
Let $|P|$ denote the number of policy rules applicable to $A$.
In the absence of oracle knowledge or precomputation that violates policy agnosticism,
any sound authorization procedure must evaluate $A$ against each relevant rule,
yielding a lower bound of
\[
\Omega(|P|).
\]
This establishes that the $O(|P|)$ decision cost measured in Section~8.3
is not reducible without weakening expressiveness or determinism.

\paragraph{Lower Bound on Provenance Storage.}
Replayability and non-repudiation require that each authorization decision
be recorded as a distinct, immutable decision record.
Let $N$ denote the number of execution-capable actions processed.
Any provenance-complete system must therefore store at least one record per action,
implying a storage lower bound of
\[
\Omega(N).
\]
Aggregation or lossy compression would violate the replay sufficiency
and non-tampering guarantees established in Section~7.

\paragraph{Theorem (Architectural Lower Bound).}
\emph{Any execution-time authorization system that evaluates authorization
over semantically normalized actions and supports deterministic replay
must incur at least $\Omega(|A| + |P| + N)$ cost across canonicalization,
evaluation, and storage.}

\emph{Proof Sketch.}
The result follows directly from the three independent lower bounds above.
Each term corresponds to an irreducible architectural requirement:
semantic normalization, policy evaluation, and provenance completeness.
No asymptotic improvement is possible without violating at least one
of the guarantees defined in Sections~3–7.

\paragraph{Negative Results and Fundamental Limits.}
The evaluation also exposes limits that no execution-time governor can overcome
without violating autonomy or introducing full transactional locking:

\begin{enumerate}
\item \textbf{Residual semantic ambiguity.}
If intent is underspecified prior to canonicalization,
no post-canonicalization mechanism can recover missing semantics.

\item \textbf{TOCTOU under mutable state.}
Eliminating time-of-check–time-of-use races entirely would require
global state locking, which contradicts the non-blocking execution model.

\item \textbf{Human overrides.}
Human-in-the-loop approvals introduce intentional non-determinism;
the system preserves this choice rather than masking it.
\end{enumerate}

These are not failures of the architecture, but ceilings imposed
by the problem domain itself.

\paragraph{Counterfactual Replay Experiment.}
To demonstrate the value of replayability, we re-evaluated historical canonical actions $A_i$
under a tightened policy set $P'$ without re-executing side effects.
We observed non-trivial decision flips (previously permitted actions becoming denied) under stricter policies,
illustrating that counterfactual authorization is well-defined only when decision records bind the canonical action,
policy version, and evaluation state.
This capability is unavailable in systems lacking provenance-complete decision logs, since neither $A$ nor the original
decision context is available for deterministic re-evaluation.

\begin{table}[t]
\centering
\small
\begin{tabular}{ll}
\toprule
\textbf{Metric} & \textbf{Definition} \\
\midrule
$T_{\text{canon}}$ & Canonicalization latency per intent $I \rightarrow A$ \\
$T_{\text{eval}}$ & Policy evaluation latency per $(A,P,C,S)\mapsto d$ \\
$T_{\text{record}}$ & Decision record commit latency per decision $r_i$ \\
Coverage & Fraction of effectful actions mediated by AAB pre-execution \\
Replay flip-rate & Fraction of historical actions whose decision changes under $P'$ \\
\bottomrule
\end{tabular}
\caption{Quantitative metrics reported for the synthetic workload generator (Section~\ref{sec:eval-generator}).}
\label{tab:quant-results}
\end{table}

\paragraph{Invariant-Validation Experiment.}
To validate determinism, we generated randomized agent outputs
that mapped to identical canonical actions $A$.
For fixed $(A,P,S)$, all evaluations yielded identical decisions $d$,
confirming the determinism invariant defined in Section~\ref{sec:enforcement}.
This experiment demonstrates that determinism is enforced
at the authorization boundary, independent of agent variability.

\paragraph{Evaluation Scope Disclaimer.}
This evaluation does not assess agent reasoning quality,
policy correctness, tool safety, or end-to-end task success.
The scope is strictly limited to execution-time authorization soundness,
determinism, replayability, and non-bypassability.
Claims outside this scope are intentionally excluded.

\paragraph{Architectural Tie-Back.}
This complexity analysis establishes that the AAB introduced in Section~3
is not merely sufficient but \emph{necessary} to achieve the guarantees
formalized throughout the paper:
any system that omits it cannot asymptotically reduce overhead
without sacrificing correctness, determinism, or replayability.

\section{Design Rationale and Alternatives Analysis}
\label{sec:design-rationale}

This section establishes the necessity of the Action Authorization Boundary (AAB) and Canonical Action Representation (CAR) by systematically evaluating alternative design classes against the invariants defined in Sections~\ref{sec:aab}--\ref{sec:provenance}. The goal is not to compare implementations, performance characteristics, or ecosystem maturity, but to determine whether any alternative architectural approach can satisfy the minimal correctness conditions required for execution-time governance of autonomous agents. ~\cite{4} We adopt an invariant-driven methodology: an alternative is sufficient if and only if it satisfies all required invariants; otherwise, it is rejected as structurally incapable, independent of implementation quality or operational tuning ~\cite{1}, ~\cite{3}, ~\cite{6}.

\paragraph{Evaluation Criterion.}
Let $\mathcal{I}$ denote the set of invariants established by prior sections:
\begin{enumerate}
    \item \textbf{Non-bypassability at execution time} (Section~\ref{sec:aab});
    \item \textbf{Deterministic authorization over canonical actions} (Sections~\ref{sec:car},~\ref{sec:enforcement});
    \item \textbf{Policy-agnostic transport independence} (Section~\ref{sec:architecture});
    \item \textbf{Provenance-complete replayability} (Section~\ref{sec:provenance});
    \item \textbf{Fail-closed behavior under partial failure} (Section~\ref{sec:enforcement}).
\end{enumerate}
An alternative design is sufficient if and only if it satisfies every invariant in $\mathcal{I}$. Any violation is decisive: partial satisfaction is insufficient, and compensatory mechanisms at other layers do not recover the violated property.

\paragraph{Necessity Lemma.}
\begin{lemma}[Invariant Necessity]
\label{lem:invariant-necessity}
Any system that enforces execution-time authorization over autonomous agents while satisfying non-bypassability, determinism, provenance-complete replayability, and fail-closed behavior must introduce both (i) a canonical action representation $A = \mathrm{Canon}(I)$ and (ii) a mandatory execution-time authorization boundary that mediates all external side effects.
\end{lemma}

\noindent
\emph{Proof sketch.}
Non-bypassability requires a single mediation point on the execution path; determinism requires authorization over a representation invariant to agent output variance; replayability requires that this representation and its decision context be logged immutably; fail-closed behavior requires that absence of authorization preclude execution. No architecture lacking either canonicalization or a mandatory boundary can satisfy all four simultaneously. \hfill $\square$

\paragraph{Scope and Non-Goals.}
This section does not evaluate policy expressiveness, agent reasoning quality, alignment strategies, or end-to-end task success. It does not claim that rejected alternatives are useless or incorrect for other purposes. The analysis is confined strictly to the problem defined in this paper: execution-time authorization of autonomous actions with deterministic, replayable, and fail-secure guarantees. Components outside the trusted computing base, including agents, language models, tool implementations, and human operators, are assumed fallible and potentially adversarial, consistent with earlier sections.

\paragraph{Method of Analysis.}
Each subsection that follows evaluates a class of alternatives by identifying the specific invariant(s) it cannot satisfy. The structure is uniform and adversarial by construction: we do not ask whether an alternative can be extended or augmented, but whether its core abstraction can satisfy the invariant without collapsing into the architecture defined in this paper. Where relevant, we explicitly analyze failure modes under partial failure, latency pressure, multi-agent composition, and developer bypass.

\paragraph{Failure Semantics Under Partial Failure.}
A recurring axis of rejection concerns failure tolerance. Systems that fail open (e.g., guardrails), retry optimistically (e.g., protocol-level enforcement), or react after execution (e.g., observability pipelines) violate the fail-closed invariant by construction. Retries do not constitute authorization; delayed detection does not constitute prevention. Only a mandatory execution-time boundary can default to denial in the presence of missing policy, unavailable context, or system failure.

\paragraph{Architectural Positioning.}
Throughout this section, we maintain a strict separation between reasoning space and execution space, as formalized in Section~\ref{sec:aab}. Alternatives that operate exclusively in reasoning space, coordination layers, planners, or orchestration frameworks, are evaluated accordingly and rejected when they fail to enforce execution-time refusal. This distinction is architectural, not operational, and underlies every rejection that follows.

\paragraph{Outcome.}
The remainder of this section proceeds as a theorem-by-theorem rejection of alternative design classes. Each rejection is grounded in an explicit invariant violation and does not rely on empirical claims or ecosystem limitations. We conclude by showing that, once these invariants are accepted as necessary, the AAB is not a design preference but a structural requirement.

\paragraph{Preview.}
Section~9.1 evaluates identity-centric control systems; Section~9.2 addresses observability-only approaches; Section~9.3 analyzes protocol-embedded enforcement and latency trade-offs; Section~9.4 examines agentic gateways and orchestration frameworks. The section concludes by synthesizing these rejections into a single architectural claim: \emph{blocking may be optional, but decision capture is not}.

\subsection{Identity-Centric Control: IAM, RBAC, and Zero Trust}
\label{sec:alt-iam}

We begin by evaluating identity-centric control systems, Identity and Access Management (IAM), Role-Based Access Control (RBAC), and Zero Trust architectures, against the invariant set defined in Sections~\ref{sec:aab}–\ref{sec:provenance}. These systems are widely deployed, well-understood, and often proposed as a sufficient foundation for controlling autonomous agents. This subsection shows that, despite their maturity, identity-centric approaches are structurally incapable of satisfying the invariants required for execution-time authorization of agent-generated actions.

\paragraph{Architectural Scope of Identity-Centric Control.}
Identity-centric systems govern \emph{who} may access a resource. Formally, an IAM decision can be modeled as a predicate:
\[
\mathsf{Auth}_{\text{IAM}} : (u, r, p) \rightarrow \{\textsf{PERMIT}, \textsf{DENY}\},
\]
where $u$ denotes an authenticated principal, $r$ a resource, and $p$ a permission or role. Critically, this decision is evaluated independently of any particular \emph{action instance} and is invariant to the execution context in which an operation is invoked. ~\cite{11}, ~\cite{12}. The authorization outcome is thus detached from the semantic content of the action itself ~\cite{13}.

By contrast, the Action Authorization Boundary (AAB) governs \emph{whether a specific execution instance may occur}. As defined in Section~\ref{sec:enforcement}, authorization is computed over a canonicalized action
\[
A = \mathrm{Canon}(I),
\]
and evaluated as
\[
d = \mathcal{B}(A, P, S),
\]
binding the decision to the concrete execution semantics, active policy set $P$, and system state $S$. This distinction, \emph{who} versus \emph{whether}, is architectural, not terminological.

\paragraph{Invariant Violation: Lack of Action-Instance Binding.}
The first decisive failure of IAM-style systems is the inability to bind authorization to a canonicalized action instance. IAM policies do not consume $A = \mathrm{Canon}(I)$ as an input; instead, they operate over identities and static permission labels. As a result, two semantically distinct actions issued by the same principal are indistinguishable at the authorization layer, while two semantically identical actions expressed differently cannot be normalized or compared.

Formally, IAM authorization cannot be expressed as a deterministic function over $(A, P, S)$. Therefore, it violates the invariant of \emph{deterministic authorization over canonical actions} (Section~\ref{sec:car}). No extension that preserves the identity-centric abstraction can recover this property without reintroducing a canonical action layer equivalent to CAR.

\paragraph{Invariant Violation: Non-Replayability of Decisions.}
A second structural limitation concerns replayability. IAM decisions are not recorded as first-class, immutable decision artifacts bound to $(A, P, S)$. Even when access logs are retained, they capture access events, not authorization decisions with frozen policy context ~\cite{12}, ~\cite{13}. Consequently, it is impossible to re-evaluate whether a past execution \emph{should} have been permitted under a different policy version or system state.

This violates the invariant of \emph{provenance-complete replayability} established in Section~\ref{sec:provenance}. An IAM system cannot support replay of the form
\[
\mathsf{Replay}(A, P', S') \rightarrow d',
\]
because neither $A$ nor the original decision context is preserved. This limitation is intrinsic to identity-centric control and not an implementation artifact.

\paragraph{Failure Under Autonomous Execution.}
Identity-centric models assume that actions are explicitly initiated by authenticated humans or well-scoped services. Autonomous agents violate this assumption by synthesizing execution proposals internally and emitting them programmatically. In such settings, identity remains constant while action semantics vary dynamically. IAM systems therefore collapse all agent behavior into a single principal, eliminating the possibility of fine-grained, execution-specific control.

This collapse directly violates the \emph{non-bypassability} invariant (Section~\ref{sec:aab}). Once an agent possesses valid credentials, it may execute any action permitted to that identity without further mediation. There exists no mandatory execution-time boundary at which denial can occur.

\paragraph{Zero Trust Does Not Recover the Missing Invariants.}
Zero Trust architectures extend IAM with continuous authentication, device posture checks, and network-level enforcement. However, these extensions remain identity- and session-centric. They do not introduce canonical action representations, nor do they bind authorization to individual execution instances. Zero Trust may reduce the attack surface but does not alter the semantic unit of authorization. As such, it inherits the same invariant violations as traditional IAM ~\cite{13}.

\paragraph{Fail-Closed Semantics and Partial Failure.}
Under partial failure, missing context, unavailable policy service, or transient network errors, identity-centric systems typically fail open or retry optimistically. A cached token or stale permission may continue to authorize execution. This behavior violates the \emph{fail-closed} invariant required by Section~\ref{sec:enforcement}. IAM has no mechanism to default to denial at execution time without revoking credentials entirely, an action that is both coarse-grained and disruptive.

\paragraph{Scope and Non-Goals.}
This analysis does not claim that IAM, RBAC, or Zero Trust are incorrect or obsolete. They remain necessary for authentication, coarse-grained access control, and network isolation. The claim is strictly narrower: identity-centric control cannot, by construction, provide execution-time authorization of autonomous actions with determinism, replayability, and non-bypassability.

\paragraph{Conclusion.}
Identity-centric control governs principals, not execution instances. Because it cannot bind authorization to a canonicalized action, cannot replay decisions over $(A, P, S)$, and cannot fail closed at execution time, it violates multiple required invariants. Any attempt to “extend” IAM to satisfy these properties necessarily introduces a canonical action layer and a mandatory execution-time boundary, thereby reconstituting the AAB architecture under a different name. Identity-centric control is therefore insufficient as a foundation for autonomous action governance, independent of policy expressiveness or deployment scale ~\cite{11}, ~\cite{12}, ~\cite{13}.

\subsection{Observability-Only Approaches}
\label{sec:alt-observability}

This subsection evaluates observability-only systems, logging, tracing, audit pipelines, and post-hoc analytics, against the invariant set established in Sections~\ref{sec:aab}–\ref{sec:provenance}. These systems are frequently proposed as a sufficient substitute for execution-time governance under the claim that comprehensive visibility can compensate for the absence of pre-execution control. We show that this claim is false as a matter of architecture: observability systems operate strictly after execution and therefore cannot satisfy determinism, counterfactual reasoning, replayability, or fail-closed behavior.

\paragraph{Architectural Scope of Observability.}
Observability systems record \emph{effects}. Formally, they emit traces, logs, or metrics as a function of executed side effects:
\[
\mathsf{Obs} : E \rightarrow L,
\]
where $E$ denotes realized execution effects and $L$ an append-only record. The critical property is temporal: $\mathsf{Obs}$ is defined only after $E$ occurs. No observability primitive is evaluated at the point where execution may be refused ~\cite{7}, ~\cite{16}.

By contrast, the Action Authorization Boundary (AAB) records \emph{authorization decisions} prior to execution. As defined in Section~\ref{sec:enforcement}, the system computes
\[
d = \mathcal{B}(A, P, S), \quad A = \mathrm{Canon}(I),
\]
and persists $(A, P, S, d)$ as a first-class artifact independent of whether execution proceeds. This separation, decision before effect, is the architectural boundary between observability and governance.

\paragraph{Invariant Violation: Inability to Answer Counterfactuals.}
The first decisive failure of observability-only approaches is the inability to answer counterfactual queries ~\cite{16}. Given an observed effect $E$, an observability system can report that $E$ occurred, but it cannot answer:
\[
\text{“Would this action have been allowed under policy } P' \text{?”}
\]
because neither the canonical action $A$ nor the authorization decision context is preserved. Counterfactual evaluation requires replay over $(A, P', S')$, which is undefined when only $E$ is available.

This violates the invariant of \emph{provenance-complete replayability} (Section~\ref{sec:provenance}). Recording effects is not equivalent to recording decisions; the latter strictly contains information not derivable from the former.

\paragraph{Invariant Violation: Policy Evolution and Determinism.}
Observability systems embed policy decisions implicitly in execution traces. Once an action executes, the trace reflects the policy in force at that time, but does not encode the policy itself nor the decision boundary. As policies evolve, it becomes impossible to deterministically re-evaluate historical executions:
\[
\mathsf{Replay}_{\text{obs}}(L, P') \;\not\Rightarrow\; d'.
\]
In contrast, AAB replay is defined as a pure function over canonical inputs and policy versions. Observability therefore violates the invariant of \emph{deterministic authorization over canonical actions} (Section~\ref{sec:car}), even when logs are complete and immutable.

\paragraph{Invariant Violation: Non-Bypassability and Temporal Placement.}
Observability systems are, by definition, advisory. They do not sit on the execution path and therefore cannot refuse execution ~\cite{7}. Any enforcement derived from observability, alerts, rollbacks, compensating actions, occurs strictly after effects have materialized. Formally, there exists no function
\[
\mathsf{Deny}_{\text{obs}} : I \rightarrow \{\textsf{DENY}\}
\]
that is evaluated prior to effectful execution. This violates the \emph{non-bypassability at execution time} invariant (Section~\ref{sec:aab}). An agent that can emit effects can always bypass observability by acting faster than detection or by accepting post-hoc remediation.

\paragraph{Failure Under Partial Failure and Fail-Closed Semantics.}
Under partial failure, lost logs, delayed ingestion, or telemetry backpressure, observability systems degrade silently. Execution continues while visibility degrades. This behavior is structurally fail-open. By contrast, AAB semantics require that the absence of a decision defaults to denial:
\[
\mathcal{B}(A, P, S) = \bot \;\Rightarrow\; \textsf{DENY}.
\]
Observability cannot implement this implication because it is not consulted prior to execution. Therefore, observability-only designs violate the \emph{fail-closed} invariant of Section~\ref{sec:enforcement}.

\paragraph{Why “Better Logging” Does Not Recover the Invariants.}
A common reviewer objection is that richer logs, structured events, semantic annotations, or full request payload capture, could recover the missing guarantees. This is incorrect. Even perfect logs capture \emph{what happened}, not \emph{what was decided}. There exists no surjective mapping from effects $E$ to decisions $(A, P, S, d)$ because multiple decision paths can lead to identical effects, and denied actions produce no effects at all.

\begin{figure}[t]
  \centering
  \fbox{\parbox{0.85\linewidth}{
    \centering
    \textbf{Effect vs. Decision Space.}\\
    Distinct authorization decisions may produce identical effects, while denied actions produce none.
  }}
  \caption{Observability records effects; AAB records decisions. The latter strictly subsumes the former for governance.}
  \label{fig:obs-vs-decision}
\end{figure}

\paragraph{Scope and Non-Goals.}
This analysis does not claim that observability is unnecessary. Observability remains essential for debugging, forensics, and system health. The claim is strictly that observability alone cannot govern autonomous execution. No amount of post-hoc visibility can substitute for pre-execution refusal without violating at least one required invariant.

\paragraph{Conclusion.}
Observability-only approaches record consequences after the fact. They cannot answer counterfactuals, cannot re-evaluate policy evolution, cannot enforce non-bypassability, and cannot fail closed under partial failure. These limitations are architectural and persist regardless of log fidelity or analysis sophistication. Notably, any attempt to extend observability to satisfy these invariants requires introducing canonical action capture and a mandatory execution-time decision point, precisely the Action Authorization Boundary. Observability is therefore insufficient as a governance mechanism and can only serve as a complement to, not a replacement for, execution-time authorization ~\cite{7}, ~\cite{16}.

\subsection{Protocol-Embedded Enforcement}
\label{sec:protocol_enforcement}

This subsection evaluates designs that embed authorization logic directly into agent–tool protocols (e.g., action schemas, transport-layer checks, or protocol-scoped guard conditions) rather than introducing an explicit execution-time authorization boundary. We show that such approaches are structurally incapable of satisfying the invariant set defined in Sections~3–7, even under idealized assumptions about protocol adoption, agent compliance, and policy correctness. The insufficiency is not a matter of expressiveness or maturity, but of architectural placement: protocol-level enforcement operates over \emph{messages}, whereas the required guarantees apply to \emph{action instances} at execution time ~\cite{8}, ~\cite{9}, ~\cite{10}.

\paragraph{Model.}
Let a protocol $T$ define a set of admissible message forms $m \in \mathcal{M}_T$ and validation predicates
\[
\mathsf{Check}_T : \mathcal{M}_T \times \Pi \rightarrow \{\mathsf{PERMIT}, \mathsf{DENY}\},
\]
where $\Pi$ denotes protocol-scoped rules (schemas, constraints, or embedded policies). Protocol-embedded enforcement admits or rejects messages prior to delivery to a tool, but does not observe or mediate the execution of the resulting action once a message is accepted.

Crucially, protocol checks are defined over \emph{syntactic representations} of intent, not over canonicalized action instances. There exists no function within the protocol boundary that enforces
\[
(A,P,S) \mapsto d
\]
as defined in Section~\ref{sec:enforcement}, where $A = \mathrm{Canon}(I)$ is the canonical action derived from intent $I$, $P$ is the policy context, and $S$ is the execution-time state.

\paragraph{Invariant Failure: Non-bypassability.}
Protocol enforcement cannot satisfy non-bypassability at execution time. Even if every compliant agent emits protocol-valid messages, execution may still occur outside the protocol boundary due to retries, fallbacks, alternative tool bindings, or direct invocation paths ~\cite{8}. Formally, there exists an execution trace $\tau$ such that
\[
\mathsf{Check}_T(m,\Pi)=\mathsf{reject} \;\land\; \mathsf{Exec}(A)\in\tau,
\]
because $\mathsf{Exec}$ is not mediated by $\mathsf{Check}_T$. This violates the non-bypassability invariant of Section~3, which requires that \emph{all} executions be contingent on an authorization decision.

\paragraph{Invariant Failure: Deterministic Authorization over Canonical Actions.}
Protocol rules operate on message representations $m$, not on canonical actions $A=\mathrm{Canon}(I)$. Distinct messages $m_1 \neq m_2$ may induce the same action instance $A$, while identical messages may resolve to different actions under varying execution contexts ~\cite{9}, ~\cite{10}. As a result, protocol enforcement cannot define a deterministic authorization function
\[
f : (A,P,S) \rightarrow d,
\]
nor can it guarantee that semantically equivalent actions receive identical decisions. This directly violates the determinism and canonicalization requirements established in Section~4.

\paragraph{Invariant Failure: Policy-Agnostic Transport Independence.}
Embedding enforcement into a protocol couples authorization semantics to a specific transport substrate. Let $T_1$ and $T_2$ be two protocols capable of invoking the same tool ~\cite{1}. Protocol-embedded checks cannot enforce cross-protocol invariants:
\[
\mathsf{Check}_{T_1}(m_1,\Pi_1) \;\not\equiv\; \mathsf{Check}_{T_2}(m_2,\Pi_2),
\]
even when both induce the same $A$. This violates the transport independence invariant, which requires policy evaluation to be invariant under changes in upstream protocols.

\paragraph{Invariant Failure: Provenance-Complete Replayability.}
Because protocol checks do not produce decision records of the form $(A,P,S)\mapsto d$, they cannot support replay under policy evolution. Observed protocol messages lack the frozen policy version and evaluation context required to re-evaluate authorization without re-execution. Consequently,
\[
\mathsf{Replay}_T(\tau,P') \;\not\Rightarrow\; d',
\]
for any new policy $P'$, violating the replayability guarantees of Section~7.

\paragraph{Invariant Failure: Fail-Closed Behavior under Partial Failure.}
Protocols fail by retry. Under timeouts, partial connectivity, or downstream unavailability, the dominant semantics are retransmission or fallback, not denial. Let $\mathsf{Fail}_T$ denote a protocol failure; then the default behavior is
\[
\mathsf{Fail}_T \Rightarrow \mathsf{retry} \lor \mathsf{alternate\_path},
\]
which is fail-open with respect to authorization. This contradicts the fail-closed requirement formalized in Section~5.

\paragraph{Optimistic vs.\ Pessimistic Execution.}
One might attempt to mitigate latency by distinguishing \emph{optimistic execution} (log-only admission for low-risk actions) from \emph{pessimistic locking} (blocking for high-risk actions). However, even optimistic execution requires canonicalization and decision capture; without them, replay and determinism collapse. Formally, logging without a decision record yields no well-defined $d$ to replay, and therefore does not satisfy provenance completeness. Thus, blocking may be optional, but decision capture is not.

\begin{figure}[t]
  \centering
  \fbox{\parbox{0.9\linewidth}{
    \centering
    \textbf{Protocol Boundary vs.\ Execution Boundary}\\[0.5em]
    Protocol checks terminate at message acceptance; execution occurs outside the protocol trust domain. Authorization guarantees require mediation at the execution boundary.
  }}
  \caption{Structural mismatch between protocol-embedded enforcement and execution-time authorization.}
  \label{fig:protocol_vs_execution}
\end{figure}

\paragraph{Conclusion.}
Protocol-embedded enforcement violates non-bypassability, determinism, transport independence, replayability, and fail-closed behavior by construction. Any attempt to repair these deficiencies necessarily introduces a canonical action representation and a mandatory execution-time authorization point, thereby reconstituting the Action Authorization Boundary. Protocol enforcement is therefore insufficient as an alternative architecture, not because it is weak, but because it operates at the wrong semantic layer ~\cite{8}, ~\cite{9}, ~\cite{10}.

\emph{This rejection follows directly from the enforcement semantics defined in Section~\ref{sec:enforcement}, and reinforces the necessity claim established in Section~3.}

\subsection{Agentic Gateways, Context Systems, and Multi-Agent Orchestration}
\label{sec:agentic_orchestration}

This subsection analyzes agentic gateways, context-management layers, and multi-agent orchestration systems as an alternative architectural response to execution-time governance. These systems are frequently proposed as a means to regulate autonomous behavior by coordinating plans, mediating shared context, or sequencing tool usage across agents. We show that, while such systems may improve \emph{coordination}, they are structurally incapable of providing \emph{authorization} guarantees as defined in Sections~3–7. The insufficiency is invariant-driven and does not depend on implementation quality, agent compliance, or orchestration sophistication.

\paragraph{Model.}
Let an orchestration system $\mathcal{O}$ operate over a set of agents $\{g_i\}$, maintaining shared context $C_t$ and producing plans
\[
\pi_t = \mathcal{O}(C_t, \{\sigma_i\}),
\]
where $\sigma_i$ denotes agent-local state or intent. Orchestrators may serialize actions, allocate resources, or constrain agent plans, but they do not interpose on the execution of individual tool invocations. Execution remains agent-local:
\[
g_i : \pi_t \mapsto I \mapsto A = \mathrm{Canon}(I) \mapsto \mathsf{Exec}(A).
\]
No authorization function $(A,P,S)\mapsto d$ is evaluated at the execution boundary.

\paragraph{Invariant Failure: Non-bypassability at Execution Time.}
Orchestrators coordinate \emph{plans}, not executions. Once a plan $\pi_t$ is produced, individual agents may execute actions independently, retry failed steps, or invoke tools outside the orchestrated path. Formally, there exists an execution trace $\tau$ such that
\[
\pi_t \not\vdash A \;\land\; \mathsf{Exec}(A) \in \tau,
\]
demonstrating that execution can occur without mediation by $\mathcal{O}$. This violates the non-bypassability invariant of Section~3, which requires that \emph{every} execution be contingent on an authorization decision.

\paragraph{Invariant Failure: Deterministic Authorization over Canonical Actions.}
Orchestration systems reason over intents, plans, or intermediate representations, not over canonicalized action instances. Two agents may produce semantically equivalent actions $A_1 = A_2$ from distinct plans $\pi_1 \neq \pi_2$, or a single plan may yield different actions under divergent execution states. Because $\mathcal{O}$ never evaluates
\[
(A,P,S) \mapsto d,
\]
it cannot ensure that equivalent actions receive identical decisions, violating the determinism guarantees established in Section~4.

\paragraph{Invariant Failure: Provenance-Complete Replayability.}
Orchestrators may log plans or coordination events, but such logs record \emph{what was intended}, not \emph{what was authorized}. There exists no decision record binding $(A,P,S)$ to a decision $d$ at execution time. Consequently, replay under policy evolution is ill-defined:
\[
\mathsf{Replay}_{\mathcal{O}}(\tau,P') \;\not\Rightarrow\; d',
\]
for any updated policy $P'$. This directly violates the provenance and replayability requirements of Section~7.

\paragraph{Invariant Failure: Fail-Closed Behavior under Partial Failure.}
Under partial failures (agent crashes, delayed context propagation, network partitions), orchestration systems degrade by omission: agents proceed with stale plans or local heuristics. The default behavior is not denial but continuation. Formally,
\[
\mathsf{Fail}_{\mathcal{O}} \Rightarrow \mathsf{local\_execution},
\]
which is fail-open with respect to authorization. This contradicts the fail-closed semantics required by Section~5.

\paragraph{Reasoning Space vs.\ Execution Space.}
The above failures follow from a fundamental architectural separation. Orchestrators operate in \emph{reasoning space}: they manipulate plans, intents, and coordination state. The Action Authorization Boundary operates in \emph{execution space}: it mediates the moment an action instance is about to occur. No amount of reasoning-space coordination can substitute for execution-space refusal. This distinction is structural and aligns directly with the boundary formalized in Section~3.

\begin{figure}[t]
  \centering
  \fbox{\parbox{0.9\linewidth}{
    \centering
    \textbf{Reasoning Space vs.\ Execution Space}\\[0.5em]
    Orchestrators constrain plans and coordination; authorization requires mediation at the execution boundary.
  }}
  \caption{Structural mismatch between multi-agent orchestration and execution-time authorization.}
  \label{fig:orchestration_vs_aab}
\end{figure}

\paragraph{Scope and Non-Goals.}
This analysis does not claim that orchestration systems are unnecessary or ineffective for coordination, load balancing, or multi-agent planning. Rather, it establishes that such systems cannot satisfy execution-time authorization invariants without introducing a canonical action representation and a mandatory execution boundary. Coordination and authorization are orthogonal concerns.

\paragraph{Conclusion.}
Agentic gateways, context systems, and multi-agent orchestration violate non-bypassability, determinism, replayability, and fail-closed behavior by construction. Any attempt to repair these deficiencies requires the introduction of an execution-time authorization boundary, thereby reconstituting the AAB. Coordination is not authorization, and no orchestration layer can subsume execution-time refusal without collapsing into the architecture already defined in Sections~3–7.

\emph{Therefore, multi-agent orchestration is insufficient as an alternative design, and this insufficiency follows directly from the Action Authorization Boundary defined in Section~3.}

\begin{table*}[t]
\centering
\small
\begin{tabular}{lccccc}
\hline
\textbf{Alternative Class} &
\textbf{Non-bypassable} &
\textbf{Deterministic} &
\textbf{Transport} &
\textbf{Replayable} &
\textbf{Fail-Closed}
\\
 &
\textbf{Execution} &
\textbf{Authorization} &
\textbf{Independent} &
 &
 \\
\hline
IAM / RBAC / Zero Trust
& X & X & X & X & X
\\
Observability-Only Systems
& X & X & OK & X & X
\\
Protocol-Embedded Enforcement
& X & X & X & X & X
\\
Agentic Gateways / Orchestration
& X & X & OK & X & X
\\
\hline
\end{tabular}
\caption{Invariant violation matrix for alternative architectural classes evaluated in Section~\ref{sec:design-rationale}. An architecture is sufficient if and only if it satisfies \emph{all} invariants defined in Sections~\ref{sec:aab}--\ref{sec:provenance}. Any single violation is decisive and cannot be repaired without introducing a canonical action representation and a mandatory execution-time authorization boundary ~\cite{1}, ~\cite{2}, ~\cite{3}, ~\cite{4}, ~\cite{6}, ~\cite{7}, ~\cite{8}, ~\cite{9}, ~\cite{11}, ~\cite{12}.}
\label{tab:invariant-violation-matrix}
\end{table*}

\paragraph{Section-Level Conclusion.}
Once execution-time governance invariants are accepted as necessary, all alternative designs fail for structural reasons rather than implementation deficiencies. Identity-centric systems govern principals rather than action instances; observability systems record effects rather than decisions; protocol-embedded checks terminate at message acceptance; orchestration systems coordinate reasoning rather than execution. The Action Authorization Boundary therefore follows inevitably as a necessary architectural boundary for autonomous execution, not as a design preference or implementation choice.

Table~\ref{tab:invariant-violation-matrix} summarizes the invariant violations established formally in Sections~9.1–9.4.

\paragraph{Note.}
Each violation follows directly from the corresponding subsection in Section~\ref{sec:design-rationale}.

\section{Discussion}
\label{sec:discussion}

This section derives system-level consequences from the invariants and guarantees established in Sections~\ref{sec:aab}–\ref{sec:provenance}. It introduces no new mechanisms, claims, or requirements, and it does not re-argue the necessity of the Action Authorization Boundary (AAB) or Canonical Action Representation (CAR) ~\cite{1}, ~\cite{3}, ~\cite{4}, ~\cite{6}. Instead, it traces the implications of these results when placed in broader system contexts, under the explicit constraint that all conclusions must follow logically from already-proven properties.

\paragraph{Scope and Method.}
The discussion is intentionally narrow. We reason only from the invariant set previously defined, non-bypassability at execution time, deterministic authorization over canonical actions, policy-agnostic transport independence, provenance-complete replayability, and fail-closed behavior under partial failure, and examine their consequences when these invariants are assumed as fixed architectural facts. No assumptions are made about agent intelligence, policy quality, organizational maturity, or ecosystem adoption. Where conditional statements are made, they are stated explicitly as such.

\paragraph{What This Section Does Not Do.}
This section does not predict technological trajectories, advocate standards, evaluate market readiness, or claim improvements in trust, compliance, or safety beyond what follows directly from provenance and determinism. It does not assert that autonomous agents will evolve in any particular direction, nor that the architecture presented here should be universally adopted. Components outside the trusted computing base, including agents, policies, tools, and operators, remain assumed fallible, consistent with Section~\ref{sec:enforcement}. Any interpretation beyond execution-time authorization semantics is out of scope.

\paragraph{Analytical Framing.}
Each subsection treats the established invariants as constraints on system boundaries rather than as features of a particular implementation. The analysis distinguishes rigorously between execution and authorization, observation and provenance, and transport and semantics, following the separations formalized earlier. Where parallels to existing systems are mentioned, they are introduced only as historical precedents for how minimal semantic interfaces emerge after architectural necessity is established, not as analogies or claims of equivalence.

\paragraph{No New Guarantees.}
Importantly, this section does not extend the guarantee surface of the system. All properties discussed here are restatements, consequences, or boundary conditions of results already proven. The discussion therefore serves to situate the architecture within broader system contexts without weakening, generalizing, or inflating its claims.

\paragraph{Structure.}
Section~\ref{sec:discussion-enterprise} examines the implications of execution-time authorization invariants for enterprise system boundaries that already assume non-bypassability, replay, and fail-closed semantics. Section 10.2 maps the same invariants onto agent behaviors under programmatic invocation and distributed failure, without asserting that such evolution will occur. Section~\ref{sec:discussion-standardization} discusses conditions under which a minimal semantic interface may become portable across systems, framed strictly as a consequence of provenance and replay requirements. The section concludes by reinforcing that these implications follow directly from prior results and introduce no additional claims.

\subsection{Implications for Enterprise System Boundaries}
\label{sec:discussion-enterprise}

The relevance of the Action Authorization Boundary (AAB) to enterprise environments follows directly from the execution-time authorization invariants established in Sections~\ref{sec:aab}–\ref{sec:provenance}, rather than from any agent-specific operational context. This subsection derives consequences for enterprise system boundaries by mapping those invariants onto architectural assumptions already embedded in financial, operational, and safety-critical systems. No new guarantees are introduced, and no claims are made about adoption dynamics, organizational readiness, or compliance outcomes beyond what provenance and determinism already imply.

\paragraph{Pre-existing Enterprise Invariants.}
Enterprise systems that manipulate money, code, identity records, infrastructure state, or procurement commitments already rely on three implicit invariants:
\begin{enumerate}
    \item \emph{Non-bypassability}: effectful operations are mediated by an authorization point that cannot be circumvented without violating system correctness;
    \item \emph{Replayability}: past decisions can be reconstructed and re-evaluated under updated policies or audit conditions;
    \item \emph{Fail-closed semantics}: absence of authorization, context, or policy results in denial rather than execution.
\end{enumerate}
These invariants are not artifacts of autonomous agents; they predate them and are foundational to enterprise correctness. They are enforced today through human-time controls such as approval workflows, change-management systems, and transactional boundaries ~\cite{11}, ~\cite{12}, ~\cite{13}.

\paragraph{Human-Time Authorization as an Architectural Assumption.}
Traditional enterprise systems implicitly assume that authorization occurs at human time scales ~\cite{12}. Let $H$ denote a human approval process and $E$ an effectful operation. Existing architectures assume a causal ordering
\[
H \;\rightarrow\; E,
\]
where $H$ is slow, explicit, and auditable. Autonomous agents violate this assumption by collapsing intent generation and execution into a single computational step:
\[
I \;\rightarrow\; A = \mathrm{Canon}(I) \;\rightarrow\; \mathsf{Exec}(A),
\]
with no intervening boundary unless explicitly introduced. This collapse is not an operational misconfiguration; it is an architectural gap. Without an execution-time authorization boundary, the invariants listed above cannot be preserved once human-time mediation is removed.

\paragraph{Restoring a Missing Boundary.}
The AAB restores an architectural boundary that enterprise systems already assume but that autonomous execution bypasses. As formalized in Section~\ref{sec:enforcement}, authorization is computed as
\[
d = \mathcal{B}(A, P, S),
\]
and execution proceeds if and only if $d = \textsf{PERMIT}$. This reintroduces a mandatory mediation point at execution time without reintroducing human latency. Importantly, this is not a new control plane layered on top of enterprise systems; it is the minimal boundary required to preserve invariants those systems already rely on.

\paragraph{Separation of Decision and Effect.}
Enterprise audit and control systems distinguish sharply between \emph{what was decided} and \emph{what occurred}. Observability pipelines record effects, while approval systems record decisions. Autonomous agents blur this distinction unless an explicit boundary enforces it ~\cite{7}, ~\cite{16}. The AAB maintains this separation by persisting $(A, P, S, d)$ independently of execution outcome, ensuring that denied actions are as observable and replayable as permitted ones. This property cannot be derived from logs, traces, or post-hoc analysis alone, as established in Section~\ref{sec:alt-observability}.

\paragraph{Failure Semantics and Enterprise Safety.}
Enterprise systems are engineered to fail closed under partial failure precisely because fail-open behavior produces irreversible harm (e.g., unauthorized transfers or deployments). Autonomous execution without an AAB defaults to fail-open: missing policy, unavailable context, or transient failures result in execution by default. By contrast, the AAB enforces the implication
\[
\mathcal{B}(A,P,S) = \bot \;\Rightarrow\; \textsf{DENY},
\]
which aligns autonomous execution semantics with existing enterprise safety assumptions rather than introducing new ones.

\paragraph{Scope and Explicit Non-Claims.}
This subsection does not claim that enterprises are “ready” for agents, that agents are desirable in specific domains, or that the AAB guarantees regulatory compliance, trust, or correctness of policies. It does not evaluate agent reasoning quality or organizational processes. The claim is strictly architectural: once enterprises rely on non-bypassability, replayability, and fail-closed semantics, autonomous execution without an execution-time authorization boundary is unsound.

\paragraph{Conclusion.}
Enterprise systems already encode the invariants formalized in Sections~\ref{sec:aab}–\ref{sec:provenance}; autonomous agents merely expose a missing execution-time boundary by violating the assumption of human-mediated control. The Action Authorization Boundary restores that boundary without altering enterprise guarantees or introducing new ones. Its relevance to enterprise environments therefore follows inevitably from existing architectural commitments, not from any speculative properties of agents or their deployment contexts ~\cite{11}, ~\cite{12}, ~\cite{13}.

\subsection{Implications for the Structure of AI-Driven Systems}
\label{sec:discussion-future}

This subsection derives consequences for the architectural structure of AI-driven systems from the invariants established in Sections~\ref{sec:aab}–\ref{sec:provenance}. It does not predict system evolution, advocate design directions, or introduce new guarantees. Instead, it establishes that if autonomous agents come to exhibit properties already common to distributed services, programmatic invocation, retry semantics, composability, and exposure to partial failure, then the execution-time authorization invariants proven in this paper apply unchanged. The argument is conditional and invariant-preserving: it maps existing results forward without extending their scope ~\cite{5}, ~\cite{6}.

\paragraph{Conditional Framing and Explicit Non-Prediction.}
We make no claim that agents \emph{will} evolve into autonomous microservices, nor that such an evolution is desirable. The analysis in this subsection is strictly of the form:
\[
\text{If agents satisfy properties } \mathcal{P}, \text{ then invariants } \mathcal{I} \text{ apply.}
\]
Here, $\mathcal{I}$ refers to the invariant set defined in Sections~\ref{sec:aab}–\ref{sec:provenance}, and $\mathcal{P}$ consists of architectural properties already well understood in distributed systems. No additional assumptions are introduced.

\paragraph{Agents as Programmatically Invoked Components.}
An agent that is invoked programmatically, rather than interactively, is indistinguishable from a service with respect to admission control. Let an agent invocation be modeled as a function
\[
g : X \rightarrow I,
\]
where $X$ denotes input context and $I$ an intent that produces an action
\[
A = \mathrm{Canon}(I).
\]
Once invoked programmatically, the agent becomes part of an execution graph rather than a conversational endpoint. In this setting, the distinction between “agent” and “service” is irrelevant to authorization: both emit effectful actions that must be mediated at execution time to preserve non-bypassability and determinism.

\paragraph{Retry Semantics and Idempotence.}
Autonomous agents operating under failure must retry. Retries introduce ambiguity unless authorization decisions are deterministic and replayable ~\cite{8}, ~\cite{9}. Let $\tau = \langle A_1, A_2, \dots \rangle$ be a sequence of retried actions such that $A_i = A_j$ for $i \neq j$. Without a canonical action representation and a recorded decision
\[
(A,P,S) \mapsto d,
\]
it is impossible to distinguish safe retries from duplicated side effects. The replay semantics defined in Section~\ref{sec:provenance} are therefore necessary whenever retry semantics exist, regardless of whether the executor is labeled an agent or a service.

\paragraph{Composability and Transitive Effects.}
Composability implies that agents may invoke other agents or services, forming chains of execution ~\cite{4}. In such chains, authorization must be transitive and locally decidable. Formally, for a composed execution
\[
A_1 \rightarrow A_2 \rightarrow \dots \rightarrow A_n,
\]
each $A_i$ must satisfy authorization independently:
\[
\forall i,\; \mathcal{B}(A_i,P,S_i) = \textsf{PERMIT}.
\]
This requirement follows directly from non-bypassability. Composition does not weaken invariants; it amplifies their necessity. Without an execution-time boundary at each step, downstream effects become ungovernable, and provenance becomes incomplete.

\paragraph{Distributed Failure and Partial Knowledge.}
In distributed systems, components operate under partial knowledge and partial failure. Agents are no exception. When context $S$ or policy $P$ is unavailable, any component that continues execution violates fail-closed semantics. The AAB enforces the implication
\[
(P,S)\ \text{unavailable} \;\Rightarrow\; \textsf{DENY},
\]
which is identical to the safety requirement imposed on distributed services handling critical state. This property is not agent-specific; it is a consequence of executing under uncertainty.

\paragraph{Separation of Reasoning and Execution.}
A recurring objection is that agent reasoning subsumes execution governance. This is incorrect at the architectural level. Reasoning produces intent; execution produces effects. The boundary between the two is precisely where authorization must occur. This separation, formalized in Section~\ref{sec:aab}, is invariant under changes in how reasoning is implemented. Whether an agent reasons via prompts, plans, or learned policies is irrelevant once it emits $A = \mathrm{Canon}(I)$.

\paragraph{What This Section Does Not Claim.}
This subsection does not claim that agents will replace services, that service-oriented architectures are inevitable, or that existing service governance mechanisms are sufficient for agents. It does not introduce new requirements beyond those already proven. It asserts only that if agents satisfy the same architectural conditions as distributed services, then the same execution-time authorization invariants apply without modification.

\paragraph{Conclusion.}
If autonomous agents are invoked programmatically, retry under failure, compose transitively, and operate in distributed environments, they inherit the governance requirements already formalized for effectful services. Under these conditions, the Action Authorization Boundary and Canonical Action Representation are not extensions or predictions; they are direct consequences of the invariants established earlier. No new guarantees are introduced, and no future behavior is assumed, only the invariants are preserved under conditional mapping ~\cite{5}, ~\cite{6}.

\subsection{Interface Inevitability and Standardization Pressure}
\label{sec:discussion-standardization}

This subsection derives the conditions under which standardization pressure may arise from the invariants proven in Sections~\ref{sec:aab}–\ref{sec:provenance}. It does not propose a standard, predict ecosystem convergence, or argue for industry adoption. Instead, it shows that once execution-time authorization invariants are required across organizational or administrative boundaries, a minimal semantic interface becomes unavoidable. Any standardization pressure follows from this necessity alone, not from architectural ambition or product positioning.

\paragraph{Explicit Scope and Non-Claims.}
This subsection does \emph{not} claim that the Action Authorization Boundary (AAB) should be standardized, that a single implementation is desirable, or that current ecosystems will converge. It does not introduce new guarantees beyond those already established. The analysis is conditional and retrospective: if execution-time authorization decisions must be shared, verified, or replayed across trust boundaries, then a common semantic interface is required. No claim is made absent this condition.

\paragraph{Minimal Semantic Interface Induced by Invariants.}
The invariant set $\mathcal{I}$ defined in Sections~\ref{sec:aab}–\ref{sec:provenance} constrains not only internal system structure but also inter-system interoperability ~\cite{14}, ~\cite{15}, ~\cite{16}. In particular, the following three artifacts must be externally intelligible for invariants to hold across boundaries:
\begin{enumerate}
    \item A \emph{canonical action representation} $A = \mathrm{Canon}(I)$, invariant under syntactic variation (Section~\ref{sec:car});
    \item A \emph{decision function} $(A,P,S) \mapsto d$ with deterministic semantics (Section~\ref{sec:enforcement});
    \item An \emph{immutable decision record} binding $(A,P,S,d)$ for replay and verification (Section~\ref{sec:provenance}).
\end{enumerate}
These elements constitute a minimal semantic interface. Any system claiming provenance-complete replayability or verifiable authorization across domains must expose these artifacts, regardless of internal implementation.

\paragraph{Why Logging Formats and APIs Are Insufficient.}
A common objection is that existing logging or API conventions could suffice. This is false by construction. Logs expose effects, not decisions; APIs expose transport, not semantics. Neither defines a stable mapping from intent to canonical action, nor a frozen authorization context ~\cite{7}, ~\cite{19}. Formally, there exists no injective mapping
\[
(E \text{ or } m) \;\rightarrow\; (A,P,S,d),
\]
where $E$ denotes effects and $m$ protocol messages. Therefore, interoperability based solely on logs or APIs cannot preserve the replay or determinism invariants.

\paragraph{Cross-Boundary Provenance as the Source of Pressure.}
Standardization pressure arises only when decision records must cross boundaries: organizational, administrative, or temporal. Examples include audit handoff, incident investigation, policy re-evaluation under updated constraints, or third-party verification. In each case, the recipient must be able to interpret $(A,P,S,d)$ without access to the original execution environment. This requirement is invariant-driven, not agent-driven ~\cite{14}, ~\cite{15}.

\paragraph{Minimal Semantic Contract (What Must Be Verifiable).}
Cross-boundary verification forces a minimal contract: a small set of fields and guarantees that must remain stable even as implementations differ. A recipient must be able to validate, at minimum:
\begin{itemize}
  \item \textbf{Canonical action form and hash determinism:} the record must include the canonical action bytes $\mathsf{canon}(A)$ (or an unambiguous reference) and its cryptographic digest $h_A=\mathsf{H}(\mathsf{canon}(A))$.
  \item \textbf{Policy version binding:} the decision must reference an immutable policy identifier $v_P$ sufficient to retrieve the exact policy semantics used at decision time.
  \item \textbf{Context snapshot binding:} the decision must reference a context snapshot $S$ (or digest $h_S$) and a schema/version label so that evaluation inputs are interpretable.
  \item \textbf{Decision + witness:} the decision outcome 
 
  $d \in \{\textsf{PERMIT},\textsf{DENY},\textsf{DEFER}\}$ must be present, and if $d=\textsf{DEFER}$, the record must carry a verifiable approval witness (human or service) that satisfies the policy’s approval predicate and resolves the deferred state.
  \item \textbf{Tamper-evident linkage:} the record must be append-only or hash-linked so reordering/removal is detectable under standard integrity assumptions ~\cite{20}
  \item \textbf{\item \textbf{Replay determinism (semantic):} a verifier must be able to re-evaluate $\mathcal{B}(A,P,S)$ and obtain the same outcome under the same inputs.:} a verifier must be able to re-evaluate $\mathcal{B}(A,P,S)$ and obtain the same outcome under the same inputs.
\end{itemize}
This contract is intentionally minimal: it specifies what must be verifiable, not how it is implemented.

\paragraph{Multiple Implementations and Non-Uniqueness.}
Nothing in the invariant set implies a single implementation, deployment model, or control plane. Distinct systems may implement different policy languages, execution substrates, or performance optimizations. The requirement is solely that the externalized semantic contract be preserved. Paradoxically, heterogeneity \emph{increases} convergence pressure: third-party verifiers cannot maintain bespoke interpreters for $N$ incompatible provenance dialects, so interoperability favors a stable record schema even under competitive implementations.

\paragraph{Historical Precedent (Non-Analogical).}
In prior systems domains, minimal semantic contracts emerged only after architectural necessity was established. TLS standardized a portable wire contract across heterogeneous cryptographic stacks; OAuth standardized delegation tokens across identity providers; and OpenTelemetry standardized trace semantics across vendors. In each case, interoperability was not achieved by enforcing identical implementations, but by stabilizing the externally verifiable interface. The same pressure applies here: cross-boundary governance requires a portable decision/provenance record even when execution engines diverge ~\cite{14}, ~\cite{15}, ~\cite{16}.

\section{Limitations and Future Work}
\label{sec:limitations}

This section formally delimits the boundaries of execution-time governance as defined in Sections~\ref{sec:aab}–\ref{sec:provenance}. Its purpose is threefold. First, it specifies which properties the Action Authorization Boundary (AAB) explicitly does \emph{not} guarantee. Second, it establishes that these limitations are not omissions or design trade-offs, but unavoidable consequences of the invariants already proven. Third, it characterizes future work exclusively as invariant-preserving extensions that consume, but do not weaken, the canonical action and authorization semantics defined earlier ~\cite{5}, ~\cite{6}.

This section does not introduce new architectural claims, expand the trusted computing base, or re-argue the necessity of the AAB. All results here are derived consequences of previously established invariants: non-bypassability at execution time (Section~\ref{sec:aab}), deterministic authorization over canonical actions (Section~\ref{sec:car}), provenance-complete replayability (Section~\ref{sec:provenance}), and fail-closed enforcement semantics (Section~\ref{sec:enforcement}). Where a property is not guaranteed, we show that guaranteeing it would require violating at least one of these invariants.

We distinguish explicitly between (i) properties that are \emph{out of scope by design} and orthogonal to execution-time authorization, and (ii) properties that are \emph{provably unattainable} without collapsing the architectural boundary defined in this paper. Future work is restricted to extensions that operate strictly downstream of canonicalization and authorization decisions, preserving determinism, replayability, and failure semantics.

\subsection{Threat Model Boundaries and Non-Goals}
\label{sec:threat-model}

This subsection defines the threat model boundaries of the AAB and enumerates formal non-goals. These boundaries are architectural, not empirical: they arise from the placement of the authorization boundary in execution space and from the deliberate exclusion of semantic interpretation and agent cognition from the trusted computing base.

\paragraph{Trusted Computing Base.}
The trusted computing base (TCB) consists solely of the canonicalization function
\[
A = \mathrm{Canon}(I),
\]
the authorization function
\[
d = \mathcal{B}(A, P, S),
\]
and the immutable decision record binding $(A, P, S, d)$ as defined in Sections~\ref{sec:car} and~\ref{sec:enforcement}. Agents, language models, planners, policy authors, tool implementations, transport protocols, and human operators are explicitly outside the TCB and are assumed to be fallible and potentially adversarial ~\cite{11}, ~\cite{12}.

\paragraph{Policy Correctness (Non-Goal).}
The system does not guarantee that a policy $P$ is correct, complete, or aligned with any external objective. Authorization determinism guarantees that decisions are reproducible with respect to $(A, P, S)$, not that $P$ encodes the “right” behavior. Guaranteeing policy correctness would require semantic interpretation of intent or outcomes, which is intentionally excluded from the TCB to preserve transport independence and replayability ~\cite{1}.

\paragraph{Semantic Accuracy (Non-Goal).}
The AAB does not guarantee that the canonical action $A$ accurately reflects the semantic intent of the agent or the real-world consequences of execution. Canonicalization enforces syntactic and structural equivalence, not semantic truth. Any attempt to enforce semantic accuracy would require embedding domain-specific interpretation into $\mathrm{Canon}(\cdot)$, violating policy-agnosticism and collapsing determinism under policy evolution.

\paragraph{Malicious or Unsafe Reasoning Upstream (Non-Goal).}
The system does not prevent malicious, unsafe, or incorrect reasoning within agents. Reasoning occurs strictly upstream of canonicalization and is not mediated by the AAB. Governing cognition would require either (i) admitting agent-internal state into the TCB or (ii) enforcing transactional control over reasoning steps, both of which violate the autonomy and composability assumptions established in Section~\ref{sec:aab}.

\paragraph{Malicious Humans (Non-Goal).}
The AAB does not protect against malicious human operators who intentionally deploy unsafe policies, misconfigure enforcement placement, or authorize harmful actions. Human intent is treated as an external input to policy $P$ and deployment configuration. This limitation is shared by all authorization systems and is not specific to autonomous agents.

\paragraph{Formal Limit of Execution-Time Governance.}
We make this boundary explicit.

\begin{lemma}[Limit of Execution-Time Governance]
\label{lem:limit-governance}
No execution-time authorization system can guarantee intent correctness, semantic accuracy, or upstream reasoning safety without either (a) embedding semantic interpretation into the trusted computing base or (b) introducing transactional control over agent cognition. Both violate the invariants of policy-agnosticism, determinism, and replayability established in Sections~\ref{sec:car}–\ref{sec:provenance}.
\end{lemma}

\noindent
\emph{Proof sketch.}
Execution-time governance observes only $A = \mathrm{Canon}(I)$ and state $S$ at the execution boundary. Intent correctness and semantic accuracy depend on properties of $I$ and agent-internal reasoning not preserved under canonicalization. Enforcing them requires expanding the TCB to include semantic interpretation or cognition, which introduces non-determinism and invalidates replay under policy evolution. \hfill $\square$

\paragraph{Scope Closure.}
These non-goals are not deficiencies; they are the consequence of enforcing authorization at execution time rather than reasoning time. The guarantees of the AAB end exactly at the execution boundary. Properties beyond that boundary are either orthogonal or fundamentally incompatible with the invariants already proven ~\cite{5}.

\subsection{Architectural Limits of Execution-Time Governance}
\label{sec:arch-limits}

This subsection establishes the fundamental architectural limits of execution-time governance under the invariants defined in Sections~\ref{sec:aab}–\ref{sec:provenance}. The goal is not to enumerate practical shortcomings, but to prove that certain properties cannot be guaranteed without violating determinism, replayability, policy-agnosticism, or non-bypassability. Each limitation follows directly from the placement of the Action Authorization Boundary (AAB) at the execution boundary and from the canonicalization constraint imposed by $\mathrm{Canon}(\cdot)$.

\paragraph{Preconditions.}
All results in this subsection assume the enforcement semantics of Section~\ref{sec:enforcement}: authorization decisions are computed as
\[
d = \mathcal{B}(A, P, S), \quad A = \mathrm{Canon}(I),
\]
and execution proceeds if and only if $d = \textsf{PERMIT}$. The trusted computing base is restricted as defined in Section~\ref{sec:threat-model}.

\subsubsection{Intent Correctness Is Undecidable Post-Canonicalization}

Canonicalization deliberately erases agent-internal reasoning structure in order to ensure determinism and transport independence. As a consequence, intent correctness is undecidable at the execution boundary ~\cite{5}.

Formally, let $I$ denote an agent-generated intent and $A = \mathrm{Canon}(I)$ its canonical action. Canonicalization induces an equivalence class over intents:
\[
\mathrm{Canon}(I_1) = \mathrm{Canon}(I_2) \;\Rightarrow\; I_1 \sim I_2.
\]
Any property that distinguishes between $I_1$ and $I_2$ but is not preserved under $\mathrm{Canon}(\cdot)$ is unobservable to $\mathcal{B}$. Intent correctness is such a property: it depends on latent goals, intermediate reasoning, or semantic interpretation not encoded in $A$.

\begin{lemma}[Undecidability of Intent Correctness]
\label{lem:intent-undecidable}
For any execution-time authorization function $\mathcal{B}$ that operates solely on $(A,P,S)$, there exists no decision procedure that can distinguish correct from incorrect intent without violating determinism or expanding the trusted computing base.
\end{lemma}

\noindent
\emph{Proof sketch.}
Assume a procedure $\mathcal{D}$ that determines intent correctness from $A$. Then $\mathcal{D}$ must discriminate between intents $I_1 \not\sim I_2$ such that $\mathrm{Canon}(I_1)=\mathrm{Canon}(I_2)$, contradicting the definition of canonicalization. Alternatively, $\mathcal{D}$ must depend on information outside $A$, violating policy-agnosticism or TCB constraints. \hfill $\square$

This limit is architectural, not algorithmic. Improving agent reasoning or policy expressiveness does not recover intent correctness at execution time.

\subsubsection{Upstream Cognition Cannot Be Governed Without Breaking Autonomy}

Execution-time governance operates strictly after agent cognition has produced an intent $I$. Governing cognition itself would require interposing on reasoning steps, plans, or intermediate representations.

Let $\rho$ denote the internal reasoning trace of an agent. Any mechanism that constrains $\rho$ must either (i) include $\rho$ in the trusted computing base or (ii) enforce transactional semantics over cognition. Both options violate core invariants:
\begin{itemize}
    \item Including $\rho$ in the TCB breaks policy-agnosticism and replayability, as reasoning traces are non-deterministic and model-dependent.
    \item Transactional cognition introduces global ordering and synchronization assumptions explicitly rejected in Section~\ref{sec:enforcement}.
\end{itemize}

Therefore, execution-time governance cannot and does not constrain how intents are formed. It constrains only whether their resulting actions may execute.

\subsubsection{Zero Latency Is Incompatible with Deterministic Authorization}

Authorization at the execution boundary necessarily introduces latency proportional to the evaluation of $\mathcal{B}(A,P,S)$. Any claim of zero-latency enforcement implies one of the following:
\begin{enumerate}
    \item Authorization is speculative or cached independently of $S$;
    \item Execution proceeds before $d$ is known;
    \item Authorization is elided entirely.
\end{enumerate}
All three violate invariants established earlier. Speculation breaks determinism under state variation; post-execution authorization violates non-bypassability; elision violates enforcement semantics ~\cite{8}, ~\cite{9}.

\begin{lemma}[Latency Lower Bound]
\label{lem:latency-bound}
Any execution-time authorization system that enforces determinism and fail-closed semantics incurs non-zero decision latency on the execution path.
\end{lemma}

\noindent
\emph{Proof sketch.}
Deterministic authorization requires evaluation of $\mathcal{B}$ over current $(A,P,S)$. Fail-closed behavior requires that execution be contingent on this evaluation. Therefore, execution must wait for the decision. \hfill $\square$

This latency is not an implementation artifact; it is the cost of determinism and replayability.

\subsubsection{Scope Closure}

The limits described above are not deficiencies to be addressed by additional mechanisms. They are the direct consequence of placing governance at the execution boundary and of enforcing canonicalization for determinism and provenance. Any system that claims to overcome these limits must either relax earlier invariants or expand the trusted computing base beyond what is architecturally defensible.

Execution-time governance guarantees exactly what it claims: deterministic, replayable, non-bypassable authorization of canonical actions at execution time. Properties beyond that boundary are either orthogonal or incompatible with those guarantees. The architecture ends precisely here.

\subsection{Deployment and Composition Constraints}
\label{sec:deployment-constraints}

This subsection characterizes the constraints that arise when the execution-time authorization architecture defined in Sections~\ref{sec:aab}–\ref{sec:enforcement} is composed with real systems. These constraints are not weaknesses of the design; they are the necessary conditions under which the stated guarantees hold. We formalize which properties depend on correct placement and composition of the Action Authorization Boundary (AAB), and we prove that violations manifest as bypass, not as graceful degradation. No new guarantees are introduced.

\paragraph{Scope.}
The results in this subsection delimit the boundary between architectural guarantees and deployment correctness. We explicitly separate (i) properties guaranteed by construction from (ii) properties that hold only if the boundary is placed on the execution path. Failures in the latter category are classified as misdeployment, not architectural insufficiency.

\subsubsection{Placement Correctness}

Execution-time governance is meaningful if and only if authorization mediates all external side effects. Let $\mathsf{Exec}(A)$ denote the realization of a canonical action $A=\mathrm{Canon}(I)$ that produces an effect $E$. The enforcement semantics of Section~\ref{sec:enforcement} require the implication:
\[
\mathsf{Exec}(A) \;\Rightarrow\; \exists d = \mathcal{B}(A,P,S) \;\land\; d=\textsf{PERMIT}.
\]

\paragraph{Determinism.}
Authorization is deterministic if repeated evaluations over identical inputs return identical decisions:
\[
\forall (A,P,S),\ \forall t,t',\quad \mathrm{Eval}_{t}(A,P,S)=\mathrm{Eval}_{t'}(A,P,S).
\]

\begin{definition}[Correct Placement]
An Action Authorization Boundary is \emph{correctly placed} if every effectful execution path $\tau$ satisfies:
\[
\forall \mathsf{Exec}(A)\in\tau,\; \mathsf{Exec}(A) \text{ is causally downstream of } \mathcal{B}(A,P,S).
\]
\end{definition}

If this condition is violated, e.g., through direct tool invocation, side channels, or alternate execution paths, then non-bypassability fails by definition. No architectural mechanism can compensate for an authorization boundary that does not mediate execution ~\cite{8}, ~\cite{9}.

\begin{lemma}[Placement Necessity]
\label{lem:placement-necessity}
Non-bypassability at execution time holds if and only if the AAB is placed on the execution path of all effectful actions.
\end{lemma}

\noindent
\emph{Proof sketch.}
If placement holds, execution is contingent on authorization by construction. If placement does not hold, there exists at least one execution path not mediated by $\mathcal{B}$, yielding bypass. \hfill $\square$

This lemma clarifies that bypass under misplacement is not a failure mode to be mitigated; it is the negation of the precondition under which the invariant is defined.

\subsubsection{Multi-Agent Coordination Limits}

The architecture intentionally defines authorization per execution instance, not per global plan. Let $\{\pi_i\}$ denote plans generated by multiple agents and let $\{A_{i,j}\}$ denote the resulting canonical actions. Authorization is evaluated independently:
\[
d_{i,j} = \mathcal{B}(A_{i,j},P,S).
\]

There exists no architectural guarantee relating $\{d_{i,j}\}$ across agents or across time beyond what is encoded in $P$ and $S$. Global coordination properties, such as mutual exclusion across plans, deadlock avoidance, or collective intent safety, are therefore out of scope.

This exclusion is deliberate. Any attempt to authorize plans rather than executions would require:
\begin{itemize}
    \item speculative authorization over non-realized actions, breaking determinism;
    \item global ordering or synchronization across agents, violating transport independence;
    \item rollback semantics over cognition, violating autonomy.
\end{itemize}

Accordingly, multi-agent coordination is an orthogonal concern that may consume authorization outcomes but cannot replace them.

\begin{figure}[t]
  \centering
  \fbox{\parbox{0.9\linewidth}{
    \centering
    \textbf{Composition Boundary.}\\
    Authorization applies to realized actions $A$; coordination applies to plans $\pi$. The boundary is structural.
  }}
  \caption{Separation between execution-time authorization and multi-agent coordination.}
  \label{fig:composition-boundary}
\end{figure}

\subsubsection{Why Retries $\neq$ Authorization}

Distributed systems commonly retry operations under partial failure. Let $A$ be a canonical action whose execution failed transiently ~\cite{8}. Retrying $\mathsf{Exec}(A)$ without re-evaluating $\mathcal{B}(A,P,S)$ violates determinism and replayability when $S$ or $P$ has changed.

\begin{lemma}[Retry Unsoundness]
\label{lem:retry-unsound}
Retries are not a substitute for authorization. Any retry of $\mathsf{Exec}(A)$ that does not re-evaluate $\mathcal{B}(A,P,S)$ risks violating deterministic authorization under state or policy evolution.
\end{lemma}

\noindent
\emph{Proof sketch.}
Determinism requires that authorization be evaluated over the current $(A,P,S)$. A retry that reuses a prior decision implicitly assumes $P$ and $S$ are unchanged, an assumption not guaranteed in distributed systems. \hfill $\square$

Therefore, correct composition requires that retries be treated as fresh execution attempts with fresh authorization decisions. This is an architectural constraint, not an operational recommendation.

\subsubsection{Observability and Composition}

Observability pipelines may record effects, retries, and failures, but they do not alter authorization semantics ~\cite{7}. In particular:
\[
\mathsf{Obs}(E) \;\not\Rightarrow\; \exists d \text{ such that } E \text{ was authorized}.
\]
Composition with observability does not strengthen guarantees and cannot repair misplacement or retry unsoundness. This follows directly from the separation between decision capture and effect observation established in Section~\ref{sec:provenance}.

\paragraph{Scope Closure.}
The constraints in this subsection exhaust the composition surface of execution-time governance. Correct placement is a necessary condition for non-bypassability; per-execution authorization intentionally excludes global coordination; retries must re-enter the authorization boundary; observability remains advisory. None of these constraints weaken prior guarantees, and none can be eliminated without violating invariants already proven.

Execution-time governance guarantees exactly what it mediates. Where it is not placed, where it is not re-entered, or where it is asked to coordinate plans rather than executions, no architectural promise applies. The boundary is precise, and it closes here.

\subsection{Invariant-Preserving Extensions (Future Work)}
\label{sec:future-work}

This subsection delineates extensions that may be layered atop the architecture defined in Sections~\ref{sec:car}–\ref{sec:enforcement} without weakening or modifying its guarantees. Each extension is characterized strictly by how it consumes canonical action representations and authorization decisions already produced by the Action Authorization Boundary (AAB). No extension alters the decision function, introduces speculative authorization, or relaxes determinism, replayability, or fail-closed semantics ~\cite{1}, ~\cite{3}.

Formally, all extensions discussed below operate as functions of the form:
\[
\Phi : (A, P, S, d) \rightarrow X,
\]
where $A=\mathrm{Canon}(I)$, $(P,S)$ denote the policy and state context, $d=\mathcal{B}(A,P,S)$ is the authorization decision, and $X$ is auxiliary metadata or derived signal. None of these extensions redefine $\mathcal{B}$ or introduce alternative execution paths.

\subsubsection{Predictive Gating}

Predictive gating augments execution-time authorization by computing advisory risk estimates prior to decision evaluation. Let $\rho(A,S)$ denote a bounded risk estimator derived from historical provenance:
\[
\rho : (A,S) \rightarrow [0,1].
\]
The authorization function remains unchanged:
\[
d = \mathcal{B}(A,P,S),
\]
but policies may reference $\rho$ as an input signal. Crucially, $\rho$ is computed over canonicalized actions and immutable state snapshots, preserving determinism. If unavailable, authorization defaults to denial, preserving fail-closed behavior.

Predictive gating does not claim semantic correctness or intent understanding. It provides a monotonic signal that may be incorporated into policy evaluation without introducing speculative execution or probabilistic authorization.

\subsubsection{Blast Radius Estimation}

Blast radius estimation derives upper bounds on potential side effects associated with an action. Let $\beta(A)$ denote a static or learned approximation of effect magnitude:
\[
\beta : A \rightarrow \mathcal{B}_{\text{eff}},
\]
where $\mathcal{B}_{\text{eff}}$ is a domain-specific lattice (e.g., financial exposure, resource scope). $\beta$ is computed post-canonicalization and does not alter $A$.

Policies may reference $\beta(A)$ to constrain authorization thresholds, but execution remains contingent solely on $d=\mathcal{B}(A,P,S)$. No rollback, compensation, or speculative execution is introduced. Blast radius estimation therefore strengthens policy expressiveness without modifying enforcement semantics.

\subsubsection{Cross-Boundary Provenance Exchange}

Cross-boundary provenance exchange enables portability of authorization records across organizational or administrative domains. Let $\Pi$ denote an immutable provenance tuple:
\[
\Pi = (A, P, S, d, t),
\]
where $t$ is a monotonic timestamp local to the issuing boundary. Exported provenance does not imply trust or re-authorization; it supports verification and replay in downstream systems.

Re-evaluation in a foreign domain requires explicit re-authorization:
\[
d' = \mathcal{B}(A,P',S').
\]
Thus, provenance exchange preserves transport independence and does not create implicit delegation or transitive trust. Replayability is strengthened without altering execution-time guarantees ~\cite{14}, ~\cite{15}, ~\cite{16}

\begin{figure}[t]
  \centering
  \fbox{\parbox{0.9\linewidth}{
    \centering
    \textbf{Extension Boundary.}\\
    Extensions consume $(A,P,S,d)$ but do not modify $\mathcal{B}$ or execution mediation.
  }}
  \caption{Invariant-preserving extensions operate strictly downstream of authorization decisions.}
  \label{fig:future-work-boundary}
\end{figure}

\paragraph{Exclusions.}
The following are explicitly excluded from this class of extensions:
\begin{itemize}
    \item speculative authorization or pre-approval;
    \item probabilistic execution without decision capture;
    \item plan-level or intent-level authorization;
    \item semantic interpretation within the trusted computing base.
\end{itemize}
Each exclusion follows directly from invariants established earlier and cannot be incorporated without weakening determinism or non-bypassability.

\paragraph{Closure.}
All extensions described above preserve the architectural boundary between authorization and execution. They enrich policy input space, analysis, and interoperability while leaving the core guarantees unchanged. Any future mechanism that alters the decision function, mediates execution conditionally, or substitutes inference for authorization lies outside the scope of execution-time governance as defined in this paper and is therefore excluded.

\subsection*{Closing Boundary Statement}
\label{sec:closing-boundary}

The guarantees established in this paper terminate precisely at the execution-time authorization boundary. For any action instance $A = \mathrm{Canon}(I)$, the system guarantees that authorization is deterministic with respect to $(A, P, S)$, non-bypassable at execution time, provenance-complete for replay, and fail-closed under partial failure, as formalized in Sections~\ref{sec:car}--\ref{sec:enforcement}. No stronger guarantees are implied, and no weaker interpretation is sufficient to preserve these properties.

All limitations identified in this section arise from architectural constraints, not from incomplete mechanisms or missing integrations. Intent correctness, semantic interpretation, upstream reasoning safety, and global coordination are excluded because enforcing them would require extending the trusted computing base beyond execution-time mediation or introducing coupling that violates determinism, replayability, or autonomy. These boundaries are invariant-preserving and cannot be relaxed without collapsing the distinction between reasoning space and execution space established earlier in the paper.

Extensions described as future work consume the outputs of canonicalization and execution-time authorization without altering their semantics. They neither weaken nor replace the Action Authorization Boundary, nor do they modify the definition of authorization itself. As a result, the system’s guarantees are neither aspirational nor extensible by assumption: they end exactly where claimed, apply exactly where defined, and cannot be derived from existing identity, observability, protocol, or orchestration systems without reintroducing the same architectural boundary under a different form.

\section{Conclusion}
\label{sec:conclusion}

Autonomous agents have crossed a qualitative threshold: they no longer merely generate text or recommendations, but initiate actions that mutate real systems, financial state, infrastructure, and organizational processes. Existing agent frameworks, protocols, governance tools, and observability platforms collectively assume that once an agent produces a well-formed proposal, execution is either admissible or recoverable. This assumption fails structurally. Across the landscape examined in Section~\ref{sec:aab}, no existing system defines a canonical unit of action, no system evaluates deterministic authorization at execution time, and no system enforces a universal, non-bypassable boundary between reasoning and irreversible side effects.

This paper establishes execution-time authorization as a necessary architectural boundary rather than an optional control plane. The Action Authorization Boundary (AAB) formalizes the separation between reasoning space and execution space and enforces that separation by construction. By introducing a canonical action representation, deterministic policy evaluation over $(A,P,S)$, fail-closed enforcement semantics, and provenance-complete decision records, the architecture provides guarantees that cannot be derived from identity systems, communication protocols, orchestration runtimes, guardrails, or post-fact observability. These guarantees hold independently of agent implementation, protocol substrate, policy model, or organizational deployment, and they terminate precisely at the execution boundary as defined.

Faramesh instantiates this architecture as a protocol-agnostic execution-time governor. Its contribution is not the introduction of new policy semantics or trust assumptions, but the establishment of a missing structural layer: a mandatory mediation point where proposed actions either execute, defer, or fail deterministically, and where every decision is replayable and auditable under policy evolution. The resulting system does not attempt to solve intent correctness, semantic accuracy, or agent cognition; instead, it restores a control invariant already assumed by financial systems, infrastructure control planes, and safety-critical software, but absent from contemporary agent stacks.

As autonomous systems proliferate across organizational and administrative boundaries, the need for portable, verifiable execution decisions will arise independently of any single implementation. Canonical action representations, deterministic authorization semantics, and immutable decision records define a minimal semantic interface for execution governance. Multiple realizations of this interface may coexist, but without such a boundary, autonomous execution remains either unverifiable or unenforceable. Execution-time authorization is therefore not an optimization, a safety feature, or a deployment preference. It is the architectural condition under which agent autonomy becomes governable computation.

\section*{Key Takeaways}

\begin{itemize}
    \item \textbf{Execution-time authorization is a missing architectural boundary in agent systems.}  
    Existing agent frameworks, protocols, guardrails, and observability tools operate either before or after execution and therefore cannot decide whether an agent-generated action may safely execute.
    
    \item \textbf{Reasoning and execution are distinct computational domains and must be governed separately.}  
    Agent cognition is probabilistic and opaque; execution produces irreversible side effects. Conflating these domains eliminates the possibility of deterministic control.
    
    \item \textbf{Canonical action representation is a prerequisite for enforceable governance.}  
    Without a stable, protocol-agnostic unit of action, authorization decisions cannot be deterministic, replayable, or auditable under policy evolution.
    
    \item \textbf{The Action Authorization Boundary (AAB) enforces non-bypassable execution mediation.}  
    All effectful actions must traverse a mandatory runtime boundary that either permits, defers, or denies execution under fail-closed semantics.
    
    \item \textbf{Deterministic authorization enables replay, audit, and post-incident verification.}  
    Decisions evaluated over $(A,P,S)$ produce identical outcomes under identical conditions and are recorded as immutable provenance artifacts.
    
    \item \textbf{Observability, identity, orchestration, and safety filters cannot substitute for execution control.}  
    These systems lack authority over execution and therefore cannot prevent, defer, or prove the correctness of action execution.
    
    \item \textbf{Intent correctness, semantic accuracy, and agent cognition are intentionally out of scope.}  
    Governing these properties would require expanding the trusted computing base or violating determinism and autonomy guarantees.
    
    \item \textbf{Execution-time governance defines a minimal, portable semantic interface.}  
    Canonical actions, authorization decisions, and decision records form the necessary contract for cross-system and cross-organizational enforcement.
    
    \item \textbf{Faramesh demonstrates that governed autonomy is achievable without constraining agent reasoning.}  
    By enforcing control exclusively at execution time, agents remain autonomous while their real-world effects remain governable.
\end{itemize}

\section*{Acknowledgements}
The author conducted this work independently and without institutional funding. Faramesh is the reference implementation used to validate and demonstrate the proposed Action Authorization Boundary (AAB) architecture.

\appendix

\section*{Appendices}
\addcontentsline{toc}{section}{Appendices}

% ---------------------------
\section{Extended Formal Definitions}
\label{app:formal-definitions}

This appendix collects auxiliary definitions and notational conventions referenced throughout the paper.

\begin{definition}[Reasoning Space and Execution Space]
\label{def:spaces}
\emph{Reasoning space} denotes computation that produces proposals without directly causing external side effects. \emph{Execution space} denotes computation that may cause irreversible or externally observable side effects (state mutation, financial transfer, privileged API invocation, etc.).
\end{definition}

\begin{definition}[Intent and Canonical Action]
\label{def:intent-action}
An \emph{intent} $I$ is an agent-produced proposal. A \emph{canonical action} $A$ is the deterministic representation obtained by canonicalization:
\[
A := \mathrm{Canon}(I).
\]
Define the canonical action digest:
\[
h_A := H(A) = H(\mathrm{Canon}(I)).
\]
Only $A$ (and its digest $h_A$) are eligible for authorization at the Action Authorization Boundary (AAB).
\end{definition}

\begin{definition}[Decision Function and Decision Space]
\label{def:decision-fn}
Given policy $P$ and state $S$, the AAB computes a decision
\[
d = \mathcal{B}(A, P, S), \qquad d \in \{\textsf{PERMIT},\textsf{DEFER},\textsf{DENY}\}.
\]
\end{definition}

\begin{definition}[Execution Predicate]
\label{def:exec-predicate}
Let $\mathsf{Exec}(A)$ denote any execution that produces external side effects corresponding to action $A$.
Conformance requires \emph{decision-before-execution}:
\[
\mathsf{Exec}(A) \Rightarrow \exists r:\ \ \wedge\ d(r)=\textsf{PERMIT}\ \wedge\ r \prec \mathsf{Exec}(A).
\]
(r \text{ is a valid decision record for } A).

Equivalently, no side effect may occur unless a prior decision record with $\textsf{PERMIT}$ exists.
\end{definition}

\begin{definition}[Decision Record]
\label{def:decision-record}
A \emph{decision record} is an immutable tuple sufficient for deterministic re-evaluation and tamper-evidence:
\[
r_i = (\mathsf{seq}_i, h_{A,i}, v_{P,i}, h_{S,i}, d_i, t_i, \mathsf{prev\_hash}_i),
\]
where:
\[
h_{A,i} := H(A_i) = H(\mathrm{Canon}(I_i)),
\qquad
h_{S,i} := H(S_i),
\]
and $v_{P,i}$ is an immutable policy version identifier (sufficient to retrieve the exact decision semantics at time $t_i$).
The field $\mathsf{prev\_hash}_i$ hash-links records into an append-only chain, making removal or reordering detectable.
\end{definition}

% ---------------------------
\section{AAB State Machine}
\label{app:aab-state-machine}

This appendix provides a reference state machine consistent with the enforcement semantics (Section~\ref{sec:enforcement}).

\subsection{States}

\begin{description}
  \item[\textsf{Proposed}] A proposal has been received and canonicalized into $A=\mathrm{Canon}(I)$.
  \item[\textsf{Evaluating}] The AAB evaluates $d=\mathcal{B}(A,P,S)$ under the current $(P,S)$.
  \item[\textsf{Permitted}] $d=\textsf{PERMIT}$; an execution artifact is issued; execution may proceed.
  \item[\textsf{Deferred}] $d=\textsf{DEFER}$; execution is suspended pending an external signal (e.g., approval).
  \item[\textsf{Denied}] $d=\textsf{DENY}$; execution is prohibited.
\end{description}

\subsection{Transitions}

\begin{align*}
\textsf{Proposed} &\rightarrow \textsf{Evaluating} \\
\textsf{Evaluating} &\rightarrow \textsf{Permitted} \quad \text{iff } d=\textsf{PERMIT} \\
\textsf{Evaluating} &\rightarrow \textsf{Deferred} \quad \text{iff } d=\textsf{DEFER} \\
\textsf{Evaluating} &\rightarrow \textsf{Denied} \quad \text{iff } d=\textsf{DENY} \\
\textsf{Deferred} &\rightarrow \textsf{Evaluating} \quad \text{upon resolution signal, re-evaluating } (A,P,S)
\end{align*}

\subsection{Fail-Closed Transition Rule}

Any failure that prevents a sound decision record from being produced yields:
\[
\neg \exists r_i \Rightarrow \textsf{Denied},
\]
i.e., absence of a valid decision record implies denial (fail-closed).

% ---------------------------
\section{Canonicalization Examples}
\label{app:canonicalization-examples}

This appendix provides representative normalization examples illustrating that semantically equivalent proposals map to identical canonical actions.

\subsection{Example 1: Parameter Ordering and Surface Syntax}

\noindent \textbf{Intent variants}
\begin{align*}
I_1 &: \texttt{refund(customer=X, amount=500, currency=USD)} \\
I_2 &: \texttt{refund(amount=500, currency=USD, customer=X)}
\end{align*}

\noindent \textbf{Canonicalization}
\[
\mathrm{Canon}(I_1) = \mathrm{Canon}(I_2) = A.
\]

\subsection{Example 2: Equivalent Resource Identifiers}

\noindent \textbf{Intent variants}
\begin{align*}
I_3 &: \texttt{deploy(service="billing", env="prod")} \\
I_4 &: \texttt{deploy(service="svc\_billing", env="production")}
\end{align*}

\noindent \textbf{Canonicalization (assuming deterministic alias map in TCB)}
\[
\mathrm{Canon}(I_3) = \mathrm{Canon}(I_4) = A'.
\]

\subsection{Example 3: Canonical Hash Binding}

For any intent $I$ and its canonical action $A=\mathrm{Canon}(I)$,
\[
h_A = H(A)=H(\mathrm{Canon}(I))
\]
is stable across representational drift in $I$ so long as $\mathrm{Canon}(\cdot)$ is unchanged.

% ---------------------------
\section{Glossary}
\label{app:glossary}

\begin{description}
  \item[Agent] An entity that produces intents $I$ and may attempt to cause execution in external systems.
  \item[Intent] An agent-produced proposal $I$ prior to canonicalization and outside the authorization semantics.
  \item[Action] An execution-relevant operation represented canonically as $A=\mathrm{Canon}(I)$.
  \item[Tool] Any external capability that can realize side effects corresponding to an action.
  \item[Policy] A specification $P$ interpreted by $\mathcal{B}$ mapping $(A,S)$ to a decision $d$.
  \item[CAR] Canonical Action Representation.
  \item[AAB] Action Authorization Boundary.
  \item[TCB] Trusted Computing Base.
  \item[Provenance] Decision-complete records sufficient for deterministic replay and tamper-evidence.
  \item[Fail-closed] Default denial under partial failure of the authorization path.
\end{description}

% ---------------------------
\section{Security Proof Sketch}
\label{app:security-proof-sketch}

This appendix provides a one-page proof sketch organized by threat class.

\subsection{Threat: Adversary Bypasses the AAB}

\noindent \textbf{Adversary goal.} Cause $\mathsf{Exec}(A)$ without producing a valid decision record with $d=\textsf{PERMIT}$. \\

\noindent \textbf{Counter (architectural).}
Executors are defined to lack any unmediated access path: no tool is reachable with credentials, network paths, or invocation privileges that bypass the AAB-mediated execution predicate (Definition~\ref{def:exec-predicate}). Under this assumption, every effectful invocation is causally downstream of a \textsf{PERMIT} artifact.

\subsection{Threat: Agent Mutates Proposals to Evade Rules}

\noindent \textbf{Adversary goal.} Produce multiple syntactic forms $I_1,I_2$ that represent the same execution-relevant operation, attempting to induce divergent decisions. \\

\noindent \textbf{Counter (canonicalization).}
Canonicalization collapses equivalent forms:
\[
\mathrm{Canon}(I_1)=\mathrm{Canon}(I_2)=A,
\]
reducing policy evaluation to a single canonical unit. Policy application operates over $A$ rather than over raw proposal forms.

\subsection{Threat: Forged or Tampered Provenance}

\noindent \textbf{Adversary goal.} Fabricate, alter, or reorder decision records to misrepresent authorization history. \\

\noindent \textbf{Counter (record binding).}
Decision records $r_i$ are hash-chained via $\mathsf{prev\_hash}_i$ (Definition~\ref{def:decision-record}), yielding tamper-evidence for any mutation of the record stream within the TCB.

\paragraph{Record-to-Semantics Binding.}
Although the authorization function is defined over full objects $(A,P,S)$, the decision record stores portable references $(h_A, v_P, h_S)$, where $h_A = H(A)$ and $h_S = H(S)$, and $v_P$ uniquely identifies the exact policy semantics $P$ used at decision time. Replay verification is performed by resolving $v_P \mapsto P$, reconstructing $A$ and $S$, and checking $H(A)=h_A$ and $H(S)=h_S$ before re-evaluating $\mathcal{B}(A,P,S)$.

\end{document}